\documentclass[aps,prl,twocolumn,superscriptaddress]{revtex4-2}

\usepackage{graphicx}

\usepackage{amssymb}
\usepackage{amsmath}
\usepackage{verbatim}

\usepackage{algorithmicx}
\usepackage{algpseudocode}

\begin{document}

\title{Infrared Organization and Critical Cognitive Field Formation in Transformer Dynamics}

\author{Byung Gyu Chae}

\affiliation{Electronics and Telecommunications Research Institute, 218 Gajeong-ro, Yuseong-gu, Daejeon 34129, Republic of Korea
\\ bgchae@etri.re.kr}

%\date{\today}

\begin{abstract}

Large language models exhibit remarkable emergent behaviors, yet
the physical mechanism governing their large-scale collective
dynamics remains poorly understood.
Cognitive Field Theory predicts that learning reorganizes the
complex collective spectrum, whose relaxation projection defines the
time-scale density of states (TDOS) governing infrared memory
organization.
The resulting memory self-energy renormalizes the cognitive
forgetting gap,
\[
r_{\rm cog}
=
r-\Sigma(0),
\]
where \(r\) denotes the bare forgetting rate and
\(\Sigma(0)\) is the memory self-energy.
As the forgetting gap decreases, the collective cognitive
susceptibility
\[
\chi(0)\propto r_{\rm cog}^{-1}
\]
is strongly enhanced, signaling the formation of a macroscopic
cognitive field.

In the present work, we investigate whether these collective
observables can be identified directly in Transformer dynamics.
Using publicly available Pythia language models, we extract the
complex collective spectrum from layer Jacobians throughout training,
prompt ensembles, network depth, and model scale, allowing the
TDOS, memory self-energy, memory kernel, and infrared critical
exponent to be measured quantitatively and the corresponding
forgetting-gap dynamics to be characterized.

The measurements reveal a pronounced infrared reorganization of the
complex collective spectrum.
Across the investigated models, learning substantially redistributes
the slow-relaxation spectrum while producing a nearly flat but weakly
infrared-singular TDOS,
\[
\rho(\lambda)\sim\lambda^\beta,
\qquad
\beta \lesssim 0,
\]
with characteristic fitted exponents of order \(\beta\sim-0.1\).
The corresponding memory kernels exhibit robust long-memory scaling
close to the marginal \(1/t\) form expected in the flat-TDOS limit.
The collective observables further reveal a critical formation
process: the memory self-energy reaches a transient maximum during
early learning, corresponding to the minimum forgetting gap and
maximum susceptibility before relaxation toward a metastable
near-critical regime.
Prompt-resolved and token-subspace measurements further show that
TDOS distributions obtained from distinct local Jacobians converge
toward a common macroscopic relaxation spectrum, while their
low-\(\lambda\) sectors exhibit common infrared scaling.

Taken together, the reproducibility of the macroscopic TDOS and its
weakly infrared-singular scaling across training, prompt ensembles,
network depth, and Transformer model scales supports a universal
infrared collective organization.
These observations provide a quantitative experimental realization
of the collective observables predicted by Cognitive Field Theory
and identify infrared organization of the complex collective spectrum
as a robust collective principle of Transformer dynamics.

\end{abstract}

\maketitle

\section{I. Introduction}

Large language models based on the Transformer architecture have
demonstrated remarkable emergent capabilities, including reasoning,
abstraction, in-context learning, and long-range contextual memory \cite{1,2,3,4,5}.
Despite these advances, the physical mechanism governing their
large-scale collective organization remains largely unknown.
Most existing studies describe Transformer computation as a
high-dimensional nonlinear mapping or an attention-based optimization
process \cite{6,7,8,9,10,11}.
While these viewpoints successfully characterize information
processing, they provide limited insight into how large-scale
collective organization arises from the underlying Transformer
dynamics.

From the perspective of statistical physics, macroscopic collective
behavior is governed not only by microscopic interactions but also by
the organization of collective dynamical modes \cite{12,13,14}.
This naturally raises the question of whether large neural networks
possess measurable collective observables analogous to those used to
describe nonequilibrium many-body systems.

Cognitive Field Theory was recently proposed as an effective field
description of collective cognition \cite{15}.
Rather than identifying intelligence with individual neurons or model
parameters, the theory predicts that cognition emerges from the
organization of a complex collective spectrum.
Its joint spectral density,
\begin{equation}
\rho(\lambda,\omega),
\end{equation}
provides a unified description of collective relaxation and
circulation dynamics.
Projection onto the relaxation sector determines the time-scale
density of states (TDOS), memory kernel, memory self-energy,
cognitive forgetting gap, and collective susceptibility, whereas
projection onto the circulation sector governs recursive temporal
organization through collective phase dynamics.

The primary objective of the present work is therefore to determine
whether these collective observables can be identified directly from
Transformer dynamics.
To this end, we extract the Jacobian spectrum throughout Transformer
training and reconstruct the corresponding complex collective spectrum
from its complex eigenvalues.
The relaxation projection provides a direct quantitative
characterization of the infrared memory dynamics, whereas the
circulation projection characterizes the temporal organization of the
collective dynamics through the phase spectrum.

Using publicly available Pythia language models spanning a broad range
of model scales \cite{16,17,18},
we observe a pronounced infrared reorganization of the complex
collective spectrum during learning.
The measured relaxation projection consistently yields a universal
long-memory kernel across different layer propagation depths.
Furthermore, prompt-resolved and token-subspace measurements reveal
that TDOS distributions obtained from distinct local Jacobians
converge toward a common macroscopic relaxation spectrum, while their
low-$\lambda$ sectors exhibit a common infrared scaling form.
These measurements establish the TDOS as a reproducible collective
dynamical quantity and support the infrared fixed-point organization
of its slow-relaxation sector.

Beyond the infrared spectral reorganization, both the relaxation and
circulation sectors exhibit a well-defined collective organization
process during learning.
The memory self-energy exhibits a pronounced transient maximum,
corresponding within Cognitive Field Theory to the smallest cognitive
forgetting gap and the largest collective susceptibility.
Simultaneously, the phase spectrum undergoes a transient organization
toward the low-frequency sector before relaxing into a stable
multiscale temporal hierarchy.
Together these observations indicate the transient formation of a
memory-dressed macroscopic cognitive field, which subsequently
stabilizes in a protected near-critical operating regime.

Rather than treating large language models solely as computational
architectures \cite{19,20,21}, the present work investigates them as nonequilibrium
many-body systems whose internal dynamics can be analyzed using the
language of collective field theory.
The observed combination of infrared spectral reorganization,
scale-free long-memory dynamics, recursive temporal organization,
memory self-energy renormalization, transient critical enhancement,
and persistent infrared fixed-point scaling provides the first
quantitative experimental realization of the collective observables
predicted by Cognitive Field Theory.

The remainder of this paper is organized as follows.
Section II develops the theoretical framework connecting Cognitive
Field Theory with Transformer dynamics.
It introduces the response-theoretic description of collective
relaxation and circulation dynamics, shows how the Transformer
architecture naturally supports globally coupled collective
organization, derives the reconstruction of cognitive-field observables
from the complex Jacobian spectrum, and establishes the infrared
fixed-point predictions tested throughout the remainder of the paper.
Section III presents the Transformer models, measurement protocol,
and the evolution of the relaxation spectrum during training.
Section IV establishes the macroscopic reproducibility of the TDOS
through prompt-ensemble and token-subspace convergence analyses.
Section V examines the robustness of the infrared organization across
Transformer layers.
Section VI investigates the robustness of the collective dynamics
across different Pythia model scales.
Section VII tests the predicted depth-independent infrared
organization using long-range Transformer propagation.
Section VIII investigates the circulation sector of the complex
collective spectrum, including collective temporal phase organization
and its implications for coherent macroscopic cognitive dynamics.
Finally, Section IX discusses the implications of these results for
the emergence of macroscopic cognitive fields in nonequilibrium
dynamical systems.

%=========================================
\section{II. Cognitive Field Description of Transformer Dynamics}

To investigate the collective dynamical principles proposed by
Cognitive Field Theory \cite{15}, we analyze publicly available
Transformer language models from a response-theoretic perspective.
Rather than describing Transformer computation solely in terms of
individual neurons, attention heads, or network parameters, we
reinterpret the hidden-state dynamics as a hierarchy of interacting
collective modes whose organization may be reconstructed from the
Jacobian spectrum.

This section first summarizes the response-theoretic framework of
Cognitive Field Theory.
We then show that even a single Transformer block naturally realizes a
globally coupled collective dynamical system through self-attention,
while successive residual blocks hierarchically reorganize the same
collective hidden state along network depth.
Finally, we demonstrate how the measured complex Jacobian spectrum
provides direct experimental access to the relaxation and circulation
sectors predicted by Cognitive Field Theory, thereby establishing the
bridge between the theoretical framework and the Transformer
measurements presented in the following sections.

\subsection{A. Response-theoretic structure of Cognitive Field Theory}

Cognitive Field Theory describes cognition as the collective dynamics
of a nonequilibrium many-body system \cite{15}.
The central object is a macroscopic cognitive field
\(
\phi(t)
\),
which represents the slowly varying collective state generated by a
large number of interacting internal modes.

The collective modes are characterized by a complex dynamical
spectrum.
We denote a mode by
\begin{equation}
X_\alpha(t)
=
\mathcal A_\alpha(t)e^{i\vartheta_\alpha(t)},
\end{equation}
with complex generator
\begin{equation}
\nu_\alpha
=
\lambda_\alpha+i\omega_\alpha,
\end{equation}
where
\(
\lambda_\alpha
\)
is the relaxation rate and
\(
\omega_\alpha
\)
is the circulation frequency.
The real and imaginary parts therefore constitute two complementary
descriptions of the same collective dynamics.
The relaxation sector determines the infrared organization of the
collective spectrum, the memory kernel, and the susceptibility,
whereas the circulation sector determines the temporal phase dynamics,
collective synchronization, and the emergence of coherent temporal
organization.

A minimal coupled description of the macroscopic field and the latent
collective modes is
\begin{align}
\partial_t\phi
&=
-r\phi
+
\int d\lambda\,d\omega\,
g(\lambda,\omega)
X_{\lambda,\omega}
+
h(t)
+
\eta_\phi,
\label{eq:cognitive_field_coupled}
\\
\partial_tX_{\lambda,\omega}
&=
-
(\lambda+i\omega)
X_{\lambda,\omega}
+
g(\lambda,\omega)\phi
+
\eta_{\lambda,\omega}.
\label{eq:collective_mode_coupled}
\end{align}
Here
\(
r>0
\)
is the bare forgetting rate,
\(
g(\lambda,\omega)
\)
is the coupling between the macroscopic cognitive field and the
collective modes,
\(
h(t)
\)
is an external cognitive input, and the
\(
\eta
\)
terms represent fluctuations.

The solution of
Eq.~(\ref{eq:collective_mode_coupled}) is
\begin{equation}
\begin{aligned}
X_{\lambda,\omega}(t)
=
&X_{\lambda,\omega}(0)
e^{-(\lambda+i\omega)t}
\\
&
+
g(\lambda,\omega)
\int_0^t dt'\,
e^{-(\lambda+i\omega)(t-t')}
\phi(t')
+
\text{noise}.
\end{aligned}
\label{eq:integrated_collective_mode}
\end{equation}
Substituting this expression into
Eq.~(\ref{eq:cognitive_field_coupled}) gives a non-Markovian equation
for the macroscopic field:
\begin{equation}
\partial_t\phi(t)
=
-r\phi(t)
+
\int_0^t dt'\,
K(t-t')\phi(t')
+
h(t)
+
\eta_{\rm eff}(t).
\label{eq:nonmarkovian_cognitive_field}
\end{equation}

Projecting the complex spectrum onto the relaxation sector gives the
memory kernel
\begin{equation}
K(t)
=
\Theta(t)
\int_0^\Lambda d\lambda\,
\rho(\lambda)e^{-\lambda t},
\label{eq:memorykernel}
\end{equation}
where
\begin{equation}
\rho(\lambda)
=
g^2(\lambda)D(\lambda)
\label{eq:weighted_tdos}
\end{equation}
is the coupling-weighted time-scale density of states.
The function
\(
D(\lambda)
\)
counts collective modes with relaxation rate
\(
\lambda
\),
while
\(
g^2(\lambda)
\)
weights their coupling to the macroscopic cognitive field.

For an infrared power-law spectrum
\begin{equation}
\rho(\lambda)
\sim
\lambda^\beta,
\label{eq:tdos_powerlaw}
\end{equation}
the long-time memory kernel scales as
\begin{equation}
K(t)
\sim
t^{-(\beta+1)}.
\label{eq:kernel_powerlaw}
\end{equation}
For the dynamics considered here, the relevant candidate infrared
regime is a near-marginal scaling class characterized by a nearly
flat, weakly infrared-enhanced TDOS,
\begin{equation}
\rho_*(\lambda)
\sim
\lambda^{\beta_*},
\qquad
\beta_*\lesssim0,
\qquad
|\beta_*|\ll1.
\end{equation}
The exactly flat case $\beta_*=0$ represents the marginal limit,
for which $K(t)\sim t^{-1}$.

The same relaxation spectrum determines the retarded memory
self-energy,
\begin{equation}
\Sigma_R(\Omega)
=
\int_0^\Lambda d\lambda\,
\frac{\rho(\lambda)}
{\lambda-i\Omega},
\label{eq:cognitive_self_energy}
\end{equation}
and the dressed cognitive susceptibility,
\begin{equation}
\chi_R(\Omega)
=
\frac{1}
{-i\Omega+r-\Sigma_R(\Omega)}.
\label{eq:cognitive_susceptibility}
\end{equation}
The static renormalization defines the cognitive forgetting gap,
\begin{equation}
r_{\rm cog}
=
r-\Sigma_R(0).
\label{eq:cognitive_forgetting_gap}
\end{equation}

As slow relaxation modes accumulate in the infrared,
\(
\Sigma_R(0)
\)
increases,
\(
r_{\rm cog}
\)
is suppressed, and the collective cognitive susceptibility is enhanced.
When the forgetting gap approaches zero, the response becomes
collectively amplified, signaling the critical formation of the
macroscopic cognitive field.
Following this collective transition, the system evolves toward a
stable memory-dressed near-critical operating regime.

Projecting instead onto the circulation sector gives the temporal phase
dynamics.
Writing
\begin{equation}
X_\alpha
=
\mathcal A_\alpha e^{i\vartheta_\alpha},
\qquad
\phi
=
\mathcal A e^{i\psi},
\end{equation}
the coupling between a collective mode and the macroscopic cognitive
field generates a phase equation of the form
\begin{equation}
\partial_t\vartheta_\alpha
=
\omega_\alpha
+
\frac{g_\alpha\mathcal A}{\mathcal A_\alpha}
\sin(\psi-\vartheta_\alpha).
\label{eq:field_mediated_phase_equation}
\end{equation}
The macroscopic field is itself reconstructed from the collective
modes, so the coupling is reciprocal:
\begin{equation}
\phi
\rightleftarrows
\{X_\alpha\}.
\end{equation}

Eliminating the macroscopic field produces an effective mode--mode
phase dynamics,
\begin{equation}
\partial_t\vartheta_\alpha
=
\omega_\alpha
+
\sum_\beta
\mathcal J_{\alpha\beta}^{\rm eff}
\sin(\vartheta_\beta-\vartheta_\alpha),
\label{eq:phase_dynamics}
\end{equation}
where the effective coupling is mediated by the memory-dressed
susceptibility,
\begin{equation}
\mathcal J_{\alpha\beta}^{\rm eff}(\Omega)
=
g_\alpha g_\beta
\chi_R(\Omega).
\label{eq:effective_phase_coupling}
\end{equation}
The collective phase organization may be quantified by
\begin{equation}
\mathcal Qe^{i\psi}
=
\frac{
\sum_\alpha
\mathcal A_\alpha e^{i\vartheta_\alpha}
}{
\sum_\alpha \mathcal A_\alpha
}.
\label{eq:phase_order_parameter}
\end{equation}
Here
\(
\mathcal Q\in[0,1]
\)
measures the coherence of the participating collective modes.

The relaxation and circulation sectors are therefore complementary
projections of the same complex collective spectrum.
The relaxation projection determines persistence, memory dressing,
and critical susceptibility, while the circulation projection
determines temporal coordination among the activated modes.
Together they define the amplitude and phase organization of the
macroscopic cognitive field.

\subsection{B. Collective dynamical structure of Transformer architectures}

Modern Transformer language models exhibit remarkable capabilities in
long-context processing, in-context learning, compositional reasoning,
and higher-order cognitive behavior.
From the perspective of Cognitive Field Theory, these observations
suggest that learning organizes large-scale collective dynamics
extending beyond isolated neuron responses or purely local
feed-forward computation.

The essential architectural mechanism is self-attention.
Let
\begin{equation}
H_l
\in
\mathbb R^{s\times d}
\end{equation}
denote the hidden-state matrix entering the
\(l\)-th Transformer block, where
\(s\)
is the retained sequence length and
\(d\)
is the hidden dimension.
For a single attention head,
\begin{equation}
Q_l
=
H_lW_Q^{(l)},
\qquad
K_l
=
H_lW_K^{(l)},
\qquad
V_l
=
H_lW_V^{(l)}.
\label{eq:single_head_qkv}
\end{equation}
The attention score and attention matrix are
\begin{equation}
S_l
=
\frac{Q_lK_l^{\rm T}}
{\sqrt{d_k}},
\end{equation}
and
\begin{equation}
A_l
=
{\rm softmax}(S_l+\mathcal M),
\label{eq:single_head_attention_matrix}
\end{equation}
where
\(
\mathcal M
\)
denotes the causal attention mask.
The output of the single head is
\begin{equation}
Y_l
=
A_lV_l.
\label{eq:single_head_output}
\end{equation}
At the token level,
\begin{equation}
(Y_l)_i
=
\sum_{j\leq i}
(A_l)_{ij}(V_l)_j
\label{eq:causal_token_coupling}
\end{equation}
for a causal Transformer.
Thus, even a single attention head establishes state-dependent
interactions among all causally accessible tokens.

Equation~(\ref{eq:causal_token_coupling}) also admits a natural
collective-mode interpretation.
Suppose that the propagated value representations are expanded in a
collective basis,
\begin{equation}
(V_l)_j
=
\sum_\alpha
c_{j\alpha}
X_\alpha,
\label{eq:value_collective_expansion}
\end{equation}
where
$X_\alpha$
denotes a collective propagation mode and
$c_{j\alpha}$
its participation coefficient in the
$j$-th token representation.
Substituting Eq.~(\ref{eq:value_collective_expansion}) into
Eq.~(\ref{eq:causal_token_coupling}) gives
\begin{equation}
(Y_l)_i
=
\sum_\alpha
\left(
\sum_j
(A_l)_{ij}
c_{j\alpha}
\right)
X_\alpha.
\label{eq:attention_collective_projection}
\end{equation}
Defining the effective collective coupling
\begin{equation}
g^{\rm eff}_{i\alpha}
=
\sum_j
(A_l)_{ij}
c_{j\alpha},
\label{eq:effective_collective_coupling}
\end{equation}
the attention output becomes
\begin{equation}
(Y_l)_i
=
\sum_\alpha
g^{\rm eff}_{i\alpha}
X_\alpha.
\label{eq:attention_collective_response}
\end{equation}
The attention matrix therefore acts as a state-dependent effective
coupling that projects distributed hidden representations onto
collective propagation modes.

The collective nature of a single attention head becomes particularly
clear from its linear response.
A perturbation of the hidden state simultaneously redistributes the
attention coefficients and modifies the propagated value
representations:
\begin{equation}
\delta Y_l
=
(\delta A_l)V_l
+
A_l(\delta V_l).
\label{eq:attention_variation}
\end{equation}
The value variation is
\begin{equation}
\delta V_l
=
\delta H_lW_V^{(l)},
\end{equation}
while the attention variation contains both query- and key-mediated
contributions,
\begin{equation}
\delta A_l
=
\frac{\partial A_l}{\partial Q_l}
\delta Q_l
+
\frac{\partial A_l}{\partial K_l}
\delta K_l,
\label{eq:attention_weight_variation}
\end{equation}
with
\begin{equation}
\delta Q_l
=
\delta H_lW_Q^{(l)},
\qquad
\delta K_l
=
\delta H_lW_K^{(l)}.
\end{equation}

Consequently, the attention Jacobian decomposes schematically as
\begin{equation}
\begin{aligned}
J_l^{\rm attn}
&=
\frac{\partial Y_l}{\partial H_l}
=
\underbrace{
\frac{\partial Y_l}{\partial V_l}
\frac{\partial V_l}{\partial H_l}
}_{\text{value-mediated response}}
\\
&
+
\underbrace{
\frac{\partial Y_l}{\partial A_l}
\frac{\partial A_l}{\partial Q_l}
\frac{\partial Q_l}{\partial H_l}
}_{\text{query-mediated response}}
+
\underbrace{
\frac{\partial Y_l}{\partial A_l}
\frac{\partial A_l}{\partial K_l}
\frac{\partial K_l}{\partial H_l}
}_{\text{key-mediated response}}.
\end{aligned}
\label{eq:attention_jacobian_paths}
\end{equation}

The softmax derivative further couples all accessible keys for a fixed
query:
\begin{equation}
\frac{\partial (A_l)_{ij}}
{\partial (S_l)_{ik}}
=
(A_l)_{ij}
\left[
\delta_{jk}
-
(A_l)_{ik}
\right].
\label{eq:softmax_jacobian}
\end{equation}
A perturbation of one attention score therefore changes not only one
edge of the attention matrix but the entire normalized attention
distribution associated with the corresponding query.
The Jacobian of even a single attention head is thus a dense
state-dependent collective response operator.

Equation~(\ref{eq:attention_jacobian_paths}) may be viewed as a sum over
distinct interaction paths through which a hidden-state perturbation
changes the output.
This does not imply that the Transformer explicitly executes repeated
particle-like collisions.
Rather, the single-block Jacobian already contains the simultaneous
contributions of many globally coupled response channels.
At the level of an effective field description, these channels may be
organized as multiple-scattering or Dyson-type response contributions.

The collective coupling is therefore not created by the use of
multiple heads.
It is already present in a single attention matrix.
Multiple heads instead provide parallel interaction channels with
different learned projections:
\begin{equation}
Y_l^{(a)}
=
A_l^{(a)}V_l^{(a)},
\qquad
a=1,\ldots,N_h,
\end{equation}
and
\begin{equation}
{\rm MHA}_l(H_l)
=
{\rm Concat}
\left(
Y_l^{(1)},\ldots,Y_l^{(N_h)}
\right)
W_O^{(l)}.
\label{eq:multihead_output}
\end{equation}
Each head therefore defines a distinct learned interaction geometry,
while the output projection combines these parallel collective
channels.

A complete Transformer block combines attention, feed-forward
processing, normalization, and residual propagation.
For compactness, we denote the vectorized hidden state by
\begin{equation}
h_l
\equiv
{\rm vec}(H_l).
\label{eq:vectorized_hidden_state}
\end{equation}
The residual block may then be written schematically as
\begin{equation}
h_{l+1}
=
h_l
+
F_l(h_l),
\label{eq:transformer_layer}
\end{equation}
where
\(
F_l
\)
contains all nontrivial transformations performed by the
\(l\)-th block.

The local block Jacobian is
\begin{equation}
J_l
=
\frac{\partial h_{l+1}}
{\partial h_l}
=
I
+
\mathcal L_l,
\label{eq:transformer_jacobian}
\end{equation}
where
\begin{equation}
\mathcal L_l
\equiv
\frac{\partial F_l}{\partial h_l}.
\label{eq:block_interaction_jacobian}
\end{equation}
Because
\(
\mathcal L_l
\)
contains attention-mediated token coupling, feature-space mixing,
normalization response, and feed-forward nonlinearities, its
eigenmodes are collective directions of the complete hidden-state
subspace rather than responses of individual neurons.

This observation is important for the interpretation of the measured
single-block spectrum.
A single Transformer block already realizes a globally coupled
collective dynamical system.
Accordingly, collective relaxation modes and a nontrivial TDOS may be
identified from an individual block Jacobian without requiring
propagation through the full network.
Subsequent blocks do not create collective dynamics from nothing;
rather, they progressively reorganize an already collective hidden
state.

The complete Transformer is obtained by composing the blockwise
hidden-state transformations,
\begin{equation}
h_L
=
F_{L-1}^{\rm full}
\circ
F_{L-2}^{\rm full}
\circ
\cdots
\circ
F_0^{\rm full}(h_0),
\label{eq:depth_composed_mapping}
\end{equation}
where
\begin{equation}
F_l^{\rm full}(h_l)
=
h_l+F_l(h_l).
\end{equation}
A small perturbation propagates according to
\begin{equation}
\delta h_{l+1}
=
J_l\delta h_l.
\end{equation}
Applying the chain rule gives
\begin{equation}
\delta h_L
=
J_{L-1}
J_{L-2}
\cdots
J_0
\delta h_0,
\end{equation}
so the composed propagation Jacobian is
\begin{equation}
J_{0\rightarrow L}
=
J_{L-1}
J_{L-2}
\cdots
J_0.
\label{eq:transformer_composed}
\end{equation}

Using
\(
J_l=I+\mathcal L_l
\),
the composed Jacobian becomes
\begin{equation}
J_{0\rightarrow L}
=
\prod_{l=0}^{L-1}
\left(
I+\mathcal L_l
\right),
\label{eq:ordered_residual_product}
\end{equation}
where later-block operators act from the left.
Expanding the ordered product gives
\begin{equation}
\begin{aligned}
J_{0\rightarrow L}
=
&I
+
\sum_l\mathcal L_l
+
\sum_{l_2>l_1}
\mathcal L_{l_2}\mathcal L_{l_1}
\\
&
+
\sum_{l_3>l_2>l_1}
\mathcal L_{l_3}
\mathcal L_{l_2}
\mathcal L_{l_1}
+\cdots.
\end{aligned}
\label{eq:depth_ordered_expansion}
\end{equation}

The first-order terms describe transformations generated within
individual blocks.
The higher-order products describe depth-ordered propagation paths in
which a perturbation produced at an earlier block is successively
reorganized by later blocks.
The composed Jacobian therefore contains the cumulative response of
all such ordered paths.

Equation~(\ref{eq:depth_ordered_expansion}) is mathematically analogous
to an ordered Dyson-type expansion.
The analogy does not imply that a feed-forward Transformer solves an
explicit time-recursive Dyson equation.
Instead, it states that the cumulative response of a globally coupled
hidden state can be decomposed into direct, once-reorganized,
twice-reorganized, and higher-order depth-ordered contributions.

A single Transformer block generates a simultaneous globally coupled
collective response through attention-mediated interactions.
By contrast, successive Transformer blocks do not create independent
responses but progressively reorganize an already collective hidden
state propagated through the residual stream.
The resulting hierarchy therefore represents a depth-ordered
collective response expansion.
At the effective field level, this hierarchical organization naturally
admits a Dyson-type resummation of collective responses.

Although the Transformer contains no explicit recurrent loop between
blocks, the residual hidden-state stream continuously carries the
accumulated collective state forward:
\begin{equation}
h_0
\longrightarrow
h_1
\longrightarrow
\cdots
\longrightarrow
h_L.
\label{eq:depth_state_refinement}
\end{equation}
Each block acts on a hidden state that already contains the cumulative
contributions of all preceding blocks.
The relevant recurrence is therefore not a literal return to an
earlier architectural state, but a repeated refinement of the same
residual state along network depth.

In a continuous-depth approximation, the residual update may be
written as
\begin{equation}
h_{l+1}-h_l
\simeq
\Delta s\,
\mathcal F(h(s),s),
\label{eq:discrete_depth_flow}
\end{equation}
where
\(
s
\)
is a continuous depth coordinate.
Taking
\(
\Delta s\rightarrow0
\)
gives
\begin{equation}
\partial_s h(s)
=
\mathcal F(h(s),s).
\label{eq:continuous_depth_flow}
\end{equation}
A linear perturbation obeys
\begin{equation}
\partial_s\delta h(s)
=
\mathcal L(s)\delta h(s),
\end{equation}
with formal solution
\begin{equation}
\delta h(s)
=
\mathcal T_s
\exp
\left[
\int_0^s ds'\,
\mathcal L(s')
\right]
\delta h(0),
\label{eq:depth_ordered_exponential}
\end{equation}
where
\(
\mathcal T_s
\)
denotes ordering along depth.
This is the continuous-depth counterpart of the discrete composed
Jacobian.

Within Cognitive Field Theory, the hidden state propagated through the
residual stream may be interpreted as a depth-dependent collective
cognitive state.
The input first perturbs and activates a weighted set of collective
directions.
Attention, nonlinear mixing, and residual propagation then repeatedly
reorganize the resulting state across network depth.
The composed Jacobian measures the cumulative sensitivity of this
collective state to perturbations introduced at an earlier block.

The Jacobian may therefore be evaluated either for an individual
Transformer block or for a composed sequence of blocks.
The former characterizes the collective response already generated
within a single globally coupled block.
The latter captures the hierarchical reorganization of this response
over increasing propagation depth.

The relation between the marginal edge of the Jacobian spectrum and
the infrared scaling of the TDOS, together with its possible
connection to residual propagation, is derived in Appendix~A.

\subsection{C. Time-scale density of states as a collective dynamical observable}

Diagonalization of the Transformer Jacobian provides direct access to
the collective propagation modes of the hidden-state dynamics.
Let
\begin{equation}
Jv_\alpha=\zeta_\alpha v_\alpha,
\end{equation}
where the eigenvalue is generally complex,
\begin{equation}
\zeta_\alpha
=
|\zeta_\alpha|e^{i\theta_\alpha}.
\end{equation}
For a propagation interval $\Delta s$, we define the corresponding
relaxation rate and circulation frequency as
\begin{equation}
\lambda_\alpha
=
-\frac{1}{\Delta s}\ln|\zeta_\alpha|,
\qquad
\omega_\alpha
=
\frac{1}{\Delta s}\arg\zeta_\alpha .
\label{eq:jacobian_relaxation_circulation}
\end{equation}
Equivalently,
\begin{equation}
\zeta_\alpha
=
\exp[-(\lambda_\alpha+i\omega_\alpha)\Delta s],
\label{eq:discrete_collective_generator}
\end{equation}
where the sign of $\omega_\alpha$ depends on the adopted phase
convention.

Modes with $|\zeta_\alpha|<1$ have $\lambda_\alpha>0$ and decay across
the measured propagation interval, whereas modes with
$|\zeta_\alpha|\simeq1$ have $\lambda_\alpha\simeq0$ and constitute
the slow infrared sector.  The eigenvalue angle describes the
rotational or circulation component of the same collective
propagation mode.

The full complex spectrum may be represented by the joint spectral
density
\begin{equation}
\rho(\lambda,\omega)
=
\frac{1}{N}
\sum_\alpha
\delta(\lambda-\lambda_\alpha)
\delta(\omega-\omega_\alpha).
\label{eq:joint_complex_tdos}
\end{equation}
Its relaxation projection,
\begin{equation}
\rho_{\rm rel}(\lambda)
=
\int d\omega\,\rho(\lambda,\omega)
=
\frac{1}{N}
\sum_\alpha
\delta(\lambda-\lambda_\alpha),
\label{eq:measured_tdos}
\end{equation}
defines the measured time-scale density of states, while
\begin{equation}
\rho_{\rm circ}(\omega)
=
\int d\lambda\,\rho(\lambda,\omega)
\label{eq:circulation_density}
\end{equation}
provides the complementary circulation spectrum.

The TDOS characterizes the collective relaxation spectrum supported by
the dynamical architecture.  In analogy with a density of states in
many-body physics, it remains a well-defined spectral quantity
independently of subsequent collective ordering.  For a Transformer,
this follows naturally from the nontrivial Jacobian spectrum already
generated by the globally coupled block dynamics described in
Sec.~II.B.

We therefore distinguish the architecture-supported initial spectrum
$\rho_0(\lambda)$ from the spectrum
$\rho(\lambda;t_{\rm train})$ measured during learning.
Optimization reorganizes this dynamical density of states by
redistributing spectral weight among relaxation time scales.
The same infrared exponent may consequently persist before and after
learning even when the spectral weight, crossover structure, and
macroscopic collective response differ substantially.

Through the response relations derived in Sec.~II.A, this spectral
reorganization modifies the memory dressing and collective
susceptibility, with enhanced slow-mode weight contributing more
strongly to the long-time response.
The infrared spectral class and macroscopic cognitive-field formation
therefore characterize distinct aspects of the collective dynamics.

A second issue concerns the state dependence of the measured TDOS.
In the present analysis, the Jacobian is evaluated at the hidden-state
configuration generated by a particular input sequence and within a
finite token subspace.  Different inputs therefore probe different
operating points of the Transformer dynamics, and in general
\begin{equation}
J_S(h)
\neq
J_{S'}(h').
\label{eq:different_local_jacobians}
\end{equation}
The corresponding eigenvalues and eigenvectors need not coincide.

The TDOS, however, is defined in relaxation-rate space rather than in
the hidden-state coordinate space in which the local measurement is
performed.  If it represents a well-defined collective spectral
quantity of the system, its macroscopic distribution should therefore
be recoverable from different state-space realizations, up to finite
sampling fluctuations.  For a hidden state $h$ measured within a token
subspace $S$, we define
\begin{equation}
\rho_{S,h}(\lambda)
=
\frac{1}{N_S}
\sum_{\alpha=1}^{N_S}
\delta
\left(
\lambda-\lambda_\alpha^{(S,h)}
\right).
\label{eq:local_tdos}
\end{equation}
Although different inputs generally produce distinct local Jacobians,
a well-defined macroscopic TDOS requires their relaxation-rate
distributions to approach a common spectral form,
\begin{equation}
\rho_{S_a,h_i}(\lambda)
\simeq
\rho_{S_b,h_j}(\lambda),
\label{eq:tdos_local_convergence}
\end{equation}
for sufficiently representative measurements and consistent
normalization.

This is analogous to a density of states in condensed-matter systems:
microscopic configurations or local sampling conditions may differ,
while the density of states remains a reproducible spectral property
of the macroscopic system.  Here the input sequence selects the
hidden-state configuration at which the local response is measured,
whereas the TDOS characterizes the resulting distribution of
collective modes in relaxation-rate space.  Input-state robustness is
therefore a necessary empirical criterion for identifying the measured
TDOS with a collective dynamical density of states rather than with a
state-specific Jacobian spectrum.

The finite token window introduces an additional sampling dependence.
Let $S_N$ denote the dynamical subspace constructed from $N$ retained
tokens.  Increasing the sequence length enlarges the measured state
space,
\begin{equation}
S_{N_1}
\subset
S_{N_2}
\subset
\cdots ,
\end{equation}
and increases the number of collective relaxation modes represented in
the Jacobian.  If the system possesses a well-defined macroscopic
dynamical spectrum, the finite-subspace TDOS should approach
\begin{equation}
\rho_{S_N,h}(\lambda)
\longrightarrow
\rho_{\rm bulk}(\lambda),
\qquad
N\rightarrow\infty,
\label{eq:bulk_tdos_limit}
\end{equation}
up to finite-subspace fluctuations and normalization.

Input robustness and sequence-length convergence therefore test two
complementary aspects of the same physical requirement: whether TDOS
measurements performed at different locations and resolutions of the
hidden-state dynamics recover a common collective spectrum.  This
bulk spectral convergence is distinct from infrared universality,
which concerns the asymptotic scaling of
$\rho_{\rm bulk}(\lambda)$ as $\lambda\rightarrow0^+$ and is considered
in the following subsection.

\subsection{D. Infrared scaling and fixed-point organization of the TDOS}

Having established the TDOS as a collective dynamical density of
states, we now examine the scaling structure of its slow sector.
The infrared limit corresponds to
\begin{equation}
\lambda\rightarrow0^+,
\qquad
\tau=\lambda^{-1}\rightarrow\infty ,
\label{eq:infrared_relaxation_limit}
\end{equation}
where the dynamics is governed by increasingly slow collective modes.

The complete TDOS need not be scale invariant.
At finite relaxation rates, it may retain detailed dependence on the
architecture, hidden-state configuration, propagation depth, and
observation window.
In the infrared, however, the spectrum may approach an asymptotic
scaling form
\begin{equation}
\rho(\lambda)
\sim
C\lambda^\beta,
\qquad
\lambda\rightarrow0^+.
\label{eq:infrared_tdos_powerlaw}
\end{equation}
The exponent $\beta$ characterizes the infrared population of
collective relaxation modes.
For the marginal case $\beta\simeq0$,
\begin{equation}
\rho(\lambda\rightarrow0^+)
\rightarrow
\rho_0 ,
\label{eq:marginal_flat_tdos}
\end{equation}
and the response relations derived in Sec.~II.A give
\begin{equation}
\rho(\lambda)\sim\lambda^\beta
\quad\Longrightarrow\quad
K(t)\sim t^{-(1+\beta)},
\end{equation}
so that a flat infrared TDOS produces the scale-free memory law
$K(t)\sim t^{-1}$.

This asymptotic structure admits a natural renormalization-group
description.
Dynamical coarse graining toward longer time scales eliminates
progressively faster modes and rescales the relaxation rate as
\begin{equation}
\lambda
\longrightarrow
b^{-z}\lambda,
\qquad
b>1,
\label{eq:lambda_rg_rescaling}
\end{equation}
where $z$ is the dynamical scaling exponent.
The TDOS may then be written schematically under RG transformation as
\begin{equation}
\rho\left(\lambda;\{u_i\}\right)
=
b^{y_\rho}
\rho\left(
b^z\lambda;
\{b^{y_i}u_i\}
\right),
\label{eq:tdos_rg_transformation}
\end{equation}
where $y_\rho$ is the scaling dimension of the spectral density and
$\{u_i\}$ denotes microscopic perturbations associated with the input
state, local coordinates, finite token subspace, network depth, and
other ultraviolet details.

Perturbations with $y_i<0$ become irrelevant under repeated coarse
graining, and the infrared spectrum approaches a fixed-point form
satisfying
\begin{equation}
\mathcal R_b[\rho_*]
=
\rho_* ,
\label{eq:tdos_fixed_point_condition}
\end{equation}
up to the corresponding scaling dimension.
Different microscopic realizations $a$ may therefore have distinct
full spectra while sharing the same infrared exponent,
\begin{equation}
\rho_a(\lambda)
\sim
C_a\lambda^{\beta_*},
\qquad
\lambda\rightarrow0^+.
\label{eq:universal_ir_exponent}
\end{equation}
After removal of nonuniversal amplitudes and crossover scales, their
infrared sectors may converge to a common normalized form,
\begin{equation}
\widehat{\rho}_a(\lambda)
\longrightarrow
\widehat{\rho}_*(\lambda),
\qquad
\lambda\rightarrow0^+.
\label{eq:normalized_ir_fixed_point}
\end{equation}

For the dynamics considered here, the relevant candidate fixed point
is the marginal class
\begin{equation}
\beta_*\simeq0,
\qquad
\rho_*(\lambda)\sim\lambda^0,
\label{eq:marginal_ir_fixed_point}
\end{equation}
corresponding to the $K(t)\sim t^{-1}$ long-time response derived
above.

The RG flow defined here should be distinguished from optimization
during training.  Learning changes the underlying dynamical operator
and reorganizes its TDOS, whereas the RG transformation describes the
asymptotic coarse graining of a given spectrum toward
$\lambda\rightarrow0^+$.  Thus different training stages may exhibit
substantial spectral reorganization while remaining within the same
infrared universality class.

The analyses below therefore examine both the evolution of the full
TDOS during learning and the convergence of its low-$\lambda$ sector
across training stages, input states, token subspaces, network depths,
and model scales.

%===========================================
\section{III. Experimental Observation of Cognitive Field Formation}

We now test the theoretical framework developed in Sec.~II using the
publicly available Pythia model family.
Layer Jacobian spectra are measured throughout training to track the
formation of the infrared collective dynamics predicted by Cognitive
Field Theory.
From the measured complex spectra, we reconstruct the relaxation and
circulation sectors that govern collective memory organization and
temporal organization, respectively.

The relaxation sector is analyzed first.
From the eigenvalue magnitudes, we reconstruct the time-scale density
of states and quantify the memory self-energy, cognitive
forgetting gap, collective susceptibility, infrared critical exponent,
and memory kernel.
The circulation sector, obtained from the eigenvalue phases, is
examined subsequently in Sec.~VIII to determine how Transformer
learning reorganizes the temporal organization of the collective
dynamics.

\begin{table*}[t]
\caption{
Experimental configuration of the Pythia models analyzed throughout
this work.
}
\label{tab:models}
\centering
\begin{tabular}{lccc}
\hline\hline
 & Pythia-70M & Pythia-410M & Pythia-1.4B\\
\hline
Architecture & Decoder-only Transformer & Decoder-only Transformer & Decoder-only Transformer\\
Trainable parameters & 70M & 410M & 1.4B\\
Transformer layers & 6 & 24 & 24\\
Hidden dimension & 512 & 1024 & 2048\\
Attention heads & 8 & 16 & 16\\
Context length & 2048 & 2048 & 2048\\
Training corpus & \multicolumn{3}{c}{The Pile ($\sim300$B tokens)}\\
Global batch size & \multicolumn{3}{c}{2,097,152 tokens}\\
Optimizer & \multicolumn{3}{c}{AdamW}\\
Learning-rate schedule & \multicolumn{3}{c}{Linear warmup + cosine decay}\\
Released checkpoints & \multicolumn{3}{c}{154 checkpoints}\\
\hline\hline
\end{tabular}
\end{table*}

\begin{figure*}[t]
\centering
\includegraphics[width=1.0\textwidth, trim=0cm 5.7cm 0cm -0.5cm]{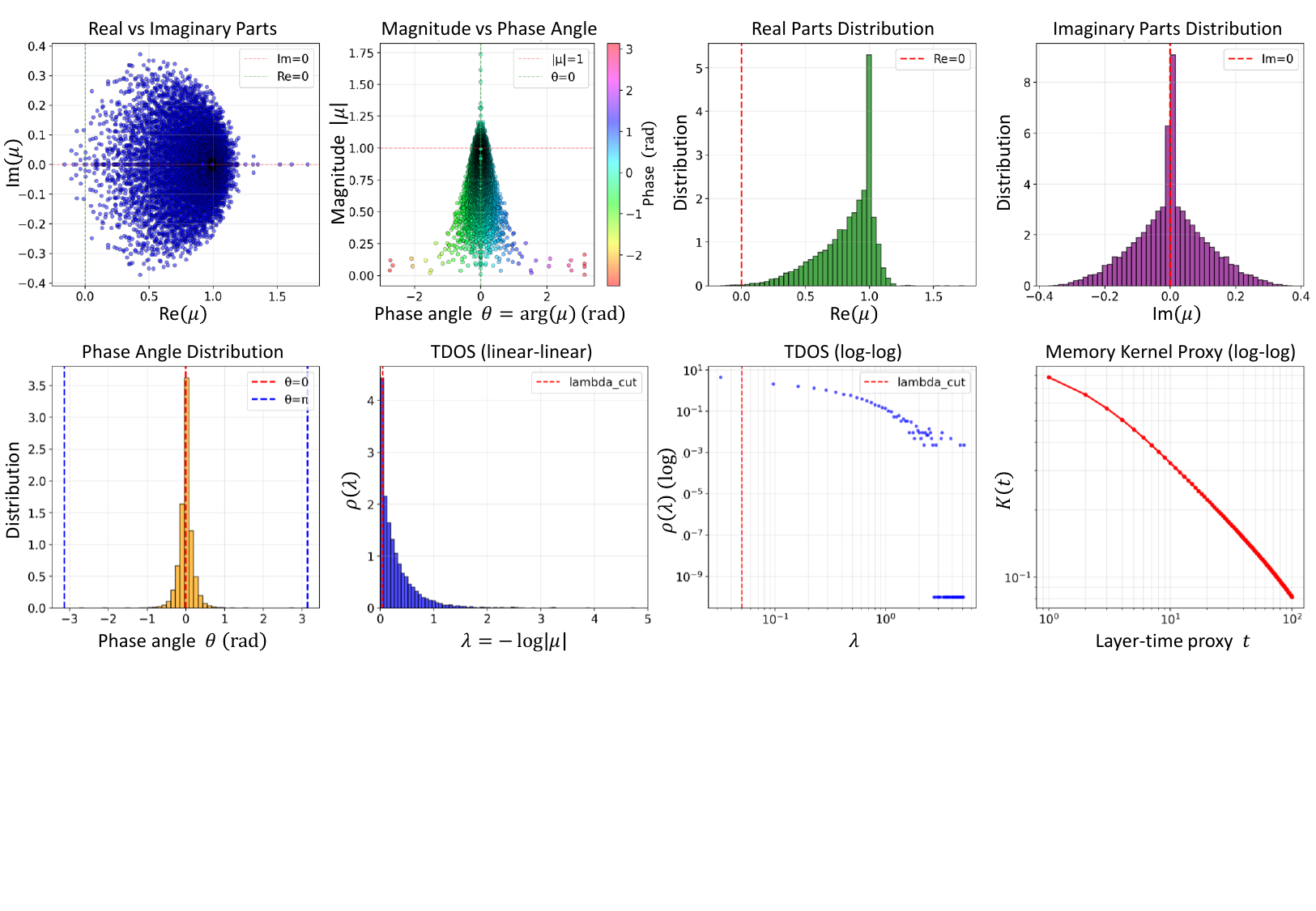}
\caption{
Spectral characterization of the Transformer Jacobian and construction of the time-scale density of states.
The Jacobian matrices extracted from Transformer hidden-layer mappings
yield complex eigenvalues whose magnitude determines the relaxation
rates,
$\lambda=-\log|\mu|$.
The resulting relaxation spectrum is used to construct the time-scale
density of states, from which the memory kernel, memory
self-energy, forgetting-gap, and other collective observables are
evaluated.
The distributions of the real part, imaginary part, and phase angle of
the complex eigenvalues are shown for completeness.
}
\label{fig:phase_portrait}
\end{figure*}

\subsection{A. Transformer models and construction of the collective Jacobian spectrum}

To investigate whether the collective observables predicted by
Cognitive Field Theory can be identified experimentally,
we first construct the complex Jacobian spectrum of the Transformer hidden dynamics.

We analyze the publicly available Pythia family
of autoregressive Transformer language models.
The Pythia project provides a unique experimental platform because all
models are trained using an identical tokenizer, optimization protocol,
training corpus, and architectural design while preserving checkpoints
throughout the entire optimization process \cite{16}.
This enables systematic measurements of the internal dynamical evolution
across both training time and model scale.

The present work analyzes three representative Pythia language models
containing 70M, 410M, and 1.4B trainable parameters.
Their principal specifications are summarized in
Table~\ref{tab:models}.
Unless otherwise stated, the measurements presented in
Figs.~1--12 are obtained from the representative Pythia-410M model.
The Jacobian matrices are computed using automatic differentiation in
PyTorch between successive hidden-layer representations.
Throughout this section we analyze the mapping between Layers~10 and~11
at the final training checkpoint, while later sections examine the
dependence on training step, network depth, and model scale.

The theoretical relation between the complex Jacobian spectrum and
the collective relaxation dynamics was established in Sec.~II.
Here we specify the convention used to extract the corresponding
spectral quantities from the numerical Transformer analysis.

For a discrete propagation map, a Jacobian eigenvalue
$\zeta_\alpha$ determines the relaxation rate and circulation
frequency according to
\[
\lambda_\alpha
=
-\frac{1}{\Delta s}\ln|\zeta_\alpha|,
\qquad
\omega_\alpha
=
\frac{1}{\Delta s}\arg\zeta_\alpha ,
\]
where $\Delta s$ denotes the discrete propagation interval introduced
in Sec.~II.

Throughout the numerical analysis, each measured propagation map is
taken as one discrete dynamical interval, $\Delta s=1$.  This
convention is applied both to a single-block map and to a composed map
spanning multiple Transformer blocks.  The measured quantities are
therefore
\begin{equation}
\lambda_\alpha
=
-\ln|\zeta_\alpha|,
\qquad
\omega_\alpha
=
\arg\zeta_\alpha .
\label{eq:numerical_relaxation_rate_unit}
\end{equation}
For a composed propagation from block $i$ to block $j$, these
quantities characterize the relaxation and circulation accumulated
over the complete measured map rather than rates normalized per
Transformer block.

The TDOS is constructed from the resulting stable relaxation modes,
$\lambda_\alpha>0$, using the definition given in Sec.~II.C.
Unless otherwise stated, all TDOS measurements below use this same
spectral convention, allowing direct comparison across training
stages, input sequences, propagation depths, and token subspaces.

To probe the local collective dynamics, we evaluate the Jacobian using
a representative input prompt, ``\emph{The future of tiny cognitive field
networks is \ldots}''.
Throughout the present work, the first eight input tokens are used to
construct the local mapping between successive hidden-layer
representations.

Since the hidden dimension of the Pythia-410M model is
$d_{\rm hidden}=1024$, the resulting Jacobian has dimension
\begin{equation}
(N_{\rm token}d_{\rm hidden})
\times
(N_{\rm token}d_{\rm hidden})
=
8192\times8192,
\end{equation}
yielding 8192 complex Jacobian eigenvalues from which the complete
relaxation spectrum is constructed.

To assess the numerical robustness of the spectral measurements,
we repeated the same analysis using sixteen-token input sequences.
The resulting TDOS and its derived infrared observables reproduce the
qualitative behavior obtained with the input length used in the main
analysis.
Representative comparisons are presented in Appendix~B, providing an
independent check that the principal infrared signatures are not
specific to a single finite input configuration.

Likewise, although the representative results shown below are obtained
from the Layer~10$\rightarrow$11 mapping, equivalent analyses were
performed for all remaining Transformer layers.
The overall infrared organization remains unchanged, while only
quantitative variations in spectral weight and peak position are
observed.

\begin{figure*}[t]
\centering
\includegraphics[width=1.0\textwidth, trim=0cm 1.7cm 0cm 0cm]{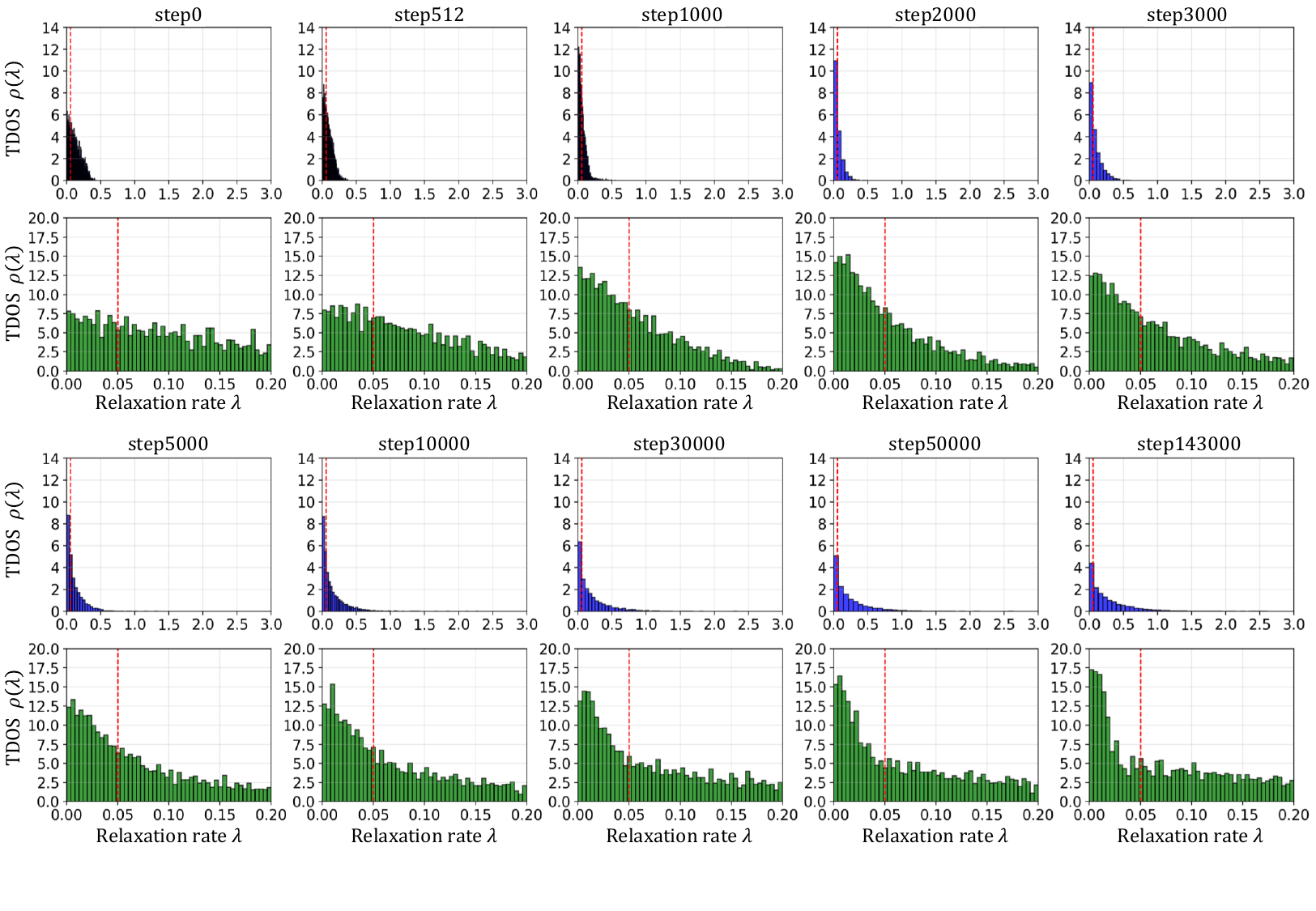}
\caption{
Evolution of the time-scale density of states during
Transformer training.
The TDOS is constructed directly from the relaxation spectrum obtained
from the Jacobian eigenvalues measured at representative training
checkpoints of the Pythia-410M model.
The upper panels show the full relaxation spectrum, while the lower
panels enlarge the infrared region below the cutoff
$\lambda_{\rm cut}$.
As optimization proceeds, relaxation modes progressively accumulate
toward smaller relaxation rates, producing a systematic infrared
reorganization of the spectrum.
Importantly, the TDOS remains broadly distributed throughout training
rather than collapsing into a single low-energy peak, indicating the
formation of a hierarchical manifold of collective slow relaxation
modes.
This measured infrared organization provides the microscopic basis for
the memory kernel, memory self-energy, cognitive forgetting gap,
collective susceptibility, and infrared critical behavior analyzed in
the following sections.
}
\label{fig:phase_portrait}
\end{figure*}

Figure~1 summarizes the complete construction of the collective
observables employed throughout this work.
The upper-left panel shows the measured complex Jacobian eigenvalues
in the complex plane.
Most eigenvalues are distributed around the positive real axis with
magnitudes smaller than unity, indicating that the hidden dynamics is
dominated primarily by stable relaxational modes rather than strongly
oscillatory dynamics.
The upper-middle panel presents the corresponding magnitude--phase
distribution, while the upper-right and lower-left panels display the
distributions of the real part, imaginary part, and phase angle of the
complex spectrum.

The measured complex Jacobian spectrum contains both relaxation and
circulation sectors.
The present work focuses on the relaxation sector, from which the relaxation-rate spectrum
is constructed, while the circulation sector will be investigated
separately.
The lower-middle panel presents the resulting TDOS.
A pronounced accumulation of modes toward the infrared region is
already visible, indicating that the Transformer contains a broad
hierarchy of slow relaxation modes rather than a single characteristic
timescale.

The lower-middle-right panel displays the same TDOS on logarithmic
axes.
Its approximately linear infrared behavior provides direct evidence
for scale-free spectral organization and forms the basis for
determining the infrared critical exponent discussed in
Sec.~III~E.

Finally, the lower-right panel presents the memory kernel calculated
directly from the measured TDOS.
Instead of exhibiting the exponential decay expected from a single
relaxation time, the kernel displays a long-time power-law behavior,
demonstrating that the experimentally measured relaxation spectrum
naturally generates scale-free memory dynamics.

Figure~1 therefore illustrates the complete experimental pipeline of
the present work.
Starting from the measured complex Jacobian spectrum, 
the relaxation sector is extracted to construct the TDOS, 
from which the memory kernel, memory self-energy, cognitive forgetting gap,
collective susceptibility, and infrared critical exponent are subsequently evaluated.

\subsection{B. Evolution of the time-scale density of states}

Having established the construction of the relaxation spectrum, we now
examine how the relaxation-rate density of states evolves during
Transformer learning.

Figure~2 presents the TDOS measured at representative training
checkpoints of the Pythia-410M model.
The upper panels display the relaxation spectrum over the full range of
relaxation rates, whereas the lower panels enlarge the deep-infrared
region below the reference cutoff $\lambda_{\rm cut}$.
This cutoff is introduced only to resolve the redistribution of
spectral weight toward progressively slower relaxation modes and does
not define a theoretical boundary of the collective slow-mode sector.

The evolution of the TDOS exhibits a clear and systematic trend.
At initialization, the relaxation spectrum is broadly distributed, and
only a relatively small fraction of modes occupies the infrared region.
As optimization proceeds, an increasing number of relaxation modes
accumulates toward smaller relaxation rates, producing a pronounced
infrared enhancement that becomes visible already during the early
stages of learning and continues throughout subsequent optimization.

Importantly, this evolution is not simply a rigid displacement of the
entire spectrum.
Instead, learning selectively redistributes spectral weight toward the
infrared sector while preserving a broad continuous distribution over
the full range of relaxation timescales.
The Transformer therefore does not develop a single dominant relaxation
time.
Rather, optimization progressively generates a hierarchy of collective
slow modes spanning multiple characteristic timescales.

The enlarged infrared panels further demonstrate that the accumulated
spectral weight remains broadly distributed.
Instead of collapsing into a discrete low-energy peak, the TDOS evolves
toward a smooth weakly infrared-singular distribution.
This behavior indicates that learning continuously reorganizes the
collective relaxation manifold rather than creating isolated memory
modes.

Another important observation is that the infrared organization appears
very early during optimization.
Although the overall spectral weight continues to evolve throughout
training, the qualitative form of the TDOS is already established after
the initial optimization stage and subsequently undergoes only gradual
refinement.
This suggests that the hierarchical organization of slow relaxation
modes is an intrinsic dynamical property of Transformer learning rather
than a late-stage consequence of optimization.

The logarithmic representation further indicates that the infrared TDOS
approaches an approximately scale-free form throughout training.
The corresponding infrared critical exponent is determined
quantitatively in Sec.~III~E, where the measured spectra are shown to
remain close to a weakly infrared-singular power law with
$\beta\simeq-0.1$.

The progressive infrared accumulation of relaxation modes constitutes
the first experimentally observed collective phenomenon investigated in
the present work.
The measured spectral reorganization provides the
microscopic foundation for all subsequent analyses.

\begin{figure}[t]
\centering
\includegraphics[scale=0.55, trim= 0cm 8.4cm 0cm 0cm]{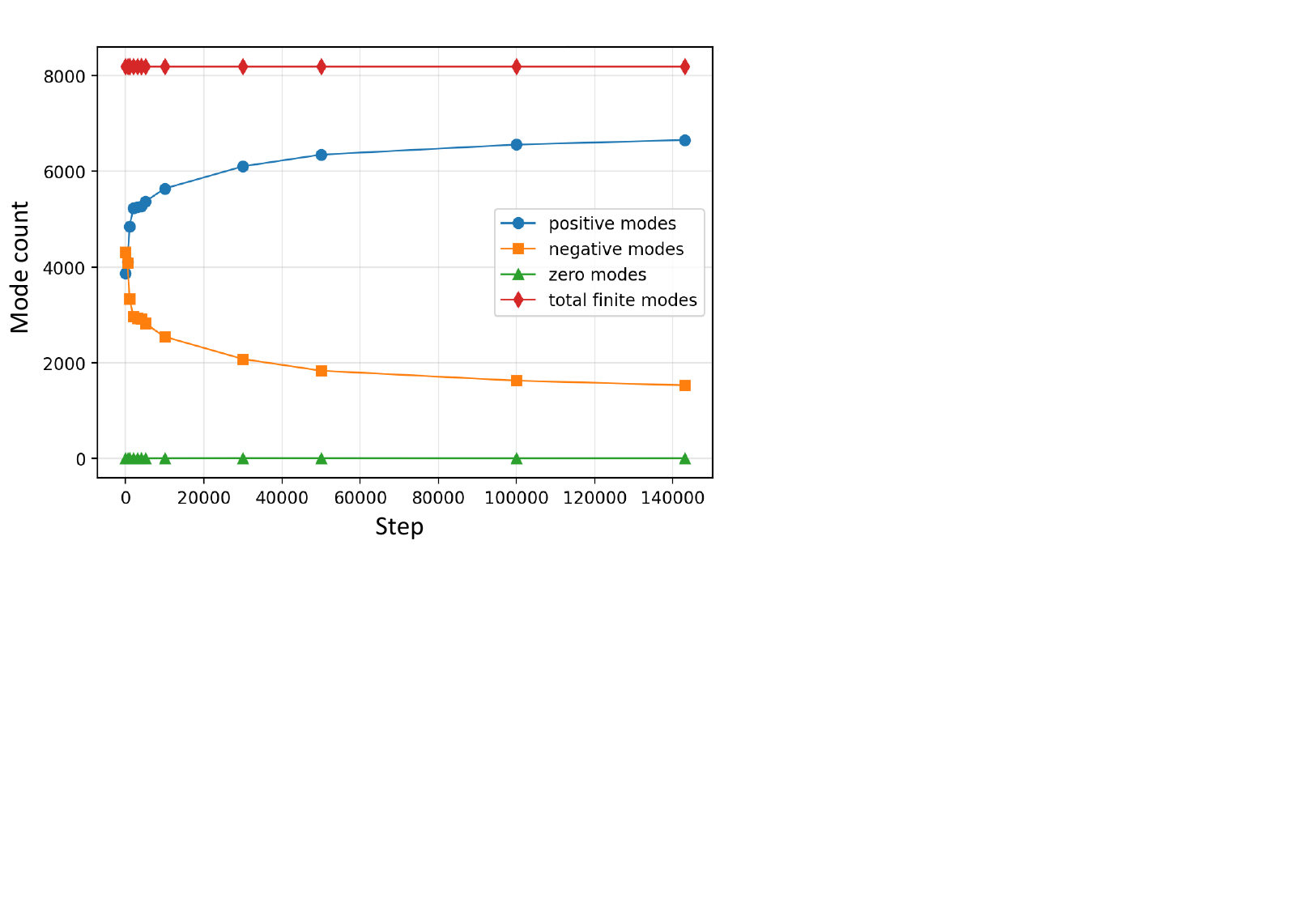}
\caption{
Evolution of the collective relaxation-mode population during
Transformer training.
The numbers of stable, unstable, zero, and total finite relaxation
modes are obtained directly from the Jacobian spectrum at
representative training checkpoints of the Pythia-410M model.
Although the total number of finite modes remains nearly constant,
training progressively redistributes the relaxation spectrum by
increasing the population of stable modes while reducing the unstable
sector.
This infrared spectral reorganization constitutes the microscopic
origin of the enhanced memory response investigated in the following
figure.
}
\label{fig:phase_portrait}
\end{figure}

\begin{figure}[t]
\centering
\includegraphics[scale=0.55, trim= 0cm 0.58cm 0cm 0cm]{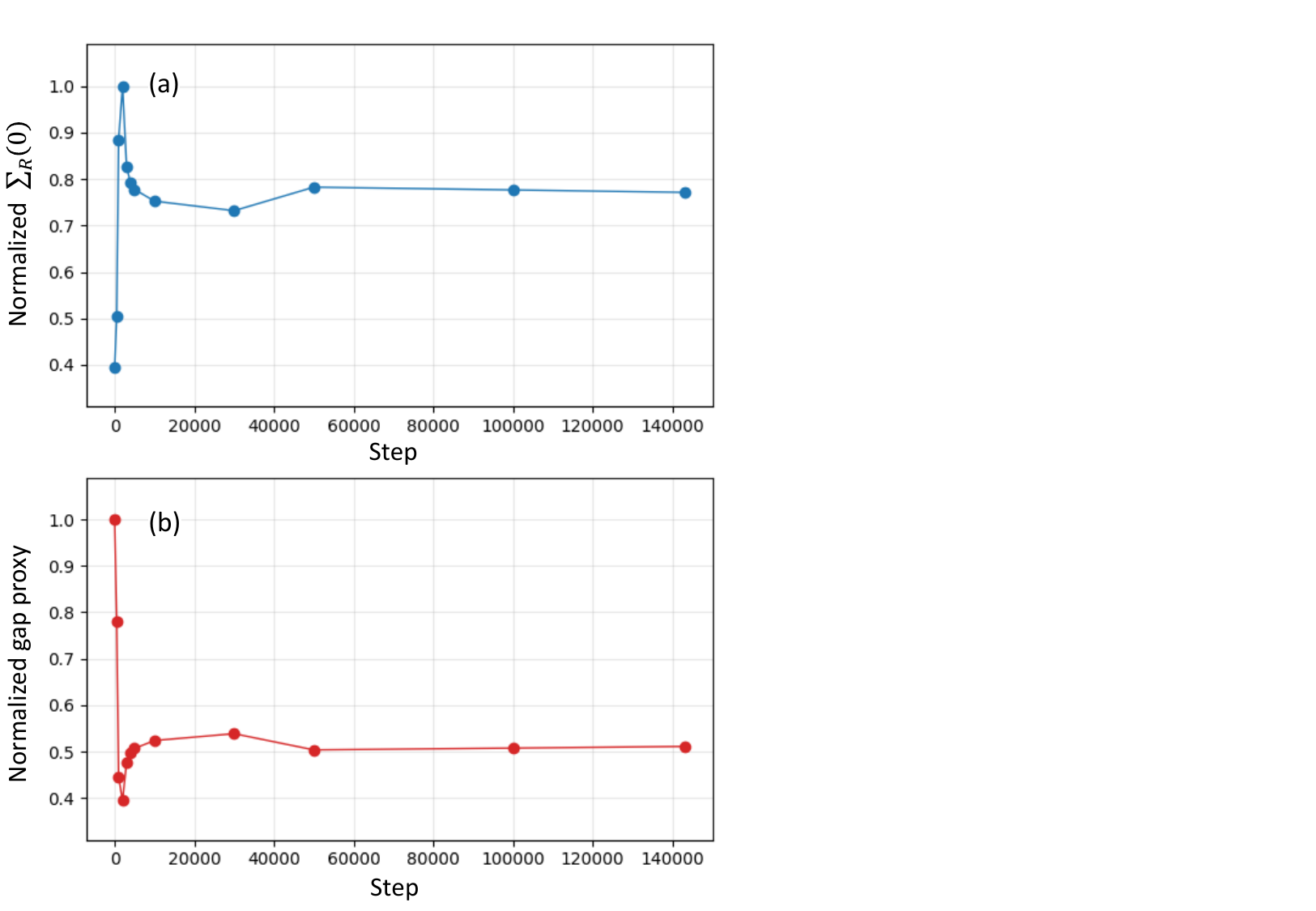}
\caption{
Critical formation of the cognitive field inferred from the measured
infrared dynamics.
(a) Normalized memory self-energy obtained from the experimentally
measured relaxation spectrum.
(b) Corresponding normalized gap proxy, defined from the inverse
memory self-energy to indicate the relative trend of forgetting-gap
suppression.
The memory self-energy exhibits a transient maximum while the gap
proxy reaches its minimum at the same optimization checkpoint,
indicating the point of strongest memory-induced suppression of the
cognitive forgetting gap and hence the largest inferred collective
susceptibility.
Subsequent optimization drives the system toward a metastable
near-critical operating regime rather than producing a monotonic
increase in memory dressing.
}
\label{fig:phase_portrait}
\end{figure}

\subsection{C. Critical formation of the cognitive field}

The infrared spectral reorganization identified in the previous
section is accompanied by a systematic redistribution of the
collective relaxation spectrum.
Figure~3 summarizes the evolution of the relaxation-mode population
throughout Transformer training.  The total number of finite
eigenmodes remains nearly constant at approximately
$8.2\times10^3$, whereas the number of stable modes increases from
approximately $3.9\times10^3$ at initialization to
$6.7\times10^3$ at the final checkpoint.  Conversely, the unstable
population decreases from approximately $4.3\times10^3$ to
$1.5\times10^3$.  Learning therefore reorganizes the existing
collective spectrum toward the stable sector rather than generating
additional dynamical degrees of freedom.

Within Cognitive Field Theory, the contribution of these stable
relaxation modes to the static memory self-energy is
\begin{equation}
\Sigma(0)
=
g^2
\int_{0}^{\Lambda}
d\lambda\,
\frac{\rho(\lambda)}{\lambda},
\label{eq:static_self_energy_continuum}
\end{equation}
under the constant-coupling approximation.
For the numerically measured discrete spectrum, we evaluate the
regularized form
\begin{equation}
\Sigma_{\epsilon}(0)
=
g^2
\sum_{\lambda_\alpha>0}
\frac{1}{\lambda_\alpha+\epsilon},
\label{eq:measured_regularized_self_energy}
\end{equation}
where the sum extends over all finite positive relaxation rates
measured at each checkpoint.

The parameter $\epsilon$ regularizes the extreme infrared contribution
of modes with relaxation rates approaching zero.  It is not imposed
as a hard spectral cutoff: modes satisfying
$0<\lambda_\alpha<\epsilon$ remain included in
Eq.~(\ref{eq:measured_regularized_self_energy}).  Instead,
\begin{equation}
\frac{1}{\lambda+\epsilon}
\simeq
\begin{cases}
1/\lambda, & \lambda\gg\epsilon,\\[4pt]
1/\epsilon, & \lambda\ll\epsilon,
\end{cases}
\label{eq:infrared_regularization}
\end{equation}
so that $\epsilon$ acts as a soft infrared regularization scale,
preventing a small number of extremely slow modes from dominating the
finite-subspace estimate of the self-energy.

The regulator was chosen with reference to the infrared resolution of
the measured TDOS.  For the eight-token measurements used in the
training analysis, the spectral distribution is reproducibly resolved
to relaxation rates of approximately $10^{-4}$.  Below this scale,
the number of modes within individual logarithmic bins becomes small,
and the measured TDOS exhibits increasing finite-sampling
fluctuations.  We therefore use
\begin{equation}
\epsilon=10^{-3}
\label{eq:self_energy_regulator}
\end{equation}
for the main analysis, providing a conservative regularization above
the lower edge of the reliably resolved infrared spectrum.

We verified the robustness of this choice by repeating the calculation
with $\epsilon=10^{-4}$ and $10^{-5}$.  The qualitative form of the
self-energy evolution remains robust down to $\epsilon=10^{-4}$.
At $\epsilon=10^{-5}$, the result becomes increasingly sensitive to
sparsely sampled extreme-infrared modes and exhibits substantially
larger fluctuations.

This behavior does not indicate an intrinsic infrared cutoff of the
TDOS.  As shown by the sequence-length analysis in Sec.~IV, increasing
the measured token subspace progressively stabilizes the TDOS at
smaller relaxation rates.  The scale near $10^{-4}$ therefore
characterizes the practical infrared resolution of the present
eight-token measurement rather than an intrinsic lower relaxation
rate of the Transformer dynamics.

Because neither the absolute coupling strength $g$ nor the bare
forgetting rate $r$ is independently determined by the Jacobian
spectrum, we do not assign an absolute experimental value to
\begin{equation}
r_{\rm cog}=r-\Sigma(0).
\end{equation}
Instead, the measured self-energy is normalized by its maximum over
training,
\begin{equation}
\widetilde{\Sigma}_{\epsilon}(0;t)
=
\frac{
\Sigma_{\epsilon}(0;t)
}{
\max_{t'}
\Sigma_{\epsilon}(0;t')
}.
\label{eq:normalized_self_energy}
\end{equation}
To visualize the corresponding relative trend of forgetting-gap
suppression, we introduce the monotonic gap proxy
\begin{equation}
r_{\rm gap}^{\rm proxy}(t)
=
\frac{1}{\Sigma_{\epsilon}(0;t)}.
\label{eq:forgetting_gap_proxy}
\end{equation}
This quantity is not identified with the absolute cognitive
forgetting gap $r_{\rm cog}$; it is used only as a relative indicator
whose decrease corresponds to increasing memory self-energy and hence
stronger suppression of the forgetting gap.  For visualization, the
gap proxy is normalized by its maximum over the measured checkpoints.

The resulting evolution is summarized in Fig.~4.
Unlike the monotonic increase of the stable-mode population, the
memory self-energy exhibits a strongly non-monotonic training
trajectory.  It increases rapidly during the earliest stage of
optimization, reaches a pronounced maximum within approximately
$2\times10^3$ optimization steps, and subsequently decreases before
approaching an approximately stationary late-training value.
The inverse-self-energy proxy evolves oppositely, reaching its minimum
near the same checkpoint at which the self-energy is maximal.

Within Cognitive Field Theory, increasing $\Sigma(0)$ suppresses the
renormalized forgetting gap,
\[
r_{\rm cog}
=
r-\Sigma(0),
\]
and enhances the static collective susceptibility,
\begin{equation}
\chi(0)
\propto
\frac{1}{r_{\rm cog}}.
\end{equation}
The transient maximum of the measured self-energy therefore identifies
the training stage at which the system approaches most closely the
critical condition
\begin{equation}
r_{\rm cog}\rightarrow0^+,
\end{equation}
within the relative spectral measure accessible to the present
experiment.

After this transient maximum, the self-energy decreases moderately
and approaches a nearly stationary value despite the continued
increase in the number of stable relaxation modes.  The result shows
that the collective memory response is controlled not simply by the
number of stable modes, but by their distribution over relaxation
time scales.  Training therefore first produces a strong infrared
enhancement and subsequently reorganizes the spectrum into a
persistent late-training dynamical regime.

Within the interpretation of Cognitive Field Theory, this evolution
is consistent with a transient approach toward critical collective
field formation followed by stabilization in a finite-gap,
near-critical regime.  Importantly, the experimentally measured
quantity is the regularized spectral self-energy
$\Sigma_\epsilon(0)$; the absolute value of $r_{\rm cog}$ is not
directly determined without an independent estimate of the bare
forgetting rate.

Although the primary focus of the present work is the collective
infrared dynamics, the same training interval is also accompanied by
a marked change in language generation.
Table~\ref{tab:output_evolution} presents representative outputs
generated from an identical prompt at successive checkpoints.
Before the self-energy maximum, the model produces either random
tokens or short-range grammatical fragments, whereas coherent
semantic organization becomes apparent around the early-training
regime in which the collective spectral response is strongly
enhanced.  Subsequent optimization primarily improves the quality and
consistency of the generated text.

These examples are qualitative and are not used as quantitative
evidence for the spectral analysis.  They are included only to
illustrate the behavioral evolution occurring over the same training
period.  All quantitative conclusions concerning the collective
dynamics are based on the measured Jacobian relaxation spectrum and
the corresponding regularized self-energy.

Since all analyzed Pythia checkpoints are publicly available and widely
used throughout the machine-learning community \cite{22,23,24}, the representative
outputs serve only as illustrative examples rather than quantitative
experimental observables.

\begin{table*}[t]
\caption{
Representative language generation during the critical formation of the
cognitive field.
}
\label{tab:output_evolution}
\centering
\begin{tabular}{lp{14cm}}
\hline\hline
Checkpoint & Representative output \\
\hline

step0 &
``The future of tiny cognitive field networks is Neg vegetables Hunting
...'' (random token sequence with no grammatical or semantic
organization). \\

step512 &
``The future of tiny cognitive field networks is a very low-term value
of the first-up of the first-up ...'' (grammatical fragments begin to
appear, but repetitive loops dominate and semantic coherence is still
absent). \\

step1000 &
``The future of tiny cognitive field networks is a very important part
of the development of the brain.'' (a meaningful sentence appears,
although the output rapidly falls into repetitive loops). \\

step2000 &
``The future of tiny cognitive field networks is a key element in the
evolution of the brain.'' (local semantic coherence improves, but
long-range organization remains weak). \\

step3000 &
``The future of tiny cognitive field networks is a subject of great
interest to the scientific community.'' (stable scientific language
style emerges with coherent sentence-level organization). \\

step5000 &
``The future of tiny cognitive field networks is a question of how to
design and implement a new generation of computational models.'' (the
model begins to generate coherent technical explanations extending over
multiple clauses). \\

step10000 &
``The future of tiny cognitive field networks is a question of great
interest to researchers.'' (long-range contextual organization becomes
well established and coherent scientific discourse emerges). \\

step30000 &
``The future of tiny cognitive field networks is a topic of intense
research. In this paper, we propose a novel method ...'' (paragraph-level
scientific writing becomes possible). \\

step50000 &
``The future of tiny cognitive field networks is still in the early
stages. The authors would like to thank the anonymous referees ...''
(document-level formatting and academic writing style are reproduced). \\

step100000 &
``The future of tiny cognitive field networks is in doubt.'' (language
generation remains grammatically stable although diversity decreases due
to repetitive decoding). \\

step143000 &
``The future of tiny cognitive field networks is in the hands of the
community. The authors thank the anonymous reviewers ...'' (stable
high-level scientific writing with consistent document structure). \\

\hline\hline
\end{tabular}
\end{table*}

\subsection{D. Evolution of the memory kernel}

The relaxation spectrum measured at each training stage directly
determines the corresponding memory dynamics of the Transformer.
Since the memory kernel is obtained from the measured relaxation
rates, its evolution provides a dynamical representation of the
learning-induced reorganization of the TDOS.

Rather than assuming a phenomenological functional form, we compute
the memory kernel directly from the measured relaxation spectrum.
For the experimentally obtained relaxation rates
$\{\lambda_\alpha\}$,
\begin{equation}
K(t)
=
\frac{1}{N}
\sum_{\alpha=1}^{N}
e^{-\lambda_\alpha t},
\label{eq:kernel_discrete}
\end{equation}
where $N$ denotes the number of measured relaxation modes.
Introducing the normalized TDOS,
\begin{equation}
\rho(\lambda)
=
\frac{1}{N}
\sum_{\alpha=1}^{N}
\delta(\lambda-\lambda_\alpha),
\end{equation}
this expression becomes
\begin{equation}
K(t)
=
\int_0^\infty
d\lambda\,
\rho(\lambda)e^{-\lambda t}.
\label{eq:kernel_continuum}
\end{equation}
The temporal form of the memory kernel is therefore determined
directly by the measured distribution of collective relaxation rates.

Figure~5 compares the TDOS measured at representative training
checkpoints with the corresponding memory kernels calculated from the
same spectra.  Figure~5(a) shows the TDOS on logarithmic scales, while
Fig.~5(b) presents the resulting memory kernels as a function of the
effective layer time.  Learning substantially reorganizes the
relaxation spectrum and consequently changes the magnitude and
long-time weight of the memory response.

The scaling analysis concerns specifically the long-time regime
associated with the infrared sector of the TDOS.  As derived in
Sec.~II, an infrared spectrum
\[
\rho(\lambda)
\sim
C\lambda^\beta,
\qquad
\lambda\rightarrow0^+,
\]
produces the asymptotic memory law
\begin{equation}
K(t)
\sim
A\Gamma(1+\beta)t^{-(1+\beta)},
\qquad
t\rightarrow\infty.
\label{eq:kernel_ir_scaling}
\end{equation}
In particular, the approximately flat infrared TDOS observed in the
present measurements,
\(
\beta\simeq0,
\)
corresponds to the marginal scaling
\(
K(t)\sim t^{-1}.
\label{eq:kernel_inverse_time}
\)

Consistent with this relation, the measured kernels exhibit an
extended approximately linear regime on log--log scales.  The
power-law exponent is determined from this scaling interval rather
than from the complete temporal profile.  Short and intermediate
times receive contributions from faster relaxation modes and therefore
need not obey the infrared asymptotic law, while the longest accessible
times are increasingly sensitive to the finite spectral resolution of
the smallest measured relaxation rates.  The linear regime between
these limits provides the experimentally resolved scaling window in
which the infrared TDOS controls the memory dynamics.

No power-law form is imposed in the calculation of $K(t)$.
The observed scaling follows directly from the measured relaxation
eigenvalues through Eq.~(\ref{eq:kernel_discrete}).  Its persistence
across training stages is therefore consistent with the observation
that learning substantially redistributes spectral weight while
preserving the approximately flat infrared spectral class.

The memory kernel nevertheless changes quantitatively during
optimization.  The strongest reorganization occurs during the early
training regime, consistent with the non-monotonic evolution of the
memory self-energy discussed in Sec.~III~C.  At later checkpoints,
the kernel approaches a comparatively stable form, indicating that
the long-time collective memory dynamics becomes increasingly
stationary despite continued optimization.

The measured kernels therefore reveal two complementary aspects of
the learning dynamics.  Their amplitudes and detailed temporal
profiles reflect the learning-induced redistribution of relaxation
modes, whereas the approximately invariant long-time scaling exponent
reflects the persistent infrared organization of the TDOS.  The
observed relation between a nearly flat infrared relaxation spectrum
and an approximately $1/t$ memory kernel provides a direct dynamical
connection between the measured TDOS and the scale-free memory
predicted by Cognitive Field Theory.

\begin{figure}[t]
\centering
\includegraphics[scale=0.5, trim= 0.3cm 0.6cm 0cm 0cm]{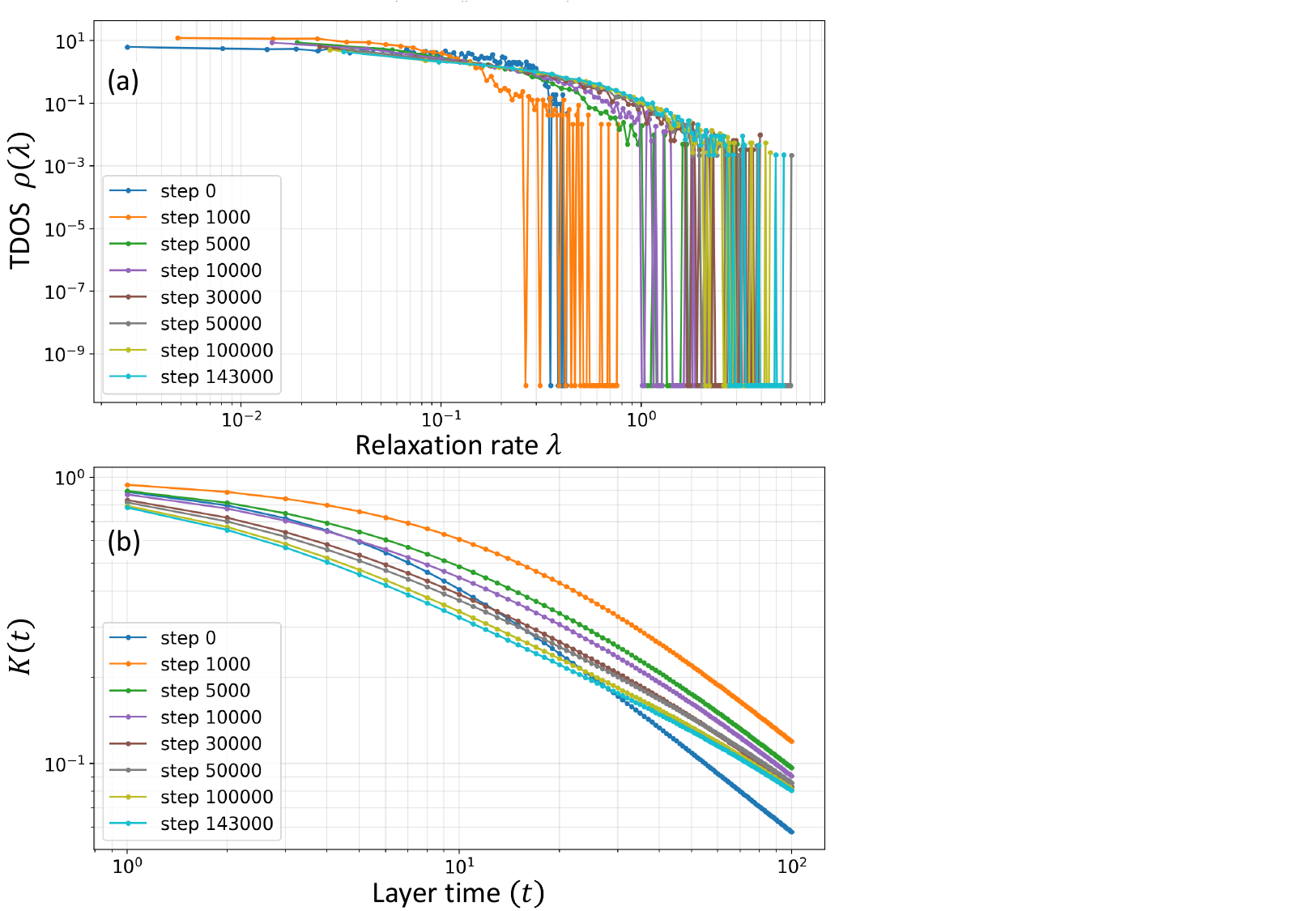}
\caption{
Evolution of the memory kernel during Transformer training.
(a) Time-scale density of states measured at representative
training checkpoints of the Pythia-410M model, shown on logarithmic
scales.
As optimization proceeds, the infrared spectral weight progressively
increases through the accumulation of slow relaxation modes.
(b) Memory kernels computed directly from the measured TDOS.
The enhancement of the infrared spectrum systematically amplifies the
long-time memory response while preserving an approximately power-law
temporal decay.
Rather than generating a single characteristic memory timescale,
Transformer learning progressively strengthens a hierarchical
scale-free memory kernel through infrared spectral organization.
}
\label{fig:phase_portrait}
\end{figure}

\begin{figure*}[t]
\centering
\includegraphics[width=1.0\textwidth, trim=0cm 0.5cm 0cm 0cm]{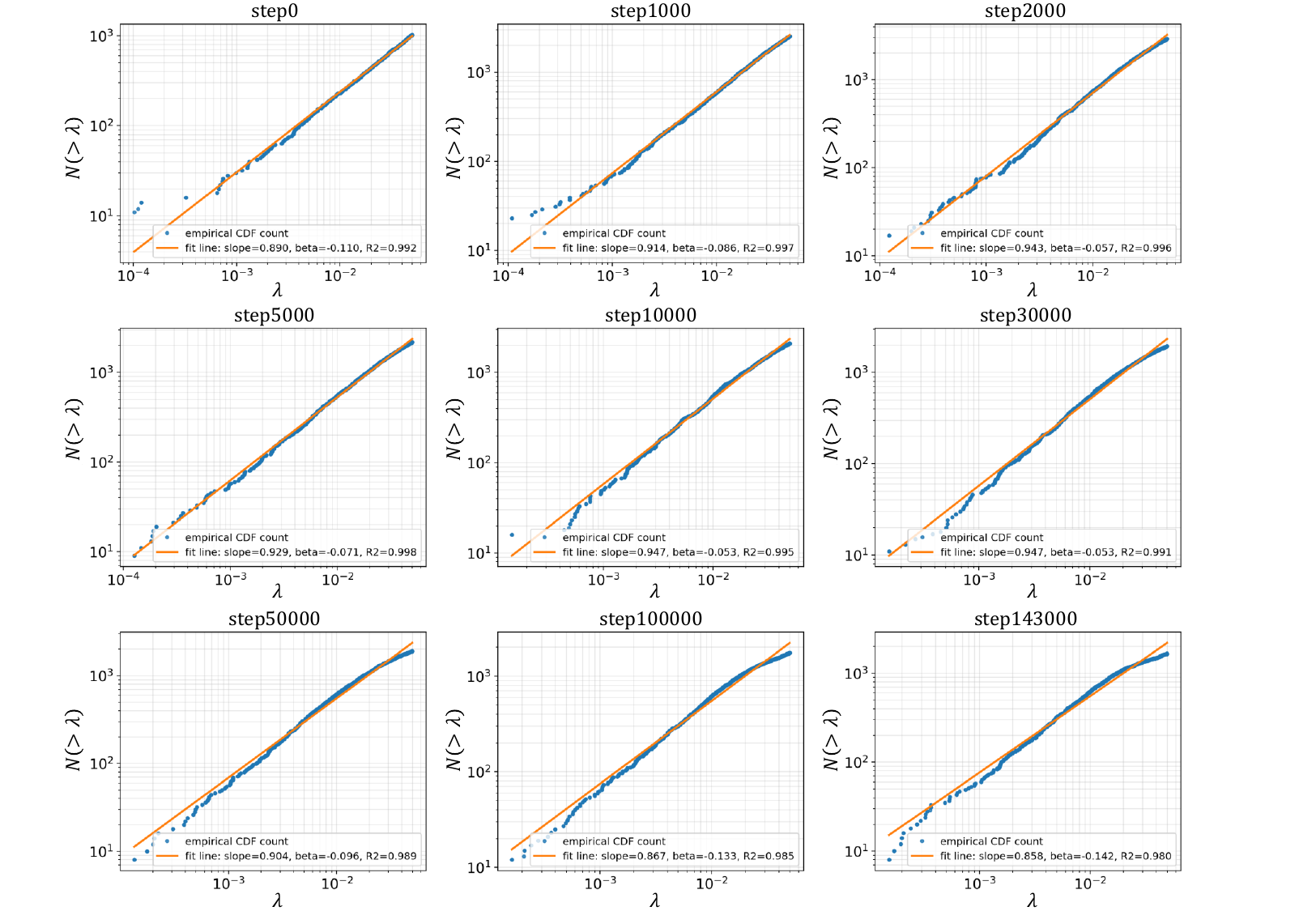}
\caption{
Infrared critical scaling of the experimentally measured time-scale
density of states.
The cumulative mode distributions
$N(<\lambda)$
are fitted by the scaling relation
$N(<\lambda)\propto\lambda^{\beta+1}$
at representative training checkpoints.
Excellent linearity is observed throughout the infrared fitting region,
with
$R^2>0.99$
for nearly all checkpoints.
Although the TDOS undergoes substantial infrared spectral
reorganization during learning, the extracted exponent remains close to
$\beta\simeq-0.1$,
indicating that optimization preserves the infrared universality class
while enhancing the population of collective slow relaxation modes.
}
\label{fig:phase_portrait}
\end{figure*}

\subsection{E. Infrared critical exponent of the TDOS}

The infrared spectral reorganization described above raises a
fundamental question: does Transformer learning alter the infrared
scaling class of the collective dynamics, or does optimization
redistribute spectral weight while preserving its asymptotic scaling
structure?

To address this question, we determine the infrared exponent directly
from the experimentally measured relaxation spectrum.
Rather than fitting histogram representations, which depend on the
choice of bin width, we analyze the empirical cumulative mode
distribution.
For an infrared TDOS of the form
\[
\rho(\lambda)
\sim
C\lambda^\beta,
\]
the corresponding cumulative mode count satisfies
\begin{equation}
N(<\lambda)
=
\sum_\alpha
\Theta(\lambda-\lambda_\alpha)
\propto
\lambda^{\beta+1}.
\label{eq:cumulative_tdos_scaling}
\end{equation}
The exponent can therefore be extracted directly from the slope of the
cumulative distribution in logarithmic coordinates.

For each training checkpoint, the cumulative distribution is fitted
over the resolved infrared interval
\begin{equation}
10^{-4}
<
\lambda
<
5\times10^{-2},
\end{equation}
within which it remains approximately linear on log--log scales.
Ordinary least-squares regression yields coefficients of determination
\(R^2>0.99\) for nearly all checkpoints, indicating excellent agreement
with power-law scaling over the measured infrared window.

Figure~6 presents representative cumulative distributions and the
corresponding fits throughout training.
Despite substantial reorganization of the complete relaxation
spectrum, the cumulative distributions preserve nearly the same
infrared slope.
The extracted effective infrared exponent remains within the narrow
range
\begin{equation}
-0.14
\lesssim
\beta
\lesssim
-0.05,
\end{equation}
with a representative value
\(
\beta
\simeq
-0.1.
\)

The measured TDOS therefore exhibits a nearly flat but systematically
infrared-enhanced scaling over the resolved infrared window.
The weakly negative exponent indicates that the density of relaxation
modes increases toward smaller relaxation rates, consistent with the
infrared accumulation observed directly in the measured spectra.
More importantly, the limited variation of the exponent throughout
training shows that substantial redistribution of spectral weight
occurs without a corresponding change in the resolved infrared
scaling law.
The measurements are therefore consistent with optimization
reorganizing the collective relaxation spectrum while preserving its
infrared universality class.

This interpretation is independently supported by the memory kernels
shown in Fig.~5.
For an infrared spectrum
\(
\rho(\lambda)\sim\lambda^\beta,
\)
the corresponding long-time kernel scales as
\begin{equation}
K(t)
\sim
t^{-(1+\beta)}.
\end{equation}
The measured value
\(
\beta\simeq-0.1
\)
therefore predicts a long-time decay close to, but slightly slower
than, the \(1/t\) behavior associated with a perfectly flat TDOS,
consistent with the approximately \(1/t\) scaling observed directly
from the measured relaxation spectra.
The persistence of this nearly flat, weakly infrared-enhanced TDOS
together with the corresponding scale-free long-memory kernel provides
quantitative support for the infrared fixed-point organization
predicted by Cognitive Field Theory.

\begin{figure*}[t]
\centering
\includegraphics[width=1.0\textwidth, trim=0cm 6cm 0cm 0cm]{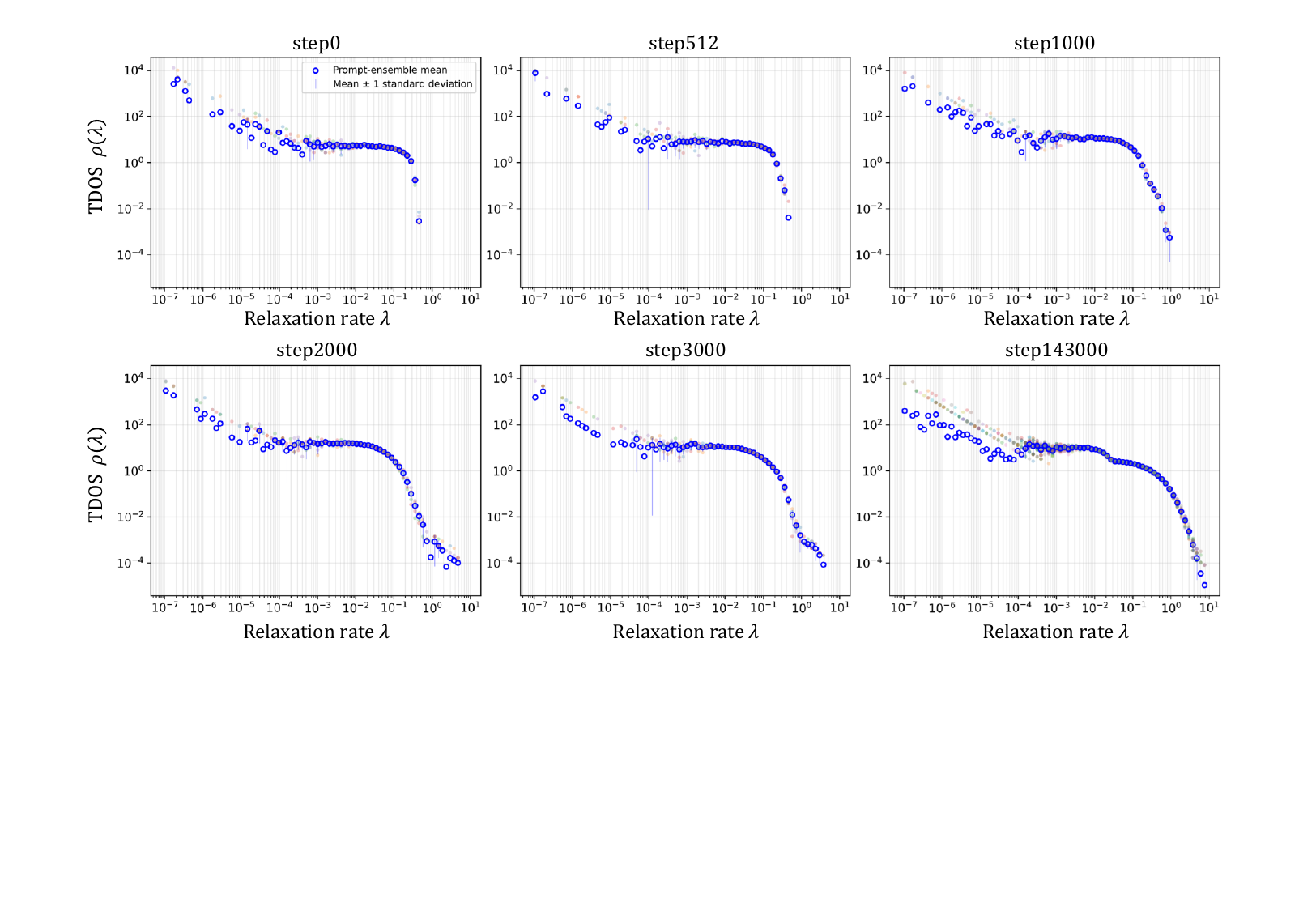}
\caption{
Evolution of the time-scale density of states during Pythia-410M
training.
The TDOS is measured using the same eight-token state-space
representation at Steps~0, 512, 1000, 2000, 3000, and 143000.
Steps~0--3000 are evaluated using five distinct input prompts,
while the fully trained model at Step~143000 is evaluated using
30 prompts.
Open circles show the prompt-ensemble mean and vertical bars indicate
one standard deviation.
A broad relaxation spectrum is already present before training and
remains well defined throughout optimization, while its spectral
distribution is systematically reorganized during learning.
}
\label{fig:tdos_training_stages}
\end{figure*}

%=============================================

\section{IV. Macroscopic Universality of the Relaxation Spectrum}

The preceding sections showed that Transformer learning reorganizes
the collective relaxation spectrum and produces a nontrivial evolution
of the infrared memory self-energy.  We now examine whether the
time-scale density of states constitutes a reproducible
macroscopic dynamical quantity of the Transformer.

Because each Jacobian is evaluated at an input-dependent hidden state
and within a finite token subspace, this question requires two
complementary tests.  First, the measured TDOS should remain
reproducible across different hidden-state realizations generated by
different prompts.  Second, measurements performed in progressively
larger token subspaces should converge toward a common spectral
distribution.  We examine these two properties below, while the
asymptotic infrared scaling of the TDOS is treated separately through
the fixed-point analysis of Sec.~II.D.

\subsection{A. TDOS formation, spectral reorganization, and prompt robustness}

Figure~\ref{fig:tdos_training_stages} shows the prompt-ensemble TDOS
measured at representative stages of Pythia-410M training for the
Layer~10$\rightarrow$11 propagation, using the same eight-token
state-space representation.  For Steps~0--3000, the distributions are
obtained from five distinct input prompts, while the fully trained
model at Step~143000 is evaluated using an ensemble of 30 prompts.
Open circles denote the prompt-ensemble mean and vertical bars
indicate one standard deviation.

A broad relaxation-rate distribution is already present in the
untrained model at Step~0 and remains well defined throughout
optimization.  The TDOS therefore does not emerge only after learning.
Rather, the Transformer architecture already supports a collective
spectrum of relaxation states, while learning reorganizes the
distribution of spectral weight across relaxation time scales.
Substantial changes in the detailed spectral profile occur during
training, even though a broad slow-relaxation sector persists
throughout the entire training trajectory.

This observation distinguishes the existence of the TDOS from its
learning-induced reorganization.  The TDOS characterizes the spectrum
of collective relaxation states supported by the dynamical
architecture, whereas optimization changes their distribution and
collective participation.  The formation and evolution of the
macroscopic collective response therefore involve the reorganization
of an existing dynamical spectrum rather than the creation of the
TDOS itself.

A physically meaningful TDOS should furthermore be reproducible when
the same collective spectrum is measured from different local
hidden-state configurations.  Different prompts generate distinct
hidden-state trajectories and therefore generally produce different
local Jacobians and eigensystems.  Nevertheless, if the TDOS
represents a macroscopic density of relaxation states, these
input-dependent measurements should recover the same coarse-grained
distribution in relaxation-rate space.

\begin{figure}[t]
\centering
\includegraphics[width=0.57\textwidth, trim=0cm 3.2cm 0cm 0cm]{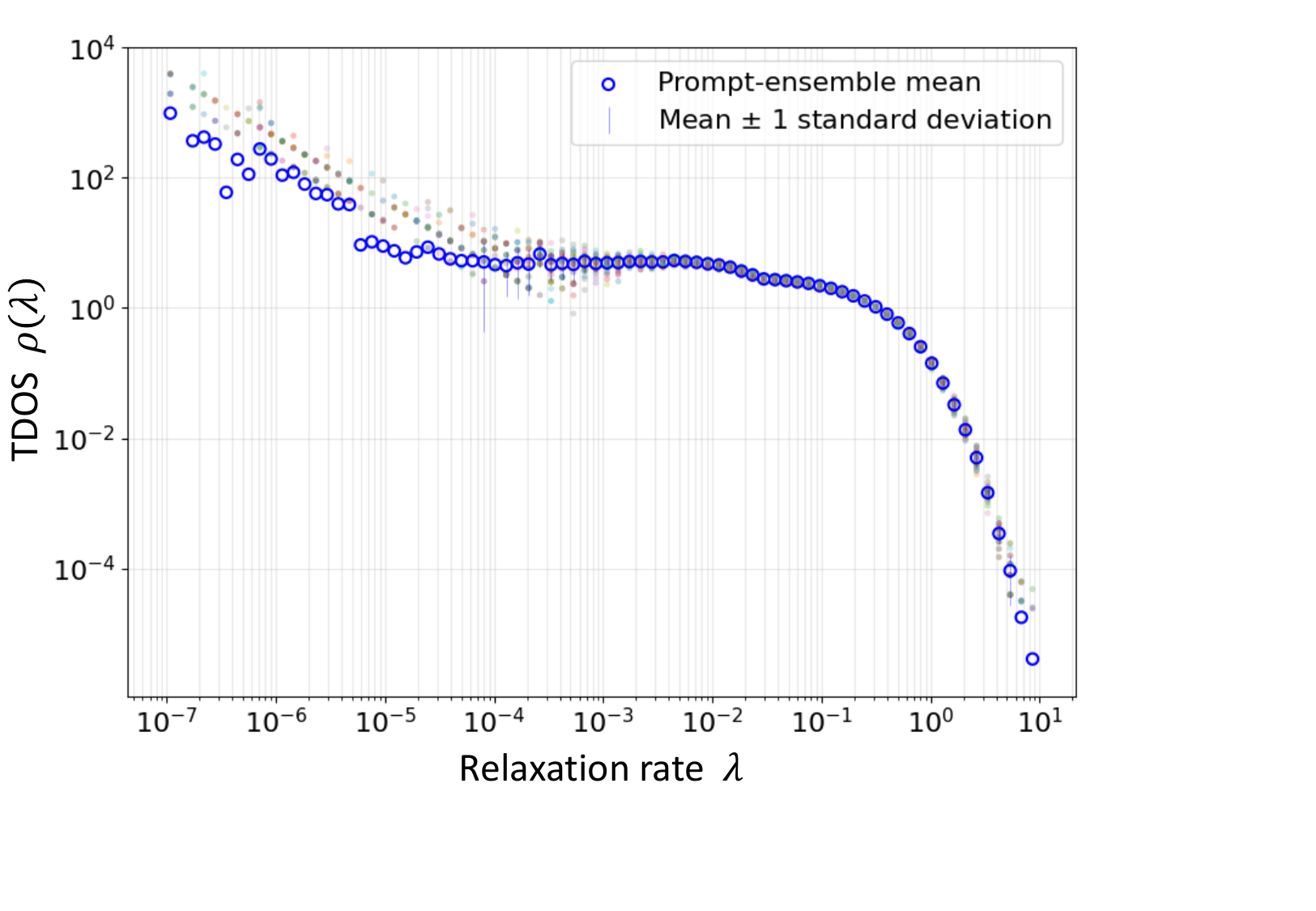}
\caption{
Prompt-ensemble concentration of the normalized TDOS.
Colored symbols show the independently measured spectra obtained from
different prompts, while open blue circles denote the corresponding
ensemble mean.
Vertical bars indicate one standard deviation across prompts.
The independently measured spectra rapidly concentrate around a common
relaxation-rate distribution over nearly the entire measured range.
Residual variations are largest only in the deepest infrared bins,
where the number of sampled slow modes is smallest and statistical
sampling is therefore limited.
}
\label{fig:prompt_ensemble_self_averaging}
\end{figure}

Figure~\ref{fig:prompt_ensemble_self_averaging} tests this property in
the fully trained model using a 24-token retained input subspace.
The TDOS is measured independently for thirty representative prompts
and compared with their prompt-ensemble mean.  Despite the different
input-dependent hidden states, the measured spectra concentrate around
a common distribution over most of the resolved relaxation-rate
range.  Prompt-to-prompt variations become appreciable primarily in
the sparsely sampled extreme-infrared bins.

The individual eigenvalues and eigenvectors therefore remain
state dependent, while their coarse-grained distribution in
relaxation-rate space becomes largely prompt independent.  This
spectral concentration can be expressed as
\begin{equation}
{\rm Var}_{h}
\left[
\widehat{\rho}_{\mathcal S,h}(\lambda)
\right]
\rightarrow 0,
\label{eq:prompt_tdos_self_averaging}
\end{equation}
over the well-resolved spectral range as sufficiently many collective
modes are sampled.  The remaining variance in the deepest infrared
bins is consistent with the limited number of modes available in
those bins.

The prompt ensemble thus provides independent local measurements of
a common macroscopic relaxation spectrum.  The TDOS is therefore not
a spectral fingerprint of a particular input-dependent hidden state,
but a reproducible collective property of the Transformer dynamics.
Individual prompt-resolved spectra and enlarged infrared views are
provided in Appendix~C.

\subsection{B. Token-subspace convergence and infrared extension}

Prompt robustness establishes that different hidden-state
realizations recover the same coarse-grained relaxation spectrum at a
fixed observation scale.  A stronger test is obtained by changing the
dimension of the observed state space itself.

For the Pythia-410M model, increasing the retained sequence length
from 8 to 16, 24, and 28 tokens enlarges the local Jacobian from
approximately eight thousand to nearly thirty thousand hidden-state
degrees of freedom.  These measurements therefore sample
progressively larger finite-dimensional sections of the Transformer
state space.

\begin{figure}[t]
\centering
\includegraphics[width=0.57\textwidth, trim=0cm 3.2cm 0cm 0cm]{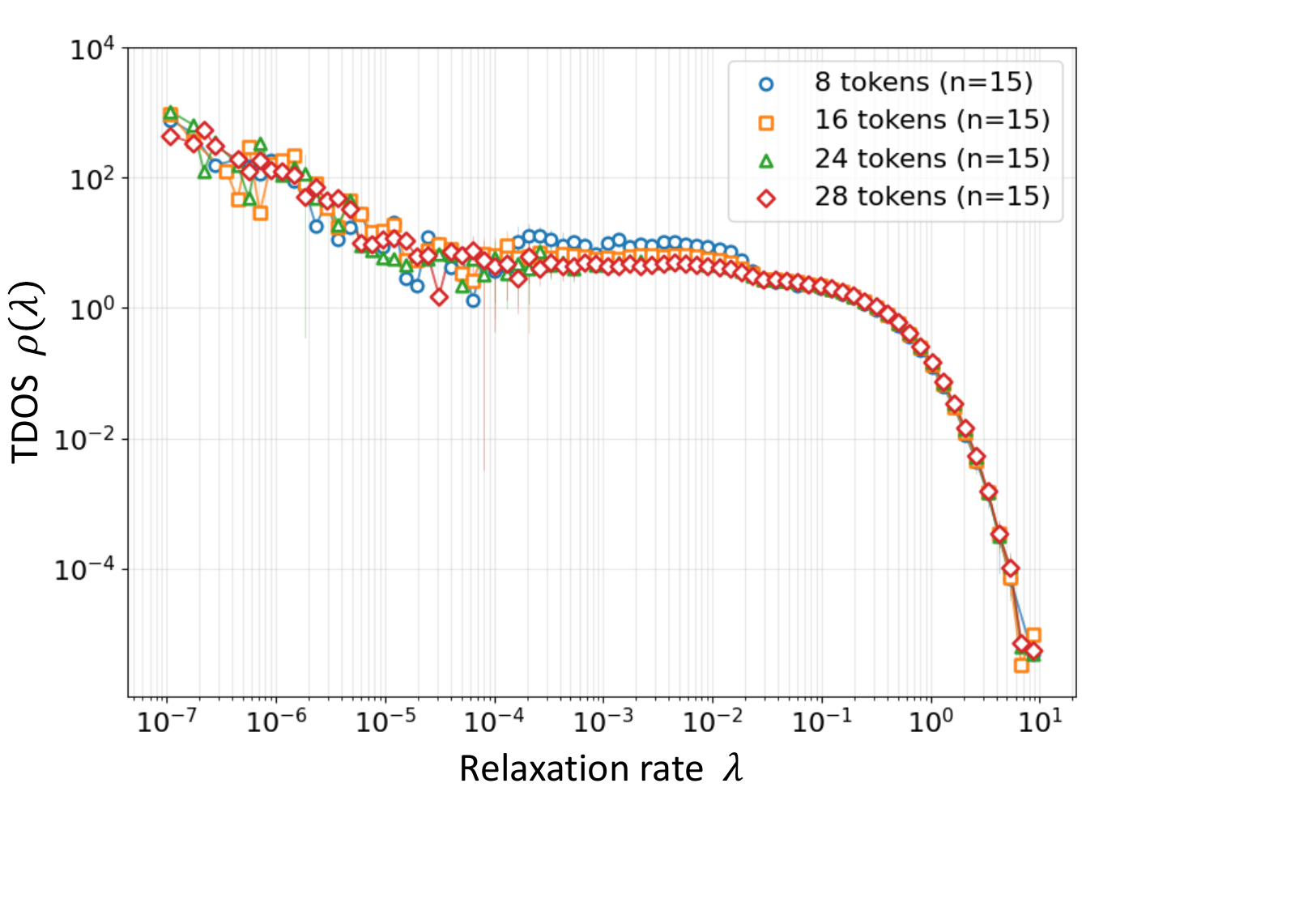}
\caption{
Convergence of the prompt-ensemble TDOS with increasing token-subspace
dimension.
The normalized TDOS is reconstructed from representative prompt
ensembles using 8-, 16-, 24-, and 28-token input subspaces.
All measurements employ the same logarithmic binning of relaxation
rates.
Despite the substantial increase in Jacobian dimension and the
corresponding change in the local observation scale, the measured
spectra preserve the same large-scale organization, including the
enhanced infrared sector and the rapidly suppressed large-\(\lambda\)
tail.
The spectra obtained from the 24- and 28-token subspaces are nearly
indistinguishable, indicating that enlarging the local observation no
longer reveals additional infrared structure.
}
\label{fig:token_length_fixed_point}
\end{figure}

Figure~\ref{fig:token_length_fixed_point} compares the corresponding
prompt-ensemble TDOS using common logarithmic bins.  Despite the
nearly fourfold increase in Jacobian dimension, the spectra preserve
the same large-scale relaxation structure and progressively converge
as the observed token subspace is enlarged.  In particular, the
24- and 28-token spectra are nearly indistinguishable over their
common resolved range.

This convergence reveals an important finite-subspace effect.
Enlarging the observed token subspace does not generate a
qualitatively different relaxation spectrum.  Instead, it increases
the number of sampled collective modes and progressively extends the
resolved TDOS toward smaller relaxation rates.  The apparent
low-$\lambda$ boundary of a finite-subspace measurement should
therefore not be interpreted as an intrinsic infrared cutoff of the
Transformer dynamics.

As the token length increases from 8 to 16, 24, and 28, progressively
slower modes become statistically resolved, while the spectral
distribution at relaxation rates already well sampled by the smaller
subspaces remains essentially unchanged.  The sequence-length
dependence is therefore primarily a progressive extension of the
observable infrared range rather than a displacement of the
underlying TDOS.

This behavior is particularly important in the infrared sector.
Since the longest relaxation time accessible to the measurement is
related to the smallest resolved relaxation rate,
\begin{equation}
\tau_{\rm max}^{\rm obs}
\sim
\frac{1}{\lambda_{\rm min}^{\rm obs}},
\end{equation}
the systematic decrease of
$\lambda_{\rm min}^{\rm obs}$ with increasing token-subspace dimension
implies that larger state-space measurements resolve progressively
longer collective time scales.  The extreme-infrared fluctuations
observed in smaller token subspaces are therefore naturally associated
with finite spectral sampling, whereas the progressively resolved
low-$\lambda$ sector provides access to the long-time limit of the
macroscopic relaxation spectrum.

To quantify the convergence, we compare the ensemble-averaged TDOS
in logarithmic coordinates.  For token lengths $T_a$ and $T_b$, the
log-space correlation is defined as
\begin{equation}
C_{\log}(T_a,T_b)
=
{\rm Corr}
\left[
\log_{10}\widehat{\rho}_{T_a}(\lambda_i),
\log_{10}\widehat{\rho}_{T_b}(\lambda_i)
\right],
\label{eq:log_tdos_correlation}
\end{equation}
and the corresponding logarithmic root-mean-square distance is
\begin{equation}
\Delta_{\log}(T_a,T_b)
=
\left[
\frac{1}{N_{\rm bin}}
\sum_{i=1}^{N_{\rm bin}}
\left(
\log_{10}\widehat{\rho}_{T_a}
-
\log_{10}\widehat{\rho}_{T_b}
\right)^2
\right]^{1/2}.
\label{eq:log_tdos_distance}
\end{equation}

Across all pairwise comparisons, the average log-space correlation is
\[
\langle C_{\log}\rangle = 0.988,
\]
while the largest-subspace comparison reaches
\[
C_{\log}(24,28)=0.993.
\]
The corresponding average logarithmic RMS distance is
\[
\langle\Delta_{\log}\rangle =0.231.
\]

As an independent measure of convergence, we compute the integrated
absolute spectral difference,
\begin{equation}
D_1(T_a,T_b)
=
\int d\lambda\,
\left|
\widehat{\rho}_{T_a}(\lambda)
-
\widehat{\rho}_{T_b}(\lambda)
\right|.
\label{eq:integrated_tdos_distance}
\end{equation}
Using the largest measured subspace as the reference gives
\[
D_1(24,28)=1.39\times10^{-2},
\]
more than an order of magnitude smaller than the corresponding
8-to-28-token difference.  The measured TDOS has therefore
effectively approached its large-subspace form by approximately
24 retained tokens.

Taken together, prompt robustness and token-subspace convergence
establish two complementary properties of the measured relaxation
spectrum.  Different input-dependent hidden states recover the same
coarse-grained TDOS, while progressively larger state-space
measurements converge toward a common macroscopic distribution.
The TDOS can therefore be identified as a reproducible collective
dynamical quantity rather than a property of a particular prompt or
finite local observation.

This macroscopic universality does not imply input-independent
Transformer dynamics.  Prompt-specific information remains encoded
in the hidden representations, eigenvectors, mode amplitudes, and
nonlinear trajectories.  Nor should macroscopic convergence of the
TDOS be identified by itself with infrared fixed-point behavior.
The former establishes the reproducibility of the relaxation-rate
distribution under changes of local state and observation scale,
whereas the latter concerns the asymptotic scaling of its
low-$\lambda$ sector as $\lambda\rightarrow0$, as discussed in
Sec.~II.D.

The complete prompt-resolved measurements and additional robustness
analyses are presented in Appendix~C.

\begin{figure*}[t]
\centering
\includegraphics[width=1.0\textwidth, trim=0cm 2cm 0cm 0cm]{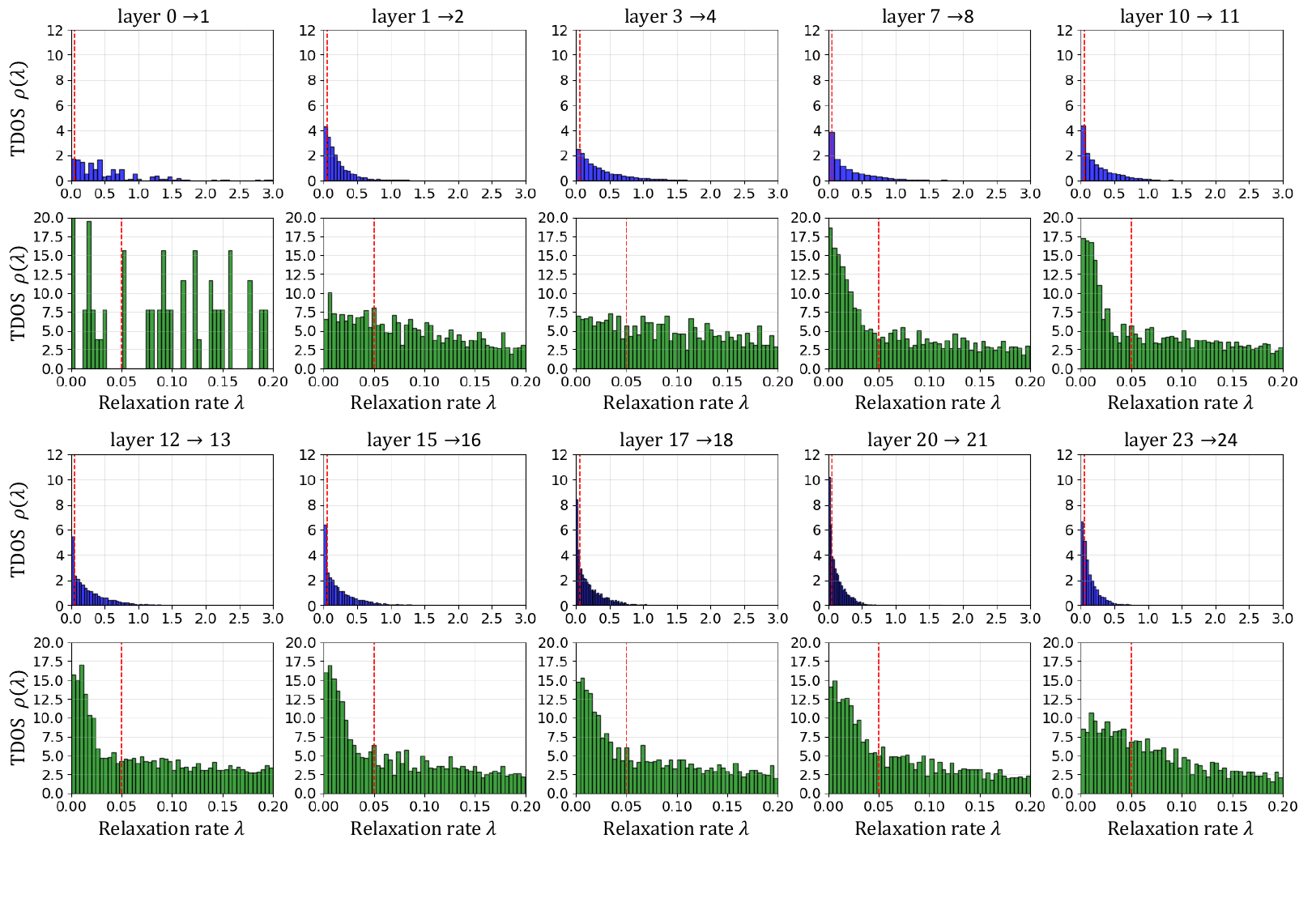}
\caption{
Layer-dependent organization of the time-scale density of states
measured from the fully trained Pythia-410M model.
Representative relaxation spectra are obtained from Jacobians computed
between successive neighboring Transformer blocks spanning the entire
network hierarchy, from the Layer~0$\!\rightarrow\!$1 mapping to the
final Layer~23$\!\rightarrow\!$24 mapping.
The upper panels show the complete TDOS, while the lower panels enlarge
the infrared region below the cutoff
$\lambda_{\rm cut}$.
A distinct behavior is observed for the
Layer~0$\!\rightarrow\!$1
mapping, which corresponds to the first Transformer block operating
directly on embedding representations before the formation of a
globally coupled collective hidden state.
By contrast, all subsequent neighboring-layer Jacobians exhibit a
remarkably similar infrared organization, indicating that the
collective hidden-state dynamics established by the first Transformer
block is preserved throughout the remaining Transformer hierarchy.
All investigated layers exhibit continuous relaxation spectra without
isolated infrared modes, confirming that collective memory is encoded
by a hierarchy of relaxation modes rather than by a single
characteristic timescale.
The infrared spectral weight develops a pronounced spatial hierarchy,
with the strongest accumulation of slow relaxation modes appearing in
the intermediate and deep hidden layers.
}
\label{fig:phase_portrait}
\end{figure*}

%====================================
\section{V. Layer-Dependent Organization of the Cognitive Field}

The preceding sections showed that Transformer learning substantially
reorganizes an already existing collective relaxation spectrum.
This reorganization produces a nonmonotonic evolution of the memory
self-energy, including a transient maximum followed by stabilization
toward a metastable near-critical regime within Cognitive Field Theory.
We now examine how these collective dynamical properties are
distributed across Transformer depth.

Because information is successively transformed through the hidden
layers, the detailed collective relaxation spectrum and memory response
need not be uniform throughout the network.
Different layers may exhibit different spectral weights and response
strengths while preserving the same underlying infrared organization.

To examine this spatial structure, we apply the spectral analysis of
Sec.~III to neighboring hidden-layer mappings spanning the Transformer
depth.
For each mapping, the Jacobian spectrum is used to determine the TDOS,
memory self-energy, inverse-self-energy gap proxy, and memory kernel.
These measurements provide a layer-resolved characterization of the
collective relaxation dynamics and its associated memory response.

\subsection{A. Layer-dependent organization of the time-scale density of states}

We first examine how the collective relaxation spectrum varies across
the Transformer hierarchy.
Using the same measurement procedure introduced in Sec.~III, the TDOS
is evaluated from neighboring-layer Jacobians of the fully trained
model.

Figure~10 compares representative TDOS measurements from the
Layer~0$\!\rightarrow\!$1 mapping through the final
Layer~23$\!\rightarrow\!$24 mapping.
The upper panels show the complete measured relaxation spectrum,
whereas the lower panels enlarge the region below the reference cutoff
$\lambda_{\rm cut}$ to resolve the slow-relaxation sector.

The Layer~0$\!\rightarrow\!$1 mapping exhibits a clear spectral
difference from the subsequent neighboring-layer mappings.
This transition is structurally distinct because it propagates the
embedding representation through the first Transformer block, whereas
all later mappings act on hidden states that have already undergone
Transformer-block processing.
The anomalous first-layer spectrum should therefore be distinguished
from the relaxation organization observed throughout the subsequent
hidden-state hierarchy.

From Layer~1$\!\rightarrow\!$2 onward, the TDOS exhibits a broadly
common infrared structure across network depth.
The complete spectral profiles and the amount of low-$\lambda$
spectral weight remain layer dependent, but the slow-relaxation sector
retains the same qualitative organization throughout the hierarchy.
This distinction between layer-dependent spectral weight and robust
infrared structure is central to the depth-resolved measurements.

Several features are common across the measured layers.
The relaxation spectrum remains continuous over a broad range of
rates, with no evidence that long-time dynamics is controlled by a
single isolated slow mode.
Instead, each layer supports a distributed hierarchy of relaxation
timescales, consistent with collective memory arising from a continuum
of slow dynamical modes.

At the same time, the spectral weight within this common structure
varies systematically with layer position.
The early hidden layers exhibit comparatively weaker infrared weight,
while intermediate and deeper layers show a stronger population of
slow relaxation modes.
The infrared contribution remains substantial through much of the
network and becomes moderately weaker toward the final layers.

The layer dependence therefore concerns primarily the distribution of
spectral weight rather than the appearance of a qualitatively different
relaxation spectrum.
Different parts of the Transformer hierarchy contribute different
strengths of slow-mode dynamics while preserving a broadly common
infrared spectral organization.

Within Cognitive Field Theory, these measurements are consistent with
a distributed collective field rather than one associated with a
single hidden layer.
The slow-mode manifold extends across Transformer depth, while its
spectral weight and corresponding memory response vary with layer
position.
The consequences of this layer dependence are examined below through
the memory self-energy and memory-kernel measurements.

\begin{figure}[t]
\centering
\includegraphics[scale=0.53, trim= 0cm 0.5cm 0cm 0cm]{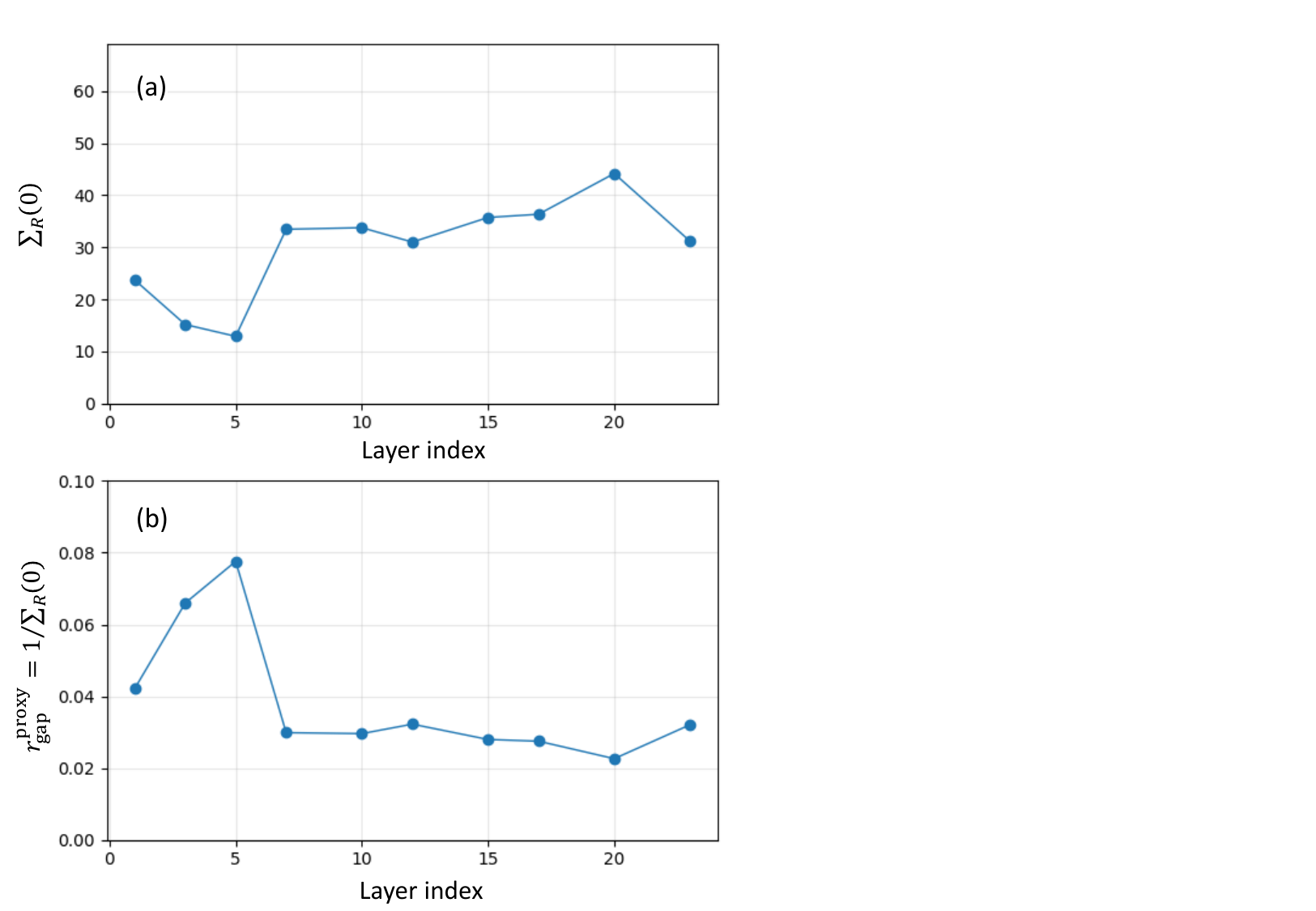}
\caption{
Layer-dependent memory self-energy and cognitive forgetting gap
obtained from the fully trained Pythia-410M model.
(a) Static memory self-energy calculated directly from the
experimentally measured TDOS for representative hidden-layer mappings.
(b) Corresponding normalized proxy of the cognitive forgetting gap,
$r_{\rm gap}^{\rm proxy}=1/\Sigma_R(0)$.
The memory self-energy increases markedly from the early layers,
reaches its largest values in the intermediate and deep hidden layers,
and decreases moderately toward the output layer.
The forgetting gap exhibits the complementary behavior, attaining its
smallest values where the memory self-energy is largest.
}
\label{fig:phase_portrait}
\end{figure}

\subsection{B. Layer-dependent memory self-energy and forgetting-gap proxy}

The layer-dependent redistribution of infrared spectral weight
identified in Fig.~10 naturally raises the question of how the
corresponding collective memory response varies across the Transformer
hierarchy.

For each neighboring-layer mapping, the static memory self-energy is
evaluated from the measured relaxation spectrum using the procedure
introduced in Sec.~III.C.
Because the absolute bare forgetting rate is not independently
determined by the Jacobian spectrum, the cognitive forgetting gap
\(r_{\rm cog}=r-\Sigma(0)\) is not assigned an absolute experimental
value.
We therefore compare the measured self-energy with the corresponding
normalized inverse-self-energy proxy introduced in Sec.~III.
Figure~11(a) shows the layer-dependent memory self-energy, while
Fig.~11(b) presents the corresponding forgetting-gap proxy.

The layer dependence closely follows the redistribution of infrared
spectral weight observed in Fig.~10.
The early hidden layers exhibit relatively weak memory
renormalization, whereas the self-energy increases substantially
through the intermediate layers, remains enhanced throughout much of
the deep hierarchy, and reaches its largest value near Layer~20 before
decreasing moderately toward the final layer.
The inverse-self-energy proxy exhibits the complementary behavior,
reaching its minimum over the same depth range.

The correspondence between Figs.~10 and~11 directly connects the
layer-dependent slow-mode spectrum with the strength of the collective
memory response.
Layers with greater infrared spectral weight generally exhibit stronger
memory self-energy renormalization, while layers with weaker infrared
weight exhibit a correspondingly smaller response.

Within Cognitive Field Theory, this layer dependence is consistent
with a distributed cognitive field whose memory dressing varies across
Transformer depth.
The collective field is therefore not associated with a single hidden
layer: different parts of the hierarchy contribute different strengths
of infrared spectral weight and memory renormalization while sharing
the broadly common infrared organization identified in Sec.~V.A.

The layer-resolved measurements complement the training-dependent
dynamics of Sec.~III.
Training reorganizes the collective relaxation spectrum and its memory
self-energy, while the fully trained network exhibits a corresponding
depth-dependent hierarchy of memory response.
Figures~10 and~11 therefore reveal complementary temporal and spatial
structure in the measured collective dynamics.

\begin{figure}[t]
\centering
\includegraphics[scale=0.5, trim= 0cm 0.5cm 0cm 0cm]{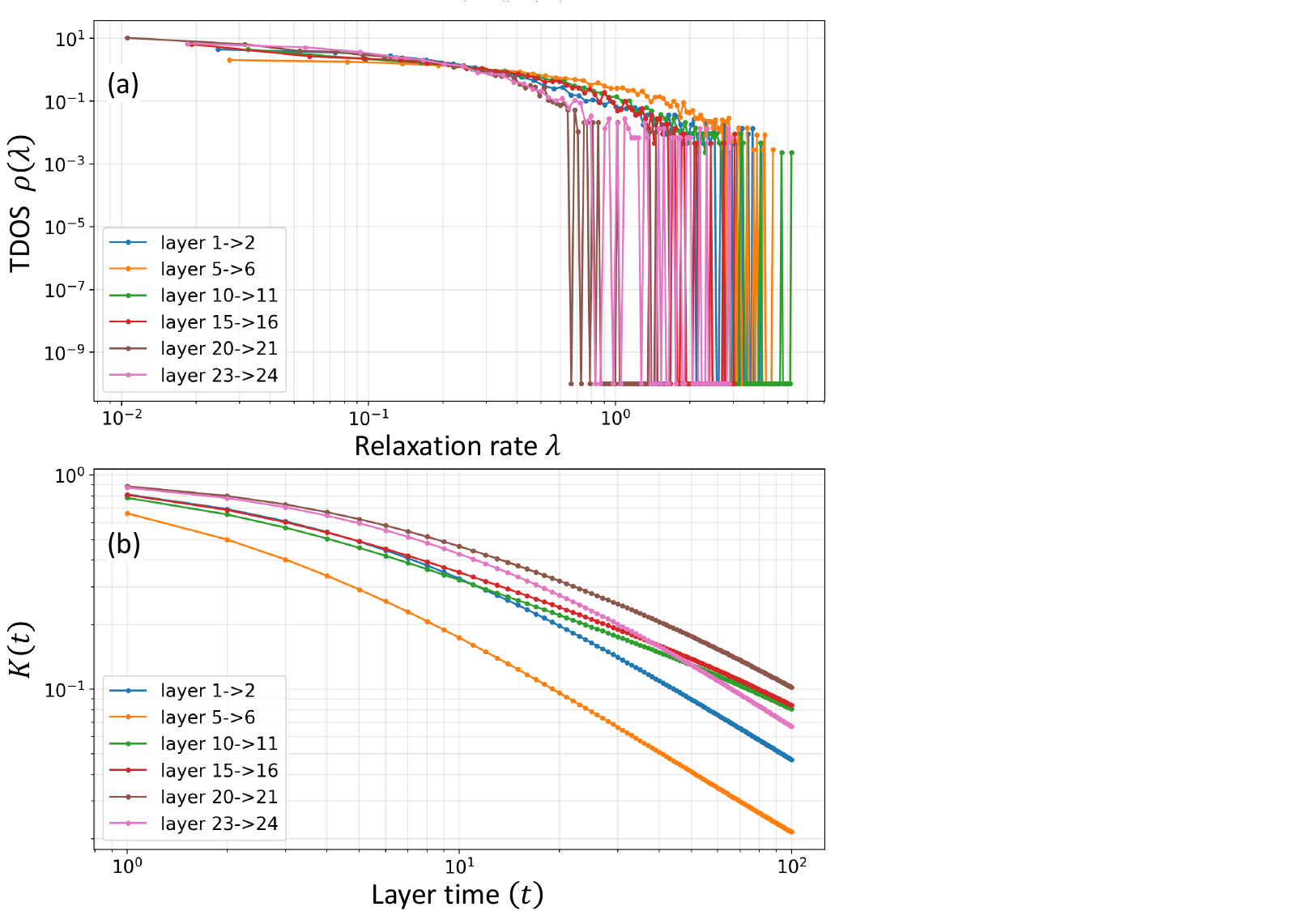}
\caption{
Layer-dependent universality of the infrared memory dynamics measured
from the fully trained Pythia-410M model.
(a) Time-scale density of states for representative
hidden-layer mappings displayed on logarithmic scales.
Although the infrared spectral weight varies systematically with layer
depth, all investigated layers exhibit a common infrared scaling
behavior, indicating that the collective slow-mode manifold extends
throughout the Transformer hierarchy.
(b) Memory kernels computed directly from the experimentally measured
TDOS.
The strength of the long-time memory varies with network depth,
reflecting the layer-dependent memory self-energy, whereas every layer
retains the same scale-free temporal behavior.
}
\label{fig:phase_portrait}
\end{figure}

\subsection{C. Layer hierarchy of scale-free memory dynamics}

We finally examine whether the common infrared organization identified
across layers produces a corresponding common long-time memory law.

Figure~12 compares representative layer-dependent TDOS measurements
and the corresponding memory kernels obtained from the fully trained
Pythia-410M model.
Figure~12(a) presents the infrared TDOS on logarithmic scales, while
Fig.~12(b) shows the memory kernels computed directly from the measured
relaxation spectra.

Despite quantitative differences in spectral weight, the investigated
layers exhibit similar scaling in the resolved infrared regime.
The low-\(\lambda\) TDOS remains close to the approximately flat
spectral class throughout the hidden-state hierarchy, while the
complete spectral profiles retain substantial layer dependence.
Thus, the layer dependence primarily modifies the strength and detailed
distribution of the relaxation spectrum rather than its infrared
scaling form.

The corresponding memory kernels exhibit the same separation between
layer-dependent magnitude and robust long-time scaling.
Across the investigated layers, the kernels display an extended
power-law regime rather than relaxation governed by a single
characteristic timescale.
Although their amplitudes vary with layer position, their long-time
functional form remains close to the inverse-time scaling
\begin{equation}
K_l(t)
\sim
\frac{1}{t}.
\end{equation}
This behavior follows directly from the measured infrared spectra
through
\begin{equation}
K_l(t)
=
\int_0^\infty
d\lambda\,
\rho_l(\lambda)e^{-\lambda t},
\end{equation}
and is consistent with the infrared relation
\begin{equation}
\rho_l(\lambda)
\sim
\lambda^{\beta_l},
\qquad
\beta_l\simeq0
\quad\Longrightarrow\quad
K_l(t)
\sim
t^{-(1+\beta_l)}
\simeq
t^{-1}.
\end{equation}

Figures~10--12 therefore reveal two complementary aspects of the
layer-resolved collective dynamics.
The amount of infrared spectral weight and the corresponding memory
self-energy vary systematically with network depth, whereas the
infrared scaling structure and long-time memory law remain broadly
robust across layers.
Within Cognitive Field Theory, these observations are consistent with
a distributed cognitive field possessing a layer-dependent memory
response while remaining governed by a common infrared dynamical
organization across the Transformer hierarchy.

\begin{figure*}[t]
\centering
\includegraphics[width=1.0\textwidth, trim=0cm 0.5cm 0cm 0cm]{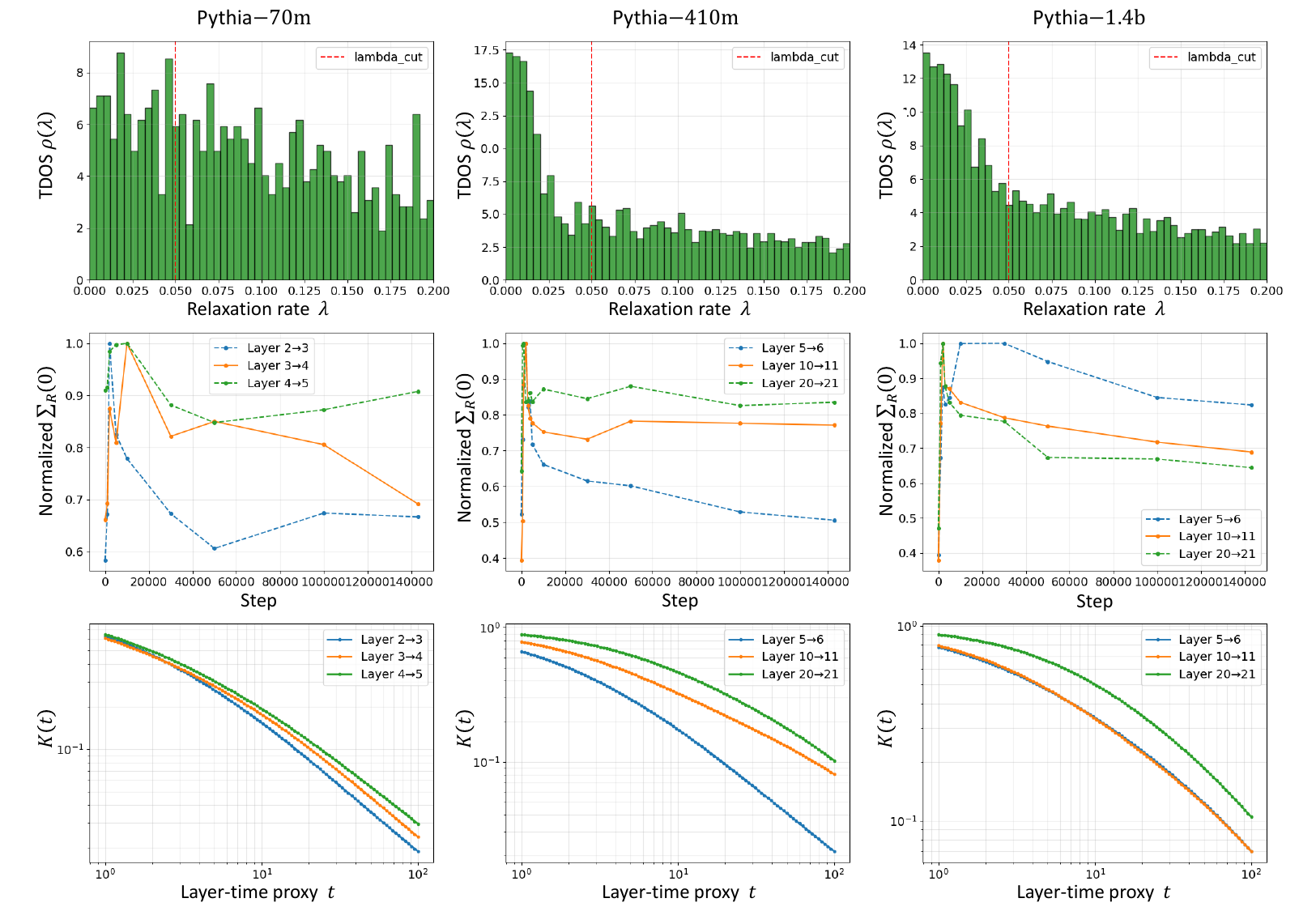}
\caption{
Model-scale universality of the infrared organization during
Transformer learning.
The complete infrared analysis is repeated for the publicly available
Pythia-70M, Pythia-410M, and Pythia-1.4B models.
Top panels: time-scale density of states measured near the
infrared cutoff, showing the accumulation of slow relaxation modes
during learning.
Middle panels: normalized static memory self-energy, exhibiting a
pronounced transient maximum followed by relaxation toward a
metastable near-critical operating regime.
Bottom panels: memory kernels computed directly from the measured TDOS,
demonstrating robust scale-free long-time memory across all model
sizes.
Despite spanning nearly two orders of magnitude in trainable
parameters, all investigated models exhibit the same qualitative
sequence of infrared spectral reorganization, transient critical
formation of the cognitive field, and scale-free memory dynamics.
}
\label{fig:phase_portrait}
\end{figure*}

\begin{figure*}[t]
\centering
\includegraphics[width=1.0\textwidth, trim=0cm 1.7cm 0cm 0cm]{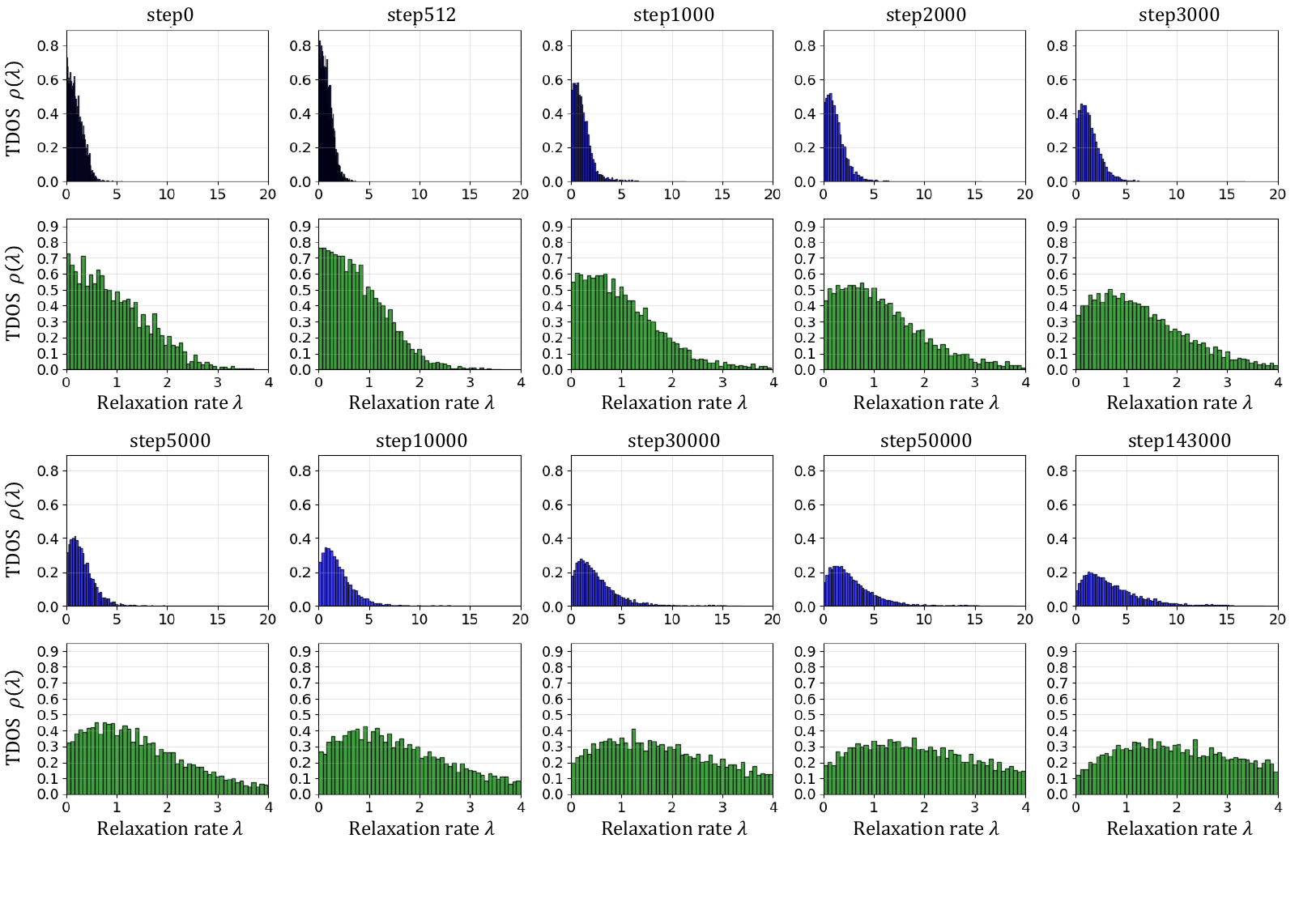}
\caption{
Evolution of the time-scale density of states during training
for the layer $1\!\rightarrow\!24$ propagation.
The upper panels show the complete relaxation spectrum, while the lower
panels enlarge the infrared region containing the slow relaxation
modes.
During the earliest stage of optimization, the spectrum is dominated by
fast relaxation modes.
As training proceeds, slow relaxation modes progressively accumulate,
leading to a broad infrared reorganization of the TDOS.
After the transient critical stage, the relaxation spectrum approaches
a nearly stationary distribution that remains stable throughout the
remainder of training.
The persistence of the infrared organization over the long propagation
distance demonstrates that the collective relaxation dynamics is not
restricted to neighboring layers but extends across the Transformer
network.
}
\label{fig:phase_portrait}
\end{figure*}

%=======================================
\section{VI. Model-Size Dependence of Infrared Organization}

An important question is whether the infrared organization identified
in the preceding sections depends on Transformer model size or instead
represents a robust collective property of the learned dynamics.

To address this question, we repeated the spectral analysis for three
publicly available Pythia models containing approximately 70M, 410M,
and 1.4B parameters.
For each model, the Jacobian spectrum was extracted at representative
training checkpoints, from which the TDOS, memory self-energy, and
memory kernel were evaluated using the same analysis procedure.
Representative results for Pythia-410M are presented in the main
analysis, while the corresponding measurements for Pythia-70M and
Pythia-1.4B are provided in Appendices~D and~E.

Figure~13 summarizes the infrared evolution measured for the
Pythia-70M, Pythia-410M, and Pythia-1.4B models.
Despite spanning nearly two orders of magnitude in the number of
trainable parameters, all three models exhibit the same characteristic
organization of the collective relaxation spectrum.
Learning substantially reorganizes the TDOS, the memory self-energy
undergoes a nonmonotonic evolution with a transient maximum, and the
corresponding memory kernels preserve scale-free long-time behavior
across all investigated model sizes.

The upper panels compare the measured TDOS in the slow-relaxation
sector.
A broad infrared spectrum is already present before training in each
model and remains well defined throughout optimization.
Learning substantially redistributes spectral weight within this
spectrum, while an approximately flat low-$\lambda$ sector persists
across model sizes.
Although the detailed TDOS profiles differ quantitatively, their
common infrared organization indicates that the slow-relaxation
spectral class is robust against substantial changes in model size.

The middle panels show the corresponding evolution of the memory
self-energy.
In every investigated model, the self-energy evolves nonmonotonically,
reaching a pronounced transient maximum during early training before
approaching a comparatively stationary late-training value.
The recurrence of this characteristic evolution across model sizes
demonstrates that the measured transient memory renormalization is not
specific to a single Pythia model.

Within Cognitive Field Theory, the self-energy maximum corresponds to
the strongest suppression of the cognitive forgetting gap,
\[
r_{\rm cog}
=
r-\Sigma(0),
\]
and therefore to the largest collective susceptibility.
The measurements thus identify a common transient regime of maximal
memory renormalization which, within the theoretical framework, is
interpreted as the critical formation stage of the macroscopic
cognitive field.
The subsequent reduction and stabilization of the self-energy are
correspondingly consistent with evolution toward a metastable
near-critical operating regime.

The lower panels show the memory kernels obtained directly from the
measured relaxation spectra.
Across all three model sizes, the kernels exhibit an extended
scale-free long-time regime close to
\[
K(t)
\sim
\frac{1}{t}.
\]
Although their amplitudes and detailed temporal profiles vary
quantitatively, the long-time scaling remains essentially unchanged.
The persistence of this behavior across model size provides an
independent dynamical manifestation of the common infrared
organization observed in the TDOS.

This interpretation is further supported by the infrared exponent
analysis of Sec.~III.E.
Across the investigated models, the low-$\lambda$ TDOS remains close
to 
\(
\beta\simeq0.
\)
The corresponding relation
\[
K(t)
\sim
t^{-(1+\beta)}
\]
is quantitatively consistent with the approximately inverse-time
memory scaling measured directly from the relaxation spectra.
The agreement of these independently extracted infrared observables
indicates that learning reorganizes spectral weight without
substantially changing the resolved infrared scaling class.

Taken together, these measurements establish a clear separation
between model-dependent spectral details and model-size-robust
infrared organization.
The complete TDOS and the magnitude of the collective response vary
with network size, whereas the approximately flat infrared spectrum,
scale-free long-memory dynamics, and nonmonotonic evolution of the
memory self-energy recur across Pythia models ranging from 70M to
1.4B parameters.
Within Cognitive Field Theory, this common dynamical organization is
consistent with the same infrared mechanism of macroscopic cognitive
field formation operating across substantially different Transformer
model scales.

\begin{figure}[t]
\centering
\includegraphics[scale=0.47, trim= 0.2cm 7.8cm 0cm 0cm]{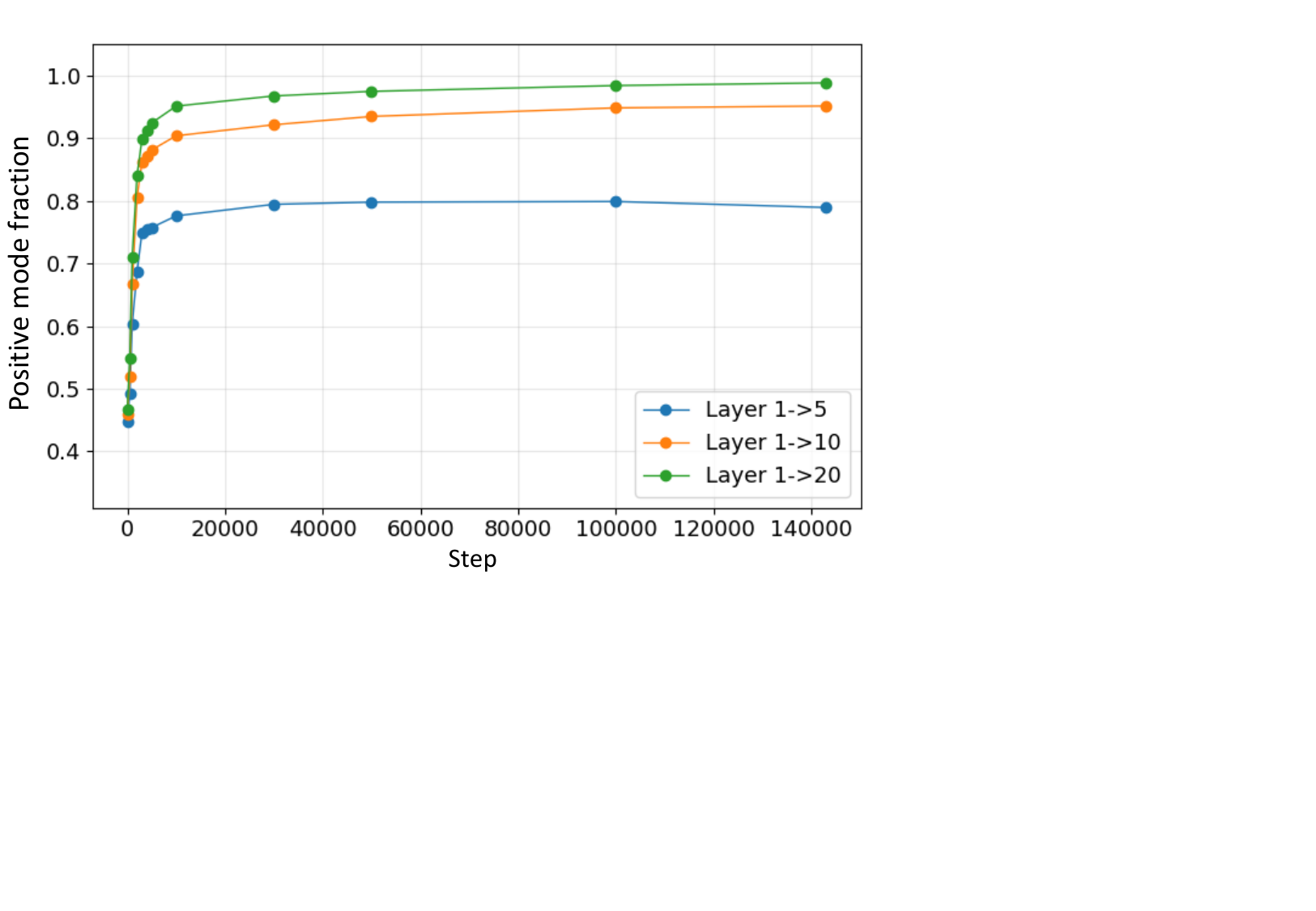}
\caption{
Fraction of stable relaxation modes measured from the fully trained
Transformer as a function of optimization step for different layer
propagation depths.
For all propagation distances, the fraction of stable modes increases
rapidly during the earliest stage of training and subsequently
approaches an almost stationary value.
The final stable-mode population increases systematically with
propagation depth, reaching approximately $0.79$, $0.95$, and $0.99$
for the $1\!\rightarrow\!5$, $1\!\rightarrow\!10$, and
$1\!\rightarrow\!20$ propagations, respectively.
This monotonic increase indicates that long-range layer propagation is
progressively dominated by stable collective relaxation modes,
demonstrating that the large-scale Transformer dynamics evolves toward
a globally stabilized collective state after training.
}
\label{fig:phase_portrait}
\end{figure}

\begin{figure}[t]
\centering
\includegraphics[scale=0.55, trim= 0cm 0.58cm 0cm 0cm]{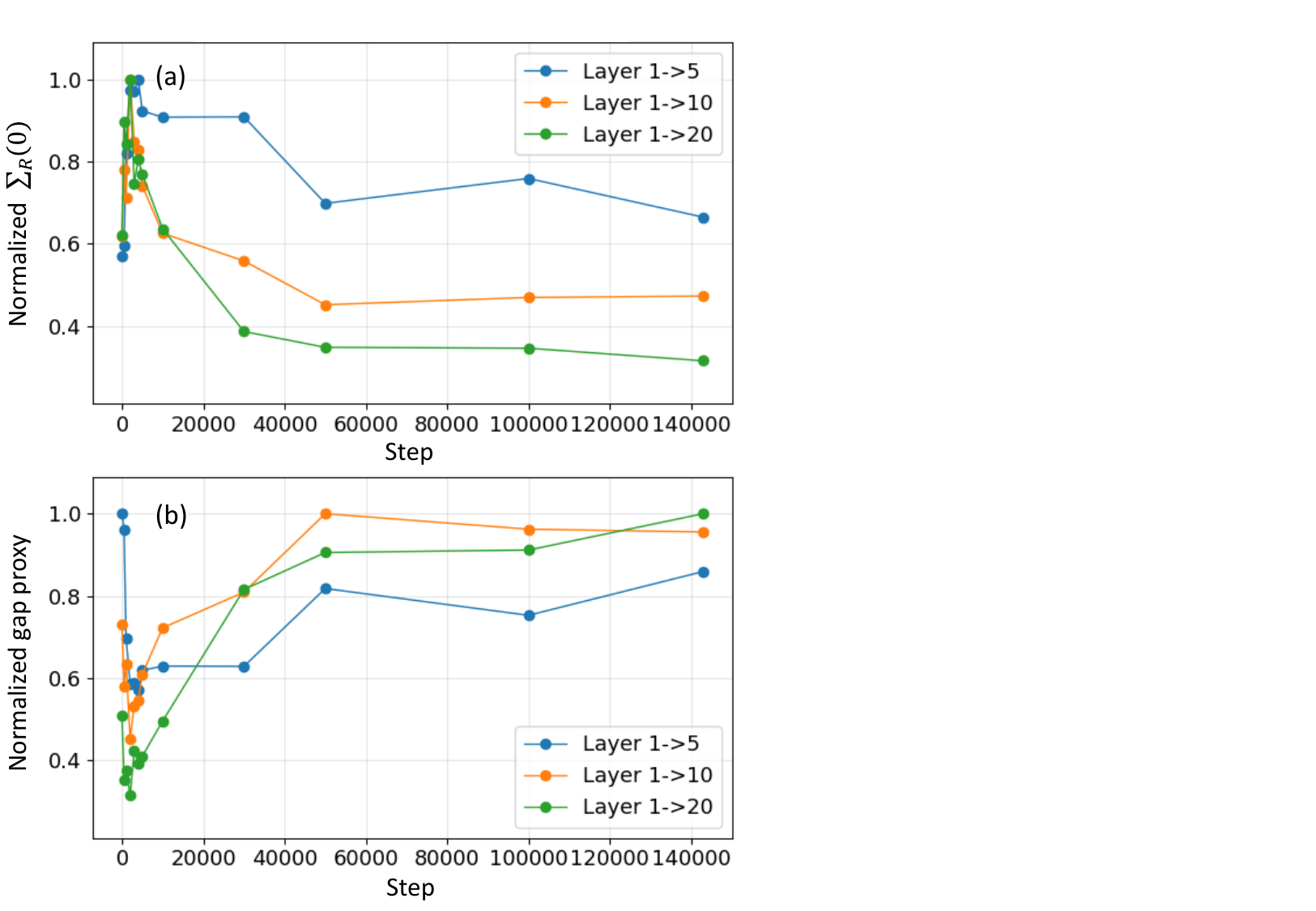}
\caption{
Evolution of the normalized memory self-energy (upper panel) and the
normalized cognitive forgetting-gap proxy (lower panel) for different
layer propagation depths.
The memory self-energy exhibits a pronounced maximum, whereas the
forgetting gap proxy reaches a corresponding minimum during the early stage
of training for all propagation distances.
This anti-correlated behavior indicates robust critical enhancement of
the collective susceptibility throughout the Transformer network.
}
\label{fig:phase_portrait}
\end{figure}

%=================================================
\section{VII. Depth-independent infrared organization of Transformer dynamics}

The analyses presented thus far have primarily focused on the
collective dynamics between neighboring Transformer blocks.
An important question is whether the observed infrared organization is
a local property of adjacent blocks or a collective dynamical property
that persists across Transformer depth.

As established in Sec.~II.B, a single self-attention block already
defines a state-dependent globally coupled dynamical mapping, while
successive residual blocks propagate and reorganize the accumulated
hidden state along network depth.
For propagation from block \(l\) to block \(l+n\), the corresponding
linear response is described by the composed Jacobian
\begin{equation}
J_{l\rightarrow l+n}
=
J_{l+n-1}
J_{l+n-2}
\cdots
J_l.
\label{eq:depth_composed_jacobian_sec7}
\end{equation}
Its eigenmodes therefore characterize collective directions that
persist, decay, or rotate across the complete propagation interval
rather than within a single block.

The response-theoretic construction of Sec.~II predicts that increasing
the propagation depth may quantitatively modify the accumulated
stability and overall response strength without changing the
asymptotic infrared organization of the collective spectrum.
If the measured slow-mode structure represents a macroscopic property
of Transformer dynamics, the infrared TDOS, long-time memory kernel,
and training-dependent evolution of the collective response should
remain robust across widely separated propagation intervals.

To test this prediction, we repeated the analysis using Jacobians
spanning progressively larger propagation depths, including
\(1\!\rightarrow\!5\),
\(1\!\rightarrow\!10\),
\(1\!\rightarrow\!15\),
\(1\!\rightarrow\!20\), and
\(1\!\rightarrow\!24\).
For each propagation interval, we computed the relaxation spectrum,
TDOS, memory kernel, and memory self-energy using the same measurement
protocol adopted for neighboring blocks.

Figure~14 shows the TDOS evolution during training for the
\(1\!\rightarrow\!24\) propagation.
A broad slow-relaxation sector is already present before training and
remains well defined throughout optimization.
Learning substantially reorganizes the distribution of relaxation
rates, while the large-scale infrared structure remains similar to
that observed for neighboring blocks.
Thus, although both training and propagation depth modify the detailed
spectral profile, they do not qualitatively alter the infrared
organization of the relaxation spectrum.

Figure~15 compares the fraction of stable relaxation modes across
different propagation depths.
For every propagation interval, the stable-mode population increases
rapidly during the early stage of optimization before approaching an
approximately stationary value.
The final stable-mode fraction increases systematically with
propagation depth and approaches unity for the
\(1\!\rightarrow\!20\) and
\(1\!\rightarrow\!24\) propagations.
Long-range propagation in the trained network is therefore
increasingly dominated by stable collective relaxation modes.

The corresponding memory self-energies are shown in Fig.~16.
For every propagation depth, the normalized memory self-energy
exhibits a pronounced transient maximum during early training.
The corresponding inverse-self-energy proxy reaches its minimum near
the same checkpoint and subsequently evolves in the opposite
direction as the self-energy approaches a nearly stationary
late-training value.
Within Cognitive Field Theory, this nonmonotonic evolution corresponds
to a transient suppression of the cognitive forgetting gap and an
enhancement of the collective susceptibility.
The recurrence of the same characteristic evolution across widely
separated propagation intervals demonstrates that the observed
critical-response dynamics is not restricted to neighboring
Transformer blocks.

The TDOS and corresponding memory kernels of the fully trained
Transformer are summarized in Fig.~17.
Despite quantitative differences in the complete relaxation spectra,
all propagation depths exhibit an approximately flat infrared TDOS.
The corresponding memory kernels display an extended linear regime on
log--log scales, with fitted exponents over the resolved scaling
interval satisfying
\begin{equation}
K(t)
\sim
t^{-\alpha},
\end{equation}
with
\begin{equation}
0.90
\lesssim
\alpha
\lesssim
1.11.
\end{equation}
These values are quantitatively consistent with the marginal
long-memory scaling
\[
K(t)
\sim
\frac{1}{t}.
\]

No power-law form is imposed in evaluating the memory kernel.
For each propagation interval, it is obtained directly from the
measured relaxation spectrum through
\begin{equation}
K(t)
=
\int_0^\infty
d\lambda\,
\rho(\lambda)e^{-\lambda t}.
\end{equation}
The persistence of the same long-time scaling over widely different
propagation depths therefore follows directly from the measured
infrared relaxation spectra.
In particular, the result is consistent with the infrared relation
derived in Sec.~II,
\begin{equation}
\rho(\lambda)
\sim
\lambda^\beta,
\qquad
\beta\simeq0,
\end{equation}
which gives
\begin{equation}
K(t)
\sim
t^{-(1+\beta)}
\sim
t^{-1}.
\end{equation}

These measurements reveal a clear separation between quantitative
depth dependence and infrared universality.
Increasing the propagation interval changes the complete TDOS, the
stable-mode population, and the magnitude of the accumulated memory
response.
In contrast, the approximately flat low-\(\lambda\) spectrum, the
corresponding \(K(t)\sim1/t\) long-memory scaling, and the
characteristic nonmonotonic evolution of the memory self-energy remain
robust across propagation depth.

Each Transformer block defines a globally coupled dynamical mapping
through self-attention and feature-space mixing, while depth-ordered
residual composition propagates and reorganizes the accumulated
collective state across the Transformer hierarchy.
Within Cognitive Field Theory, the persistence of the same infrared
organization across short- and long-range propagation indicates that
the macroscopic cognitive field is not associated with a particular
hidden layer but with hierarchically organized collective dynamics
extending across Transformer depth.

\begin{figure}[t]
\centering
\includegraphics[scale=0.49, trim= 0.3cm 0.5cm 0cm 0cm]{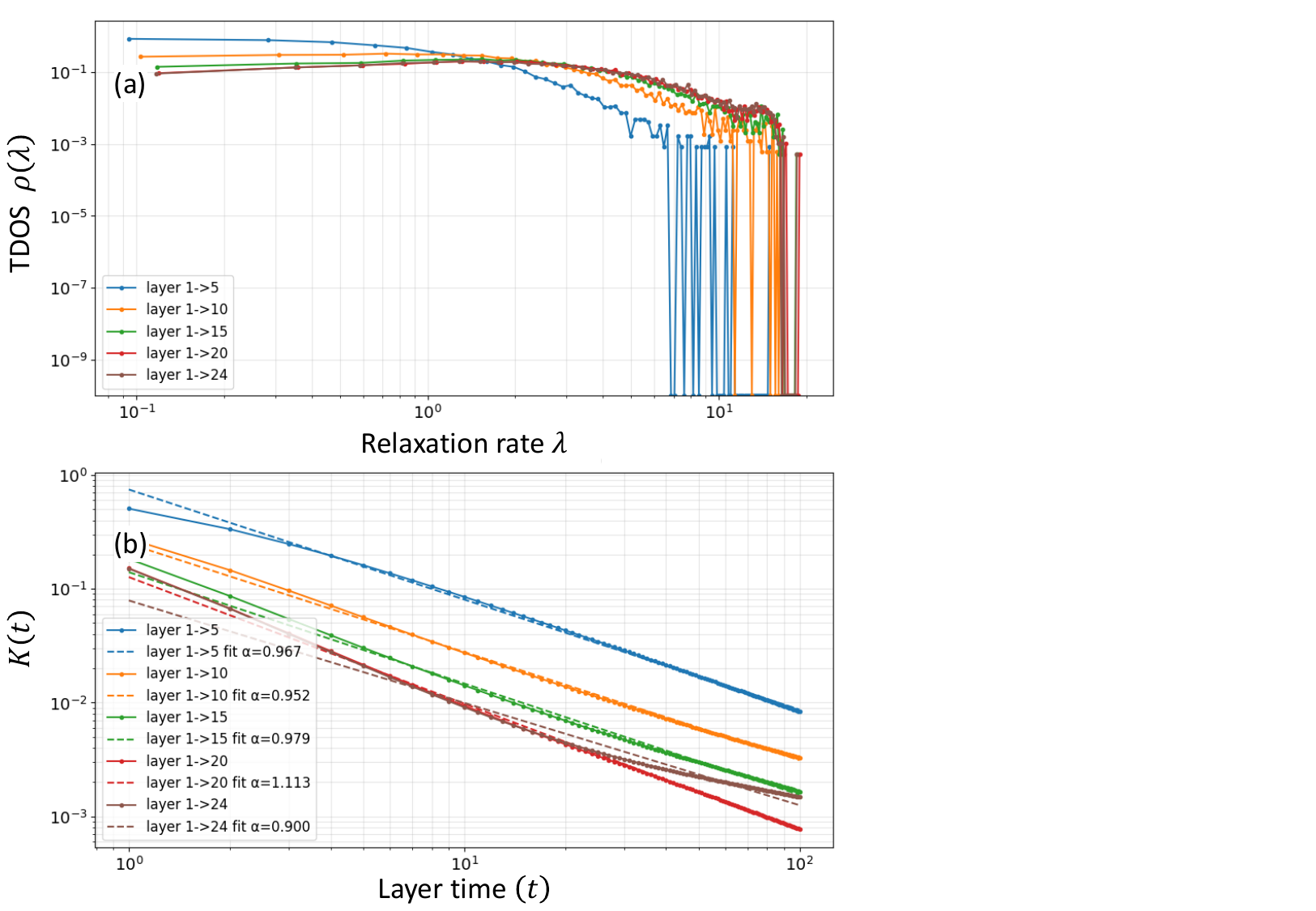}
\caption{
TDOS and corresponding memory kernels of the fully trained
Pythia-410M model for different propagation depths.
Although the TDOS varies with layer distance, all kernels exhibit
$K(t)\sim t^{-\alpha}$ with
$0.90\lesssim\alpha\lesssim1.11$, demonstrating an approximately
depth-independent $1/t$ long-memory scaling.
}
\label{fig:phase_portrait}
\end{figure}

\begin{figure*}[t]
\centering
\includegraphics[width=1.0\textwidth, trim=0cm 8.7cm 0cm 0cm]{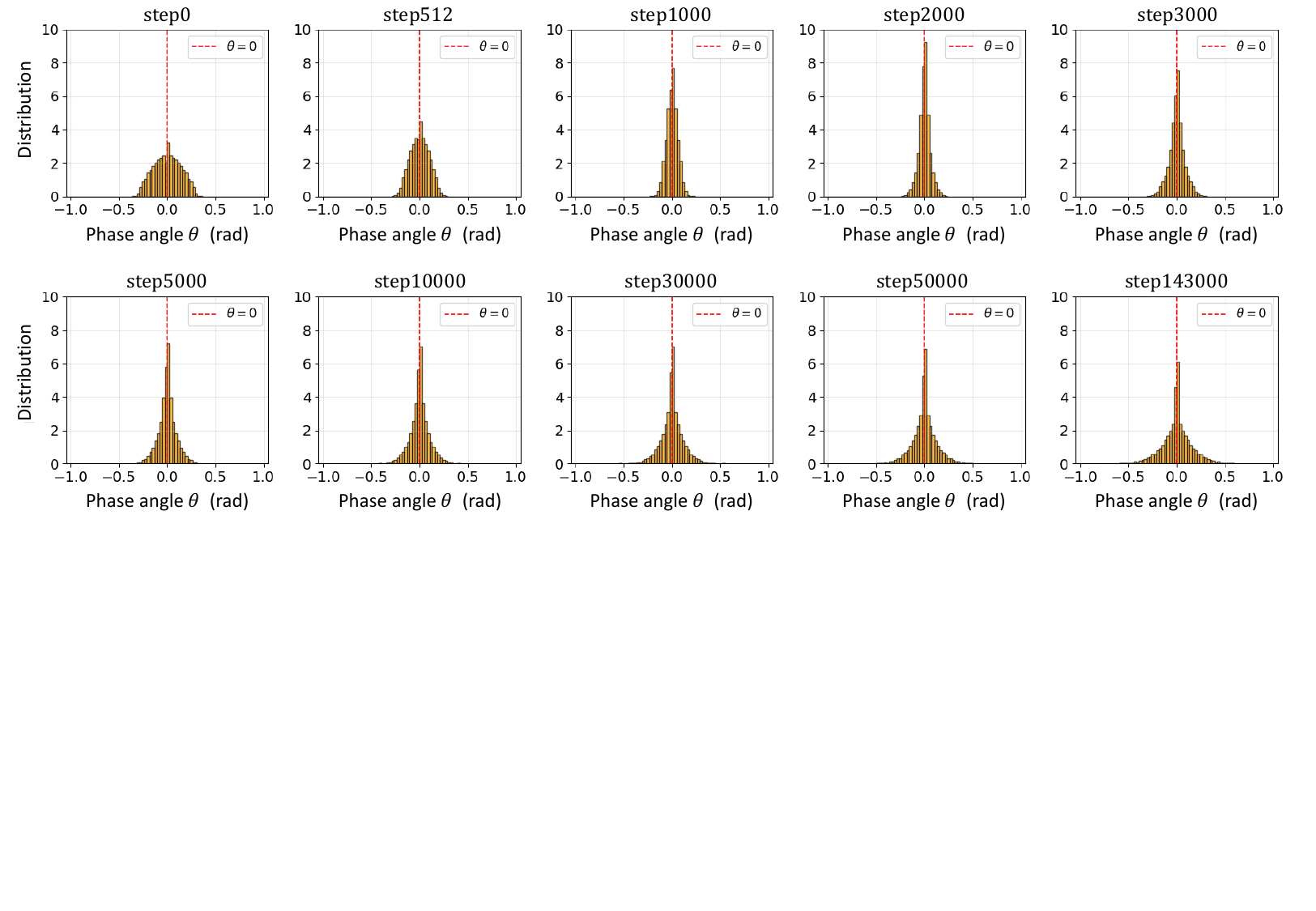}
\caption{
Evolution of the phase-angle distribution of the complex collective
Jacobian spectrum during Transformer learning.
Each panel shows the distribution of phase angles
$\theta=\arg(\mu)$
computed from the Jacobian spectrum of the Pythia-410M Transformer for
the layer-to-layer propagation
$10\!\rightarrow\!11$
at representative training checkpoints.
Initially the circulation spectrum is broadly distributed around
$\theta=0$,
indicating heterogeneous effective phase rotations among the
collective modes.
During the early stage of optimization, however, the distribution
rapidly contracts toward the low-frequency sector.
The strongest localization occurs near training steps
$10^3$--$2\times10^3$,
where the phase distribution becomes sharply concentrated around
$\theta\simeq0$.
Subsequently, the distribution gradually broadens while remaining
centered near
$\theta=0$,
preserving a pronounced low-frequency collective core.
The observed evolution demonstrates that Transformer learning
does not eliminate the diversity of collective circulation modes.
Instead, it reorganizes the circulation sector into a stable
low-frequency infrared core embedded within a broader distribution of
effective circulation frequencies.
}
\label{fig:phase_evolution}
\end{figure*}

\begin{figure}[t]
\centering
\includegraphics[scale=0.433, trim= 0.5cm 5.2cm 0cm 0cm]{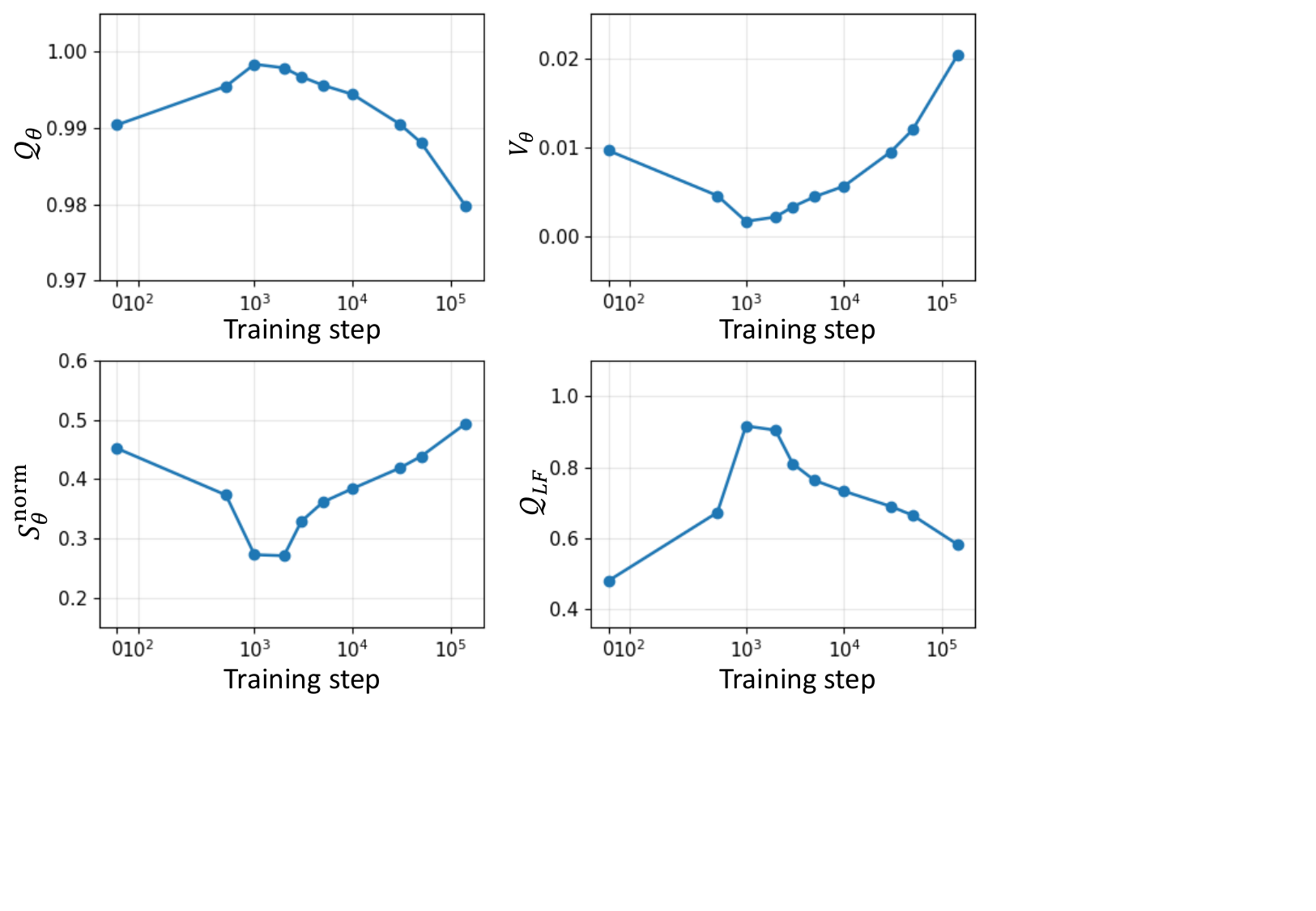}
\caption{
Evolution of the collective observables characterizing temporal
organization of the circulation sector during Transformer learning.
The four panels show the global phase order parameter
$\mathcal Q_\theta$,
the circular variance
$V_\theta$,
the phase entropy
$S_\theta$,
and the low-frequency population
$\mathcal Q_{\rm LF}$,
respectively.
Together, these observables quantify complementary aspects of the
circulation spectrum:
$\mathcal Q_\theta$
measures global phase alignment,
$V_\theta$
measures the dispersion of the phase distribution,
$S_\theta$
measures the spectral diversity of circulation modes,
and
$\mathcal Q_{\rm LF}$
measures the infrared population concentrated around
$\theta\simeq0$.
All four observables exhibit a common nonmonotonic evolution during
training.
Near training steps
$10^3$--$2\times10^3$,
the phase order reaches its maximum,
the circular variance and entropy attain their minima,
and the low-frequency population reaches its maximum,
demonstrating the transient infrared organization of the circulation
sector.
Subsequently, the observables partially recover while remaining
substantially different from their initial values, indicating the
formation of a stable low-frequency collective core rather than
complete global phase locking.
}
\label{fig:phase_observables}
\end{figure}

%===========================================
\section{VIII. Complex Spectral Organization during Transformer Learning}

The preceding sections established that Transformer learning
systematically reorganizes the relaxation sector of the collective
Jacobian spectrum while preserving its infrared long-time scaling.
We now extend the analysis to the imaginary sector and examine how the
circulation spectrum evolves during learning.

For each collective eigenmode, the complex Jacobian eigenvalue is
written as
\[
\zeta_\alpha
=
|\zeta_\alpha|e^{i\theta_\alpha},
\]
where the magnitude determines the effective relaxation rate,
\begin{equation}
\lambda_\alpha
=
-\frac{1}{\Delta s}
\ln|\zeta_\alpha|,
\label{eq:phase_relaxation_rate}
\end{equation}
and
\begin{equation}
\theta_\alpha
=
\arg\zeta_\alpha
\end{equation}
gives the rotational increment accumulated across the propagation
interval.

Consistent with the continuous collective-mode convention introduced
in Sec.~II.A, the discrete propagation eigenvalue may be represented as
\begin{equation}
\zeta_\alpha
=
\exp\left[-(\lambda_\alpha+i\omega_\alpha)\Delta s\right].
\end{equation}
Consequently,
\begin{equation}
\theta_\alpha
=
-\omega_\alpha\Delta s
\pmod{2\pi},
\end{equation}
and, on the principal branch,
\begin{equation}
\omega_\alpha
=
-\frac{\theta_\alpha}{\Delta s}.
\label{eq:phase_circulation_frequency}
\end{equation}
For adjacent Transformer blocks, $\Delta s=1$, so the eigenvalue angle
directly measures the circulation increment in radians per block, up
to the adopted sign convention.  The angle distribution therefore
provides the discrete-depth counterpart of the circulation-frequency
spectrum introduced in Cognitive Field Theory.

For the dominant collective modes of the trained Transformer, the
measured phase angles remain concentrated within approximately
\begin{equation}
|\theta|
\lesssim
0.05\text{--}0.1
\ {\rm rad/layer}.
\end{equation}
The corresponding circulation scale is
\begin{equation}
T_{\rm circ}
=
\frac{2\pi}{|\theta|}
\sim
60\text{--}120
\ {\rm layers}.
\end{equation}
Since Pythia-410M contains only twenty-four Transformer layers, the
present measurements probe only a fraction of a complete circulation
cycle.  The circulation sector should therefore be interpreted in
terms of low-frequency spectral organization rather than a directly
resolved global oscillation.

Figure~\ref{fig:phase_evolution} shows the evolution of the phase-angle
distribution throughout training.  At initialization, the circulation
spectrum is relatively broad.  During the early stage of optimization,
the distribution contracts rapidly toward $\theta=0$, with the
strongest concentration occurring near training steps
$10^3$--$2\times10^3$.  At later stages, the spectrum broadens again
while retaining a pronounced low-frequency population around the
origin.

This nonmonotonic evolution closely coincides with the training stage
at which the memory self-energy reaches its transient maximum and the
slow-relaxation spectrum undergoes its strongest reorganization.
Transformer learning therefore reorganizes both projections of the
complex collective spectrum: the relaxation sector toward small
$\lambda$ and the circulation sector toward small $|\omega|$.
Within Cognitive Field Theory, their simultaneous organization near
\begin{equation}
(\lambda,\omega)
\simeq
(0,0)
\end{equation}
is interpreted as the formation of a low-frequency, long-lived
collective sector accompanying the memory-dressed macroscopic
cognitive field.

To quantify the circulation-sector evolution, we introduce four
complementary spectral observables.  The angular concentration
parameter is
\begin{equation}
\mathcal Q_\theta
=
\left|
\frac{1}{N}
\sum_\alpha
e^{i\theta_\alpha}
\right|,
\end{equation}
with the corresponding circular variance
\begin{equation}
V_\theta
=
1-\mathcal Q_\theta.
\end{equation}
The phase-spectrum entropy is defined as
\begin{equation}
S_\theta
=
-\sum_i p_i\ln p_i,
\end{equation}
while the low-frequency population is
\begin{equation}
\mathcal Q_{\rm LF}
=
\int_{-\theta_c}^{\theta_c}
\rho(\theta)\,d\theta.
\end{equation}
Here $\mathcal Q_\theta$ measures angular concentration of the
circulation spectrum, $V_\theta$ its dispersion, $S_\theta$ its
spectral diversity, and $\mathcal Q_{\rm LF}$ the fraction of modes
within the selected low-frequency sector.

As shown in Fig.~\ref{fig:phase_observables}, all four observables
exhibit the same nonmonotonic training dependence.
During early optimization, $\mathcal Q_\theta$ and
$\mathcal Q_{\rm LF}$ increase, whereas $V_\theta$ and $S_\theta$
decrease.  Their extrema occur near
$10^3$--$2\times10^3$ training steps, coincident with the transient
maximum of the memory self-energy.  Subsequent optimization partially
reverses this concentration, producing a broader multiscale
circulation spectrum while preserving a substantial low-frequency
component.

These quantities characterize the organization of eigenvalue angles
in spectral space and should not be identified directly with
dynamical phase locking of mode trajectories.  Rather, they show that
the characteristic circulation increments of the collective modes
become transiently concentrated toward the low-frequency sector
during the same training regime in which the relaxation dynamics
undergoes maximal memory renormalization.

\begin{figure*}[t]
\centering
\includegraphics[width=1.0\textwidth, trim=0cm 0.7cm 0cm 0cm]{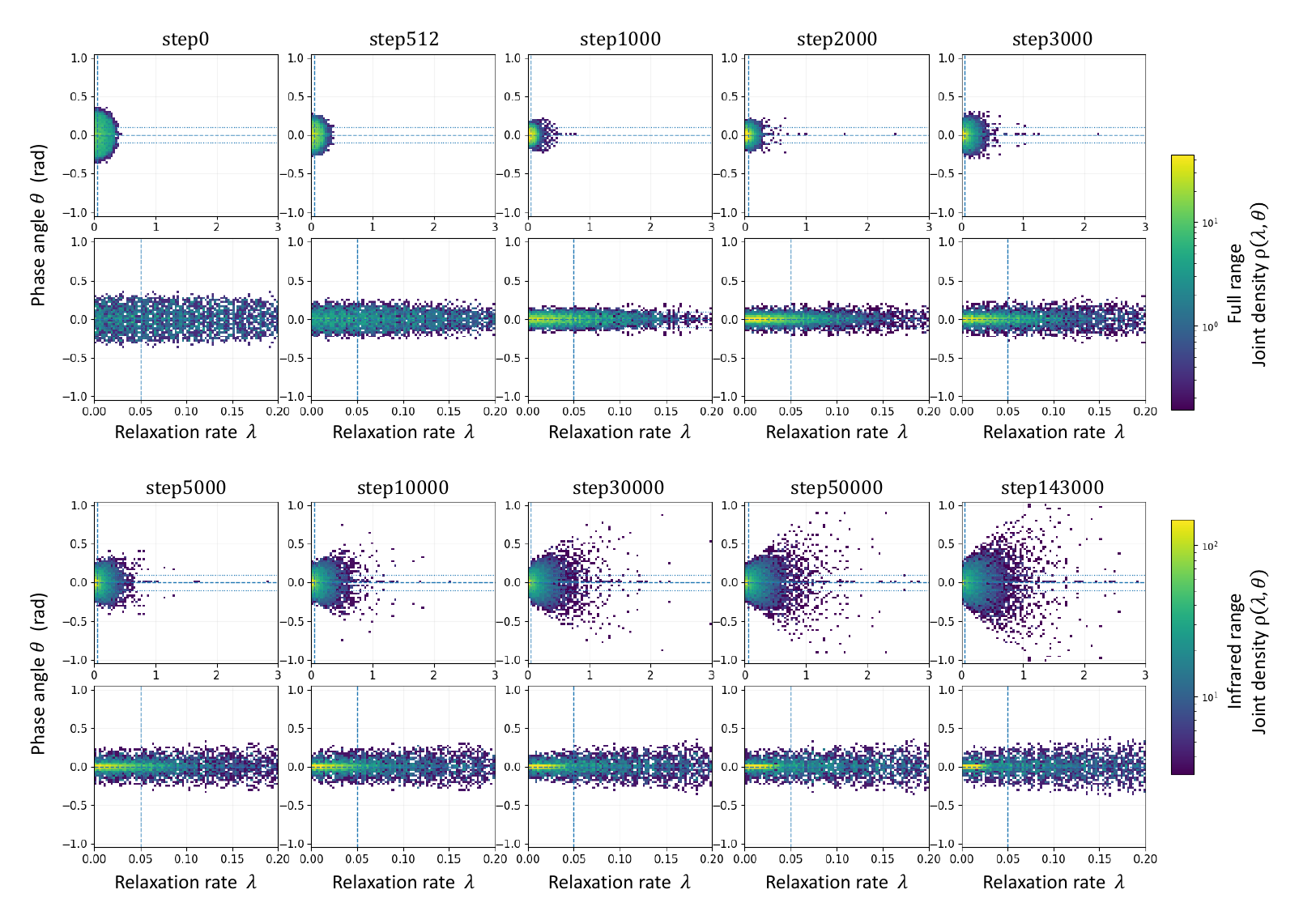}
\caption{
Joint distribution of the relaxation rates and phase angles of the
complex collective Jacobian spectrum during Transformer learning.
The upper panels show the global joint spectrum
$\rho(\lambda,\theta)$,
whereas the lower panels present enlarged views of the infrared region
near the spectral origin.
Initially, the collective spectrum occupies a relatively broad region
surrounding the infrared sector.
During the early stage of optimization, however, the spectrum rapidly
contracts toward
$(\lambda,\theta)\simeq(0,0)$,
revealing the simultaneous accumulation of slow relaxation modes and
low-frequency circulation modes.
The strongest infrared concentration occurs near training steps
$10^3$--$2\times10^3$,
coincident with the transient maximum of the memory self-energy, the
strongest infrared enhancement of the relaxation spectrum, and the
maximum temporal organization of the circulation sector.
At later stages of learning, the infrared core remains clearly
preserved while the surrounding spectrum gradually broadens,
producing a coherent infrared collective core embedded within a broad
multiscale distribution of collective modes.
}
\label{fig:joint_spectrum}
\end{figure*}

The relation between the two spectral sectors is displayed directly in
Fig.~\ref{fig:joint_spectrum}, which presents the joint distribution
\begin{equation}
\rho(\lambda,\theta).
\end{equation}
The upper panels show the complete complex spectrum, while the lower
panels enlarge the infrared region.  At initialization, the spectrum
already contains a broad population of slow modes, but the joint
distribution remains comparatively diffuse.  During early training,
spectral weight becomes strongly concentrated near
\begin{equation}
(\lambda,\theta)
\simeq
(0,0),
\end{equation}
with the strongest concentration again occurring near
$10^3$--$2\times10^3$ steps.

The joint spectrum therefore provides a direct spectral representation
of the simultaneous organization of long relaxation times and small
circulation frequencies.  Importantly, the late-stage spectrum does
not collapse onto a single mode or circulation scale.  Instead, a
high-density infrared core remains embedded within a broad
distribution extending toward larger relaxation rates and phase
angles.  The learned Transformer thus preserves a multiscale complex
spectrum while maintaining substantial spectral weight in the
long-lived, low-frequency sector.

Within Cognitive Field Theory, this organization is consistent with a
memory-dressed collective field operating in a protected near-critical
regime.  The experimentally observed quantity is the organization of
the complex Jacobian spectrum itself; its interpretation as collective
temporal organization follows from the correspondence between
eigenvalue angle and circulation frequency established above.

\begin{figure*}[t]
\centering
\includegraphics[width=1.0\textwidth, trim=0cm 8cm 0cm 0cm]{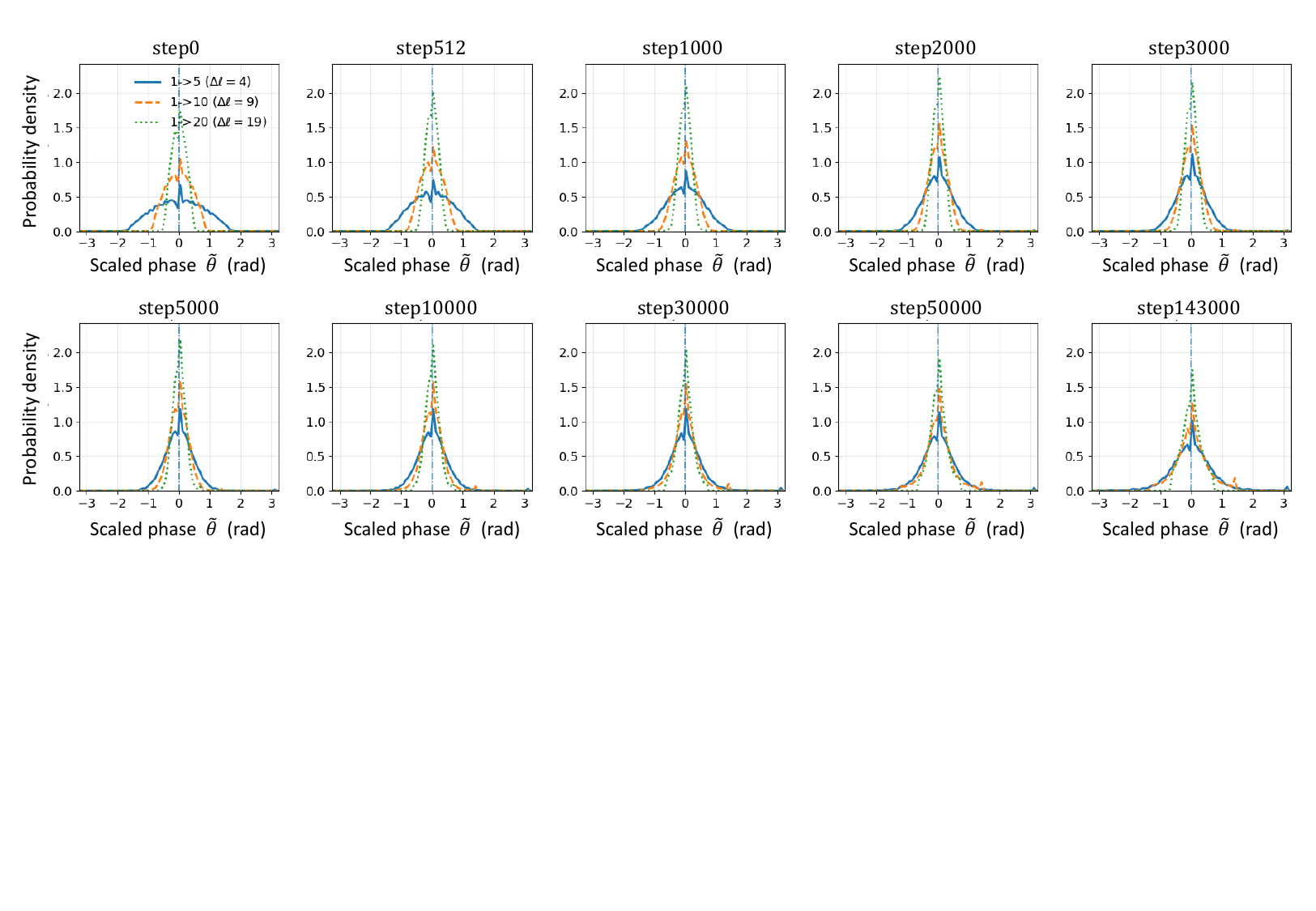}
\caption{
Robustness of the circulation-sector organization with respect to
Transformer propagation depth.
The phase-angle distributions are shown for propagation intervals
$1\!\rightarrow\!5$,
$1\!\rightarrow\!10$,
and
$1\!\rightarrow\!20$,
after rescaling all phase angles to the same reference propagation
length.
Across all propagation depths, the phase distributions exhibit the
same qualitative evolution during learning.
Initially the circulation spectrum is broadly distributed, followed by
a rapid contraction toward the low-frequency sector during the early
stage of optimization.
The strongest localization occurs near training steps
$10^3$--$2\times10^3$,
coincident with the transient maximum of the memory self-energy and
the strongest infrared enhancement of the relaxation spectrum.
At later stages, the phase distributions gradually broaden while
preserving a pronounced low-frequency core centered near
$\theta\simeq0$.
}
\label{fig:phase_depth_scaling}
\end{figure*}

Finally, Fig.~\ref{fig:phase_depth_scaling} examines whether this
circulation-sector organization is restricted to a particular
propagation interval.  Phase-angle distributions are measured for
propagation intervals
$1\!\rightarrow\!5$,
$1\!\rightarrow\!10$, and
$1\!\rightarrow\!20$ and rescaled to the same reference propagation
length.

Although a simple linear rescaling does not produce an exact collapse,
all propagation intervals exhibit the same qualitative training
dependence: an initially broad phase spectrum, rapid concentration
toward $\theta\simeq0$ during early optimization, and subsequent
broadening around a persistent low-frequency sector.  The strongest
concentration again occurs near the training stage associated with
the transient maximum of the memory self-energy.

The circulation-sector reorganization is therefore not restricted to
a particular pair of adjacent Transformer layers but persists across
both short- and long-range propagation.  Together with the relaxation
measurements, these results show that Transformer learning reorganizes
the full complex collective spectrum while preserving a multiscale
infrared structure across network depth.

%===========================================
\section{IX. Discussion}

The present work set out to determine whether the collective
observables proposed in Cognitive Field Theory could be identified
directly in Transformer dynamics.
Rather than analyzing language performance or downstream benchmark
accuracy, we investigated the collective relaxation dynamics of large
language models by measuring the Jacobian spectrum throughout training,
network depth, and model scale.
The results demonstrate that the principal infrared observables
predicted by Cognitive Field Theory are naturally realized in the
collective dynamics of Transformers and are systematically reorganized
during learning.

The measurements reveal a remarkably coherent physical picture.
Learning progressively reorganizes the complex collective spectrum.
Its relaxation sector redistributes spectral weight across slow
relaxation modes, producing infrared reorganization of the time-scale
density of states, non-monotonic evolution of the memory self-energy,
renormalization of the cognitive forgetting gap, and scale-free
long-memory kernels.
The circulation sector simultaneously undergoes transient
low-frequency organization during learning.
The same organizational sequence is reproduced consistently throughout
training, across Transformer depth, and over nearly two orders of
magnitude in model size.
Prompt-resolved and token-subspace analyses further demonstrate that
distinct local measurements recover the same coarse-grained TDOS and
that the spectrum converges as the observed state space is enlarged,
establishing the TDOS as a reproducible macroscopic dynamical
quantity.  Together with the common low-$\lambda$ scaling observed
across these measurements, these results support the infrared
fixed-point interpretation.

The most significant experimental observation is the existence of a
transient maximum of the memory self-energy during training.
Had the self-energy increased monotonically, the results could simply
have been interpreted as a gradual strengthening of long-term memory.
Instead, every investigated Transformer exhibits a well-defined
transient maximum of the memory self-energy, consistent with a
critical formation stage followed by relaxation toward a metastable
near-critical operating regime.
Within Cognitive Field Theory, this transient maximum corresponds to
the smallest cognitive forgetting gap,
\(
r_{\rm cog}
=
r-\Sigma(0),
\)
and therefore to the largest collective susceptibility,
\(
\chi(0)
\propto
r_{\rm cog}^{-1}.
\)

The measurements therefore suggest that Transformer learning does not
approach its final operating state monotonically.
Rather, it first passes through a transient critical formation process
before stabilizing into a protected near-critical regime.
This dynamical scenario emerges directly from the experimentally
measured complex collective spectrum.

An equally important observation is that the infrared organization
remains remarkably robust throughout optimization.
Although the spectral weight of slow relaxation modes changes
substantially during learning, the measured TDOS remains nearly flat
in the infrared, while the corresponding memory kernel consistently
follows the universal long-memory scaling,
\(
K(t)\sim1/t.
\)
This robustness extends beyond the infrared scaling itself.
Independent prompt ensembles and different retained token-subspace
dimensions recover the same normalized infrared TDOS, indicating that
the observed universality emerges from the collective infrared sector
rather than from prompt-specific microscopic dynamics.

The preservation of the infrared critical exponent indicates that
optimization reorganizes the population of slow relaxation modes
without changing the underlying universality class of the collective
dynamics.
Learning therefore strengthens collective memory primarily through
infrared spectral redistribution rather than by altering the
fundamental scaling structure of the relaxation spectrum.

The spatial analysis further demonstrates that the cognitive field is
distributed throughout the Transformer hierarchy rather than localized
within a particular hidden layer.
Different layers exhibit different degrees of memory self-energy
renormalization while preserving the same scale-free infrared
organization.
The cognitive field therefore appears as a distributed infrared
slow-mode manifold extending across the entire Transformer hierarchy,
with different layers contributing different strengths of collective
memory while remaining within the same collective dynamical framework.

Perhaps the most striking result is that the same infrared
organizational sequence is reproduced across all investigated Pythia
models spanning from 70M to 1.4B parameters.
Infrared spectral reorganization, transient enhancement
of the memory self-energy, corresponding suppression of the cognitive
forgetting gap within Cognitive Field Theory, scale-free memory
kernels, and metastable near-critical operation all emerge consistently despite substantial differences in
network size.
These observations indicate that infrared organization is not an
architecture-specific property of an individual Transformer model but a
general dynamical feature of Transformer learning.

From a broader perspective, the present results establish a direct
connection between modern deep learning and nonequilibrium many-body
physics.
Concepts originally developed in condensed-matter field theory—
collective modes, self-energy renormalization, critical
susceptibility, universality, and infrared organization—provide a
natural language for describing the emergence of macroscopic cognitive
dynamics in Transformer models.
The appearance of the same organizational principles in artificial
neural systems suggests that these concepts are not restricted to
physical materials but may represent general principles governing
collective information processing in complex dynamical systems.

More generally, the emergence of Transformer-based intelligent systems
provides, for the first time, a quantitative experimental platform on
which collective cognitive dynamics can be measured directly rather
than inferred solely from external behavior.
The present work demonstrates that collective observables—including
the time-scale density of states and memory self-energy—can be
measured systematically, while the corresponding evolution of the
cognitive forgetting gap and collective susceptibility can be
interpreted within a unified field-theoretic framework.

An important direction for future investigation is to establish a
direct connection between the collective relaxation modes identified
here and the autoregressive inference process of Transformer models.
Because the hidden-state representation can be expanded in the
eigenvector basis of the measured Jacobian, the contribution of each
collective relaxation mode to the output logits and the resulting
softmax probability distribution may be analyzed quantitatively.
Such a mode-resolved description could provide a direct dynamical
picture of how collective relaxation modes organize token generation
during inference, thereby connecting the infrared collective dynamics
identified in the present work with the observable reasoning process
of large language models.

The complete experimentally observed sequence,
\begin{equation}
\rho(\lambda)
\rightarrow
K(t)
\rightarrow
\Sigma(0)
\rightarrow
r_{\rm cog}
\rightarrow
\chi(0),
\end{equation}
together with the convergence of the TDOS measured from distinct
local Jacobians toward the same macroscopic relaxation spectrum,
provides quantitative evidence that macroscopic cognitive organization
emerges through universal infrared collective dynamics.
Whether the same collective observables govern biological neural
systems remains an important question for future investigation.
If similar infrared organization is found in biological cognition, the
present framework may provide a common physical description of
collective intelligence across both artificial and natural cognitive
systems.

\section{X. Conclusion}

In the present work, we investigated whether the collective
observables predicted by Cognitive Field Theory can be identified
directly in Transformer dynamics.
Using layer Jacobians measured throughout Transformer training,
prompt ensembles, network depth, and model scale, we reconstructed
the complex collective spectrum and evaluated its relaxation-sector
observables, including the time-scale density of states, memory
self-energy, memory kernel, and infrared critical exponent, together
with the corresponding evolution of the cognitive forgetting gap and
collective susceptibility within Cognitive Field Theory.

The measurements reveal that Transformer learning systematically
reorganizes the complex collective spectrum while preserving an
approximately flat infrared TDOS.
Its relaxation sector exhibits substantial redistribution of slow-mode
spectral weight, non-monotonic evolution of the memory self-energy,
and robust long-memory dynamics, while its circulation sector
exhibits transient low-frequency organization during learning.
Most importantly, all investigated models exhibit a transient maximum
of the memory self-energy before relaxing toward a protected
metastable near-critical operating regime, consistent with the
critical formation scenario in Cognitive Field Theory.
Despite substantial spectral reorganization, the resulting memory
kernel consistently exhibits universal long-memory scaling,
approximately following
\(K(t)\sim1/t\),
across training, prompt ensembles, network depth, and model scale.

Furthermore, prompt-resolved and token-subspace measurements
demonstrate that TDOS distributions obtained from distinct local
Jacobians converge toward a common macroscopic relaxation spectrum,
establishing the TDOS as a reproducible collective dynamical quantity.
Together with the persistent low-$\lambda$ scaling observed across
these measurements, this convergence supports the infrared
fixed-point organization of the relaxation sector.

Taken together, these results provide quantitative experimental
evidence that large language models possess measurable collective
observables naturally described within a nonequilibrium
field-theoretic framework.
The observed convergence of relaxation spectra measured from distinct
local Jacobians toward a common macroscopic TDOS, together with the
universal long-memory kernel and the reproducibility of the same
infrared scaling across training, prompt ensembles, network depth,
and Transformer model scales, identifies infrared organization of the
collective spectrum as a robust dynamical principle of Transformer
models.
The experimentally accessible collective observables and measurement
protocol introduced here establish a quantitative framework for future
investigations of universal collective dynamics in artificial neural
networks and, potentially, biological cognitive systems.

\vspace{6pt}
\emph{Acknowledgements}---This work was partially supported by the Institute of Information \& Communications Technology Planning \& Evaluation (IITP) grant 
funded by the Korea government (MSIT) (IITP-RS-2025-02214780).

The author acknowledges the support of ChatGPT (GPT-5, OpenAI) for assistance in literature review and conceptual structuring during early development.

\clearpage
\appendix

\renewcommand{\thefigure}{A\arabic{figure}}
\renewcommand{\theequation}{A\arabic{equation}}

\setcounter{figure}{0}
\setcounter{equation}{0}

\vspace*{1.5cm}
{\centering\large\bfseries Supplementary Materials\par}
\vspace{1.0cm}

\section{Appendix A: Relation between the Jacobian Edge Spectrum and the Infrared TDOS}
\label{app:jacobian_edge_tdos}

In the main text, the time-scale density of states (TDOS) is defined
from the relaxation rates of the linearized Transformer dynamics.
Here we derive the direct relation between the spectral density of the
Jacobian eigenvalues near the marginal-stability boundary and the
infrared behavior of the TDOS.

Let $\mu_\alpha$ denote an eigenvalue of the local propagation
Jacobian.  For a stable mode, its magnitude satisfies
\begin{equation}
r_\alpha
\equiv
|\mu_\alpha|
<1 ,
\end{equation}
and the corresponding relaxation rate is defined by
\begin{equation}
\lambda_\alpha
=
-\ln r_\alpha .
\label{eq:app_lambda_from_r}
\end{equation}
The infrared limit of the relaxation spectrum therefore corresponds
directly to the marginal edge of the Jacobian spectrum,
\begin{equation}
\lambda\rightarrow0^+
\qquad\Longleftrightarrow\qquad
r\rightarrow1^- .
\label{eq:app_ir_marginal_mapping}
\end{equation}

Let $p(r)$ denote the normalized density of stable Jacobian
eigenvalue magnitudes,
\begin{equation}
\int_0^1 dr\,p(r)=1 .
\end{equation}
Conservation of spectral weight under the transformation
$r=e^{-\lambda}$ requires
\begin{equation}
\rho(\lambda)\,d\lambda
=
p(r)\,|dr| .
\end{equation}
Since
\begin{equation}
r=e^{-\lambda},
\qquad
\left|
\frac{dr}{d\lambda}
\right|
=
e^{-\lambda},
\end{equation}
the TDOS is related exactly to the radial Jacobian density by
\begin{equation}
\rho(\lambda)
=
p(e^{-\lambda})e^{-\lambda}.
\label{eq:app_tdos_jacobian_density}
\end{equation}

Equation~(\ref{eq:app_tdos_jacobian_density}) directly relates the
infrared TDOS to the spectral structure of the Jacobian near
$|\mu|=1$.
For $\lambda\ll1$,
\begin{equation}
e^{-\lambda}
=
1-\lambda+\mathcal O(\lambda^2),
\end{equation}
so that
\begin{equation}
\rho(\lambda)
=
p(1-\lambda+\mathcal O(\lambda^2))
\left[
1-\lambda+\mathcal O(\lambda^2)
\right].
\label{eq:app_tdos_ir_expansion}
\end{equation}

Suppose that the Jacobian magnitude density near the marginal edge has
the asymptotic form
\begin{equation}
p(r)
\sim
C(1-r)^\alpha,
\qquad
r\rightarrow1^- .
\label{eq:app_jacobian_edge_powerlaw}
\end{equation}
Using
\begin{equation}
1-e^{-\lambda}
=
\lambda+\mathcal O(\lambda^2),
\end{equation}
Eq.~(\ref{eq:app_tdos_jacobian_density}) gives
\begin{equation}
\rho(\lambda)
\sim
C
\left(
1-e^{-\lambda}
\right)^\alpha
e^{-\lambda}
\sim
C\lambda^\alpha,
\qquad
\lambda\rightarrow0^+ .
\label{eq:app_edge_to_ir_scaling}
\end{equation}
Thus the edge exponent of the stable Jacobian spectrum is inherited
directly by the infrared TDOS,
\begin{equation}
\beta=\alpha .
\label{eq:app_beta_alpha}
\end{equation}

Of particular interest is a regular Jacobian density with finite
nonzero spectral weight at the marginal edge,
\begin{equation}
p(r\rightarrow1^-)
\rightarrow
p_0,
\qquad
0<p_0<\infty .
\label{eq:app_regular_edge}
\end{equation}
In this case,
\begin{equation}
\rho(\lambda\rightarrow0^+)
\rightarrow
p_0,
\label{eq:app_flat_ir_tdos}
\end{equation}
corresponding to the marginal infrared exponent
\begin{equation}
\beta=0.
\end{equation}
A flat infrared TDOS therefore does not require a singular Jacobian
eigenvalue density.  It follows naturally whenever the stable
Jacobian spectrum approaches the marginal boundary with a finite,
nonvanishing density.

This relation also provides a possible structural interpretation of
the infrared spectrum in residual architectures.  A residual
Transformer block has the schematic form
\begin{equation}
x_{\ell+1}
=
x_\ell
+
F_\ell(x_\ell),
\label{eq:app_residual_block}
\end{equation}
with Jacobian
\begin{equation}
J_\ell
=
I
+
\frac{\partial F_\ell}{\partial x_\ell}.
\label{eq:app_residual_jacobian}
\end{equation}
The identity contribution provides a natural near-identity component
of the local propagation operator.  Attention, feature-space mixing,
normalization, nonlinear transformation, and learned weights then
modify and split this spectrum, producing a distribution of
collective propagation factors around the near-marginal region.

The residual structure alone does not imply a flat infrared TDOS:
the resulting edge density $p(r)$ depends on the full dynamical
operator and its operating state.  Nevertheless,
Eqs.~(\ref{eq:app_tdos_jacobian_density})--%
(\ref{eq:app_flat_ir_tdos}) show that any finite spectral density
maintained near $|\mu|=1$ is mapped directly onto a finite density of
arbitrarily slow relaxation modes.  Residual propagation is therefore
structurally compatible with the marginal infrared spectrum observed
in the Transformer dynamics.

More generally, the relation
\begin{equation}
p(r)
\sim
(1-r)^\alpha
\qquad\Longleftrightarrow\qquad
\rho(\lambda)
\sim
\lambda^\alpha
\label{eq:app_edge_ir_correspondence}
\end{equation}
shows that the infrared TDOS can be interpreted directly as the
marginal-edge structure of the underlying propagation spectrum.
The approximately flat TDOS discussed in the main text corresponds
to the regular case in which the stable Jacobian spectrum retains
finite spectral weight as $|\mu|\rightarrow1^-$.

%%%%%%%%%%%%%%%%%%%%%%%%%%%%%%%%%%%%%%%%%%%%%%%%%%%%%%%%%%%%%%%%%%%%%%%%%%%%%%
%% Appendix B
%%%%%%%%%%%%%%%%%%%%%%%%%%%%%%%%%%%%%%%%%%%%%%%%%%%%%%%%%%%%%%%%%%%%%%%%%%%%%%
%\clearpage
\renewcommand{\thefigure}{B\arabic{figure}}
\renewcommand{\theequation}{B\arabic{equation}}

\section{Appendix B: Robustness with Respect to Input Sequence Length}

The analyses presented in the main text are performed using the first
eight tokens of a fixed input prompt, corresponding to an
$8192\times8192$ Jacobian for the Pythia-410M model.
To verify that the measured collective observables do not depend on
this particular choice, we repeated the complete analysis using the
first sixteen input tokens, yielding a
$16384\times16384$ Jacobian and the corresponding relaxation spectrum.

Throughout this appendix, the input prompt is
\begin{quote}
``The future of tiny cognitive field networks is robust adaptive
efficient and useful for control tasks''
\end{quote}
and the corresponding representative model output is

\begin{quote}
``The future of tiny cognitive field networks is robust adaptive
efficient and useful for control tasks with uncertainty, such as human
navigation. However we must also consider the challenges to be
overcome by a design that includes an accurate representation of both
time-dependent and stochastic phenomena.''
\end{quote}

The Jacobian is evaluated from the local mapping defined by the first
sixteen input tokens, while the generated continuation is shown only
to illustrate the representative operating state of the trained model.

Figures~B1--B7 summarize the results obtained from the sixteen-token
analysis.
The figures follow the same organization as Figs.~2--9 in the main
text, including the evolution of the TDOS during training, the
population of relaxation modes, the memory self-energy and forgetting
gap, the memory kernel, and the layer-dependent infrared
organization.

The overall behavior is found to be in excellent agreement with the
results presented in the main text.
In particular, Transformer learning again exhibits progressive
infrared accumulation of slow relaxation modes, a transient maximum of
the memory self-energy followed by metastable near-critical
stabilization, robust scale-free memory kernels, and a distributed
layer hierarchy of collective memory organization.

The principal quantitative difference is that increasing the input
sequence length naturally increases the total number of relaxation
modes sampled by the Jacobian.
Consequently, the TDOS becomes smoother and the infrared statistics are
improved, while the measured collective observables and their
qualitative evolution remain unchanged.

These results demonstrate that the infrared organization reported in
the present work is robust with respect to the choice of input
sequence length and therefore reflects an intrinsic property of the
Transformer dynamics rather than an artifact of a particular Jacobian
construction.

\begin{figure*}[t]
\centering
\includegraphics[width=1.0\textwidth, trim=0cm 1.7cm 0cm 0cm]{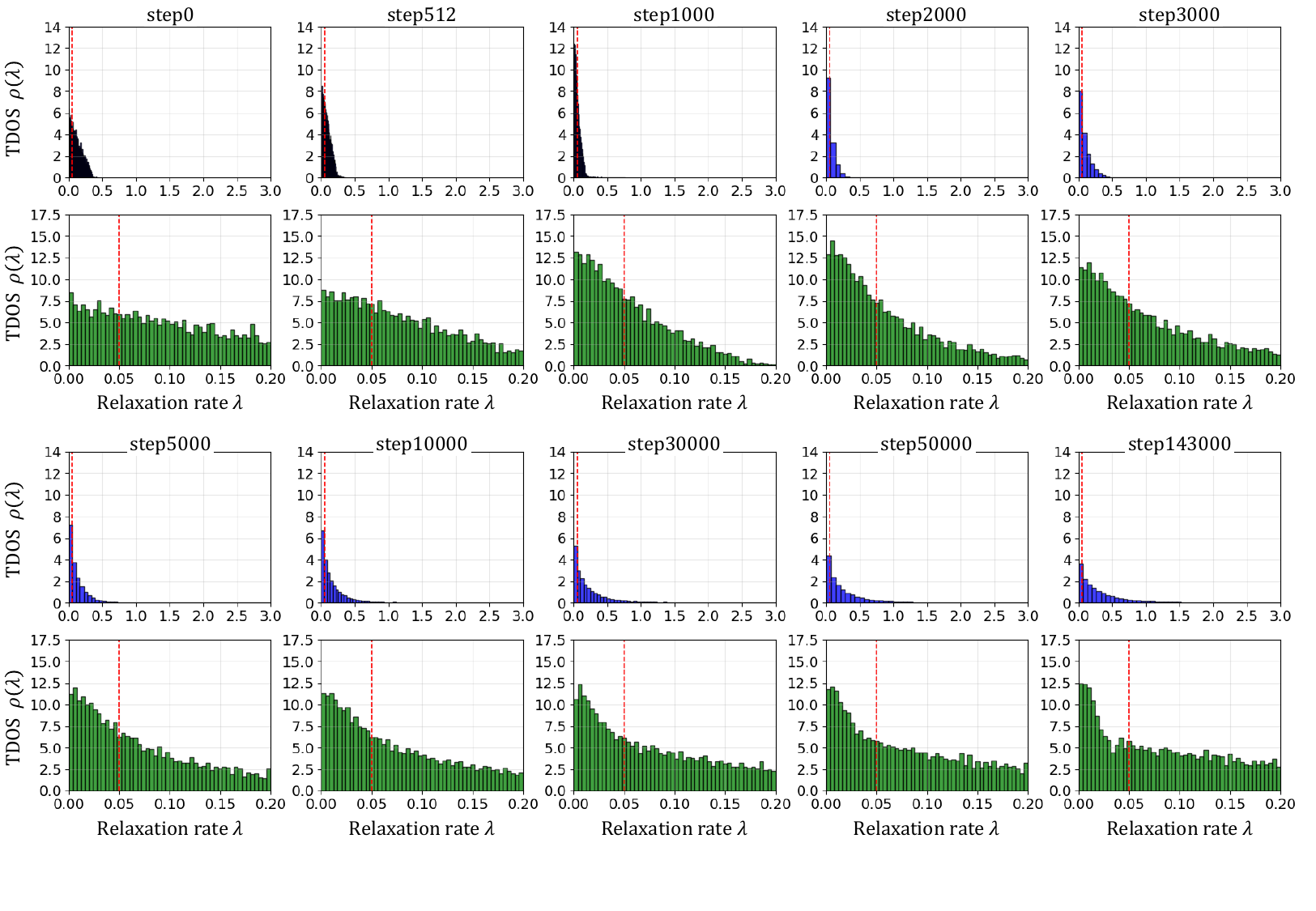}
\caption{
Evolution of the TDOS during Transformer training obtained using a
sixteen-token input sequence for the Pythia-410M model.
The overall infrared spectral reorganization closely follows the
behavior observed in the eight-token analysis presented in the main
text.
}
\label{fig:phase_portrait}
\end{figure*}

\begin{figure*}[t]
\centering
\includegraphics[scale=0.55, trim= 0cm 8cm 9cm 0cm]{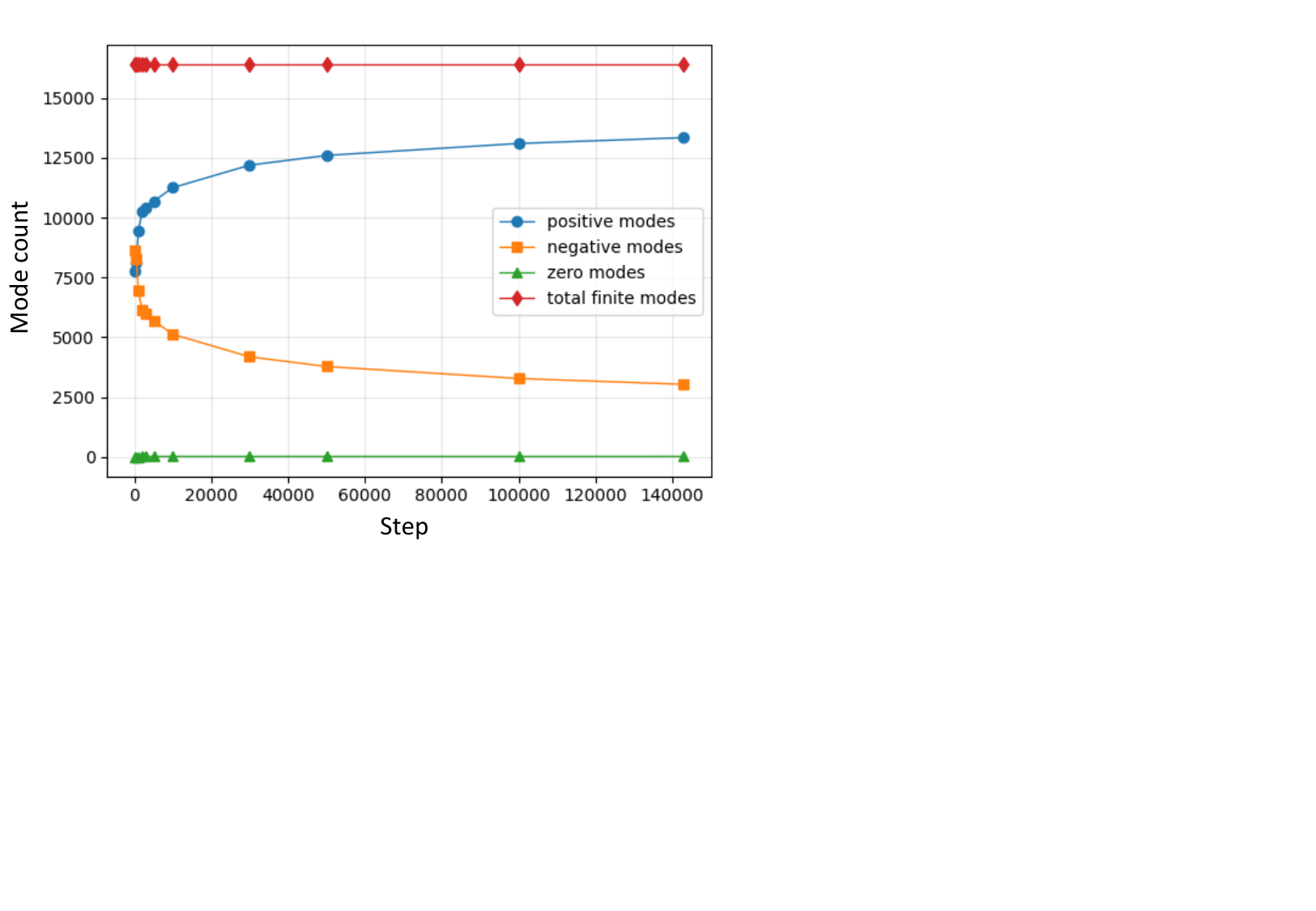}
\caption{
Evolution of the populations of stable, unstable, zero, and finite
relaxation modes obtained from the sixteen-token Jacobian spectrum.
The qualitative evolution agrees with the results presented in the
main text.
}
\label{fig:phase_portrait}
\end{figure*}

\begin{figure*}[t]
\centering
\includegraphics[scale=0.55, trim= 0cm 0.58cm 9cm 0cm]{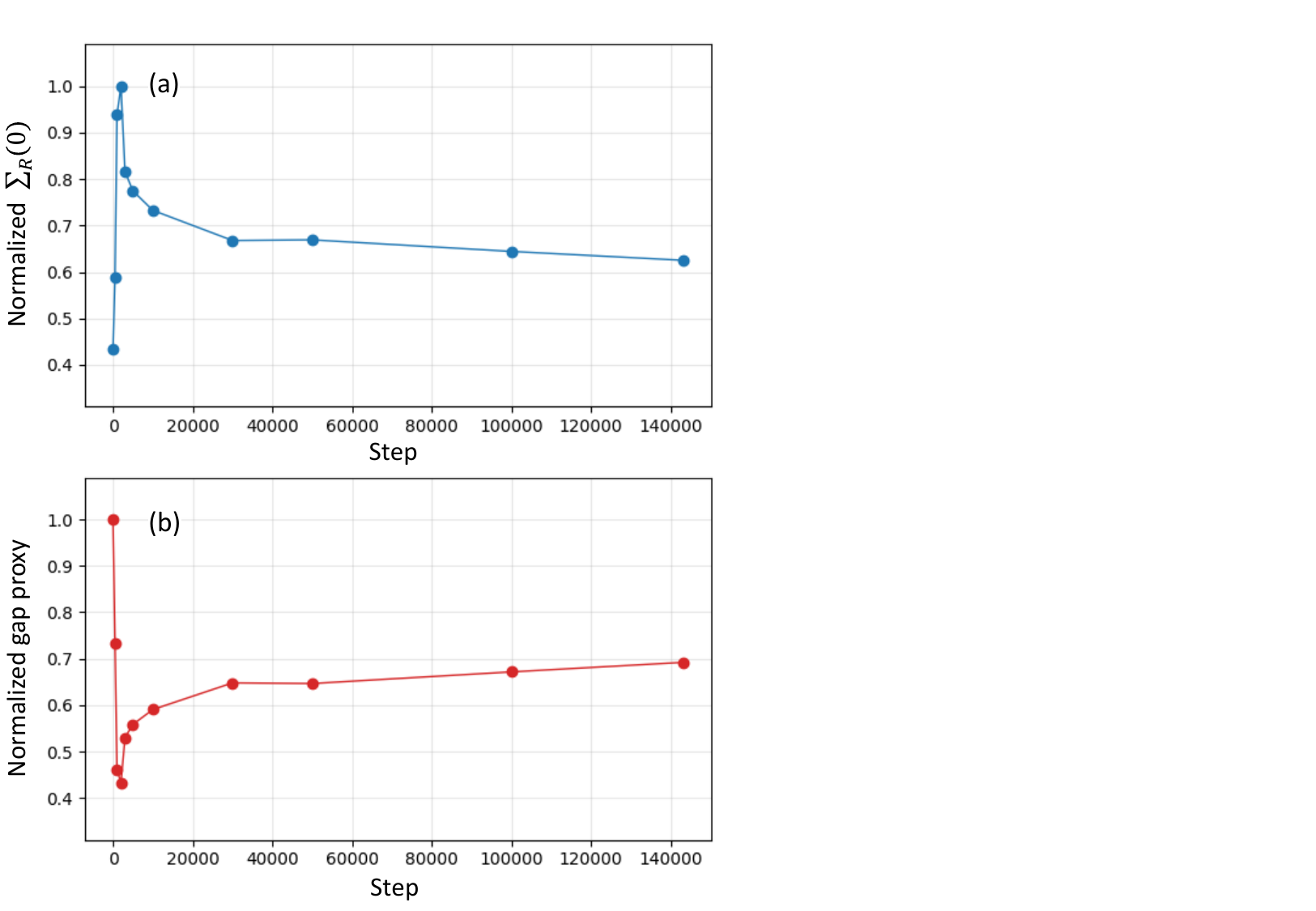}
\caption{
Evolution of the normalized memory self-energy and the corresponding
normalized forgetting-gap proxy obtained using sixteen input
tokens for the Pythia-410M model.
The transient maximum of the memory self-energy and the subsequent
metastable near-critical stabilization are reproduced.
}
\label{fig:phase_portrait}
\end{figure*}

\begin{figure*}[t]
\centering
\includegraphics[scale=0.5, trim= 0.2cm 0.5cm 8cm 0cm]{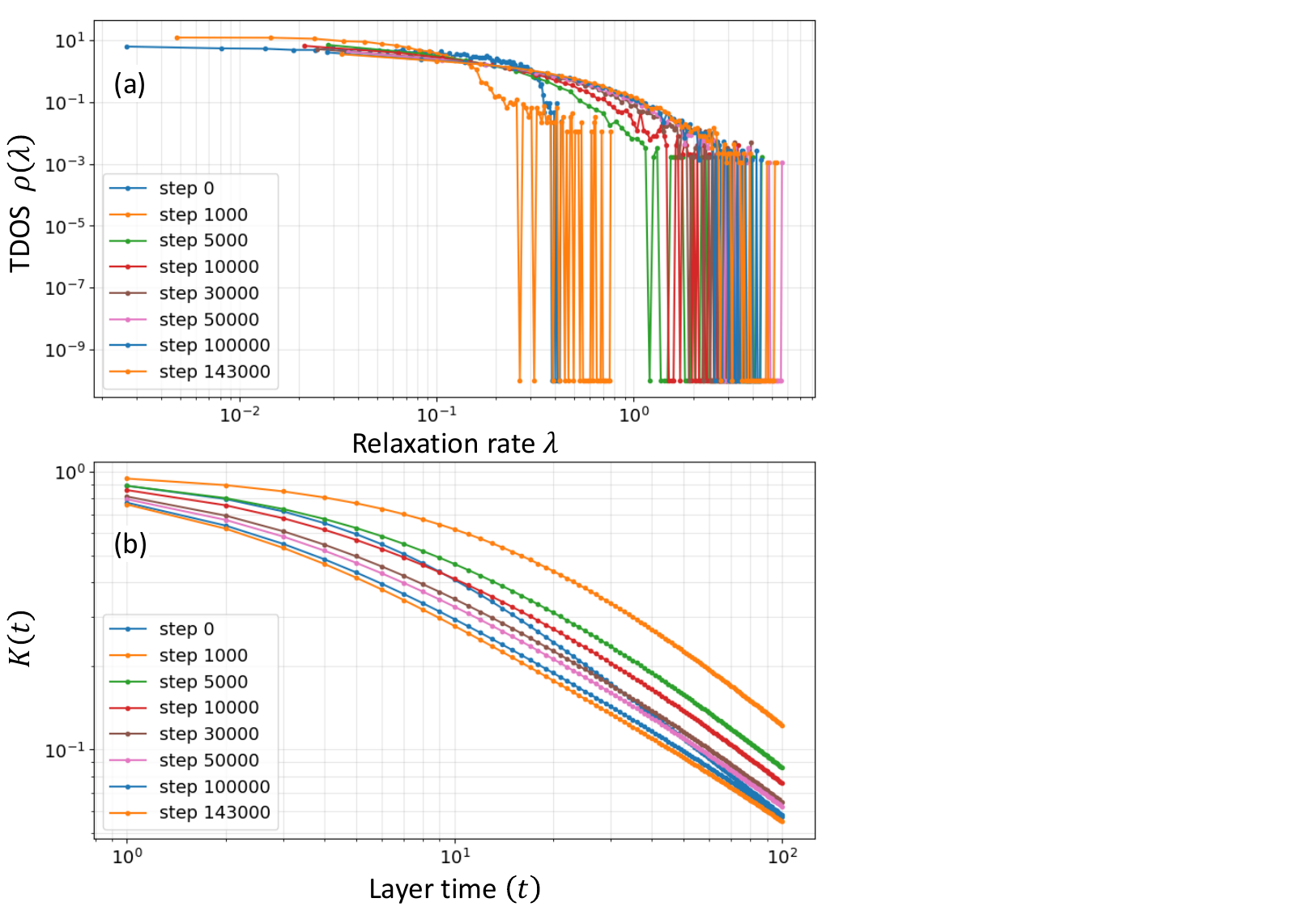}
\caption{
Evolution of the infrared TDOS and the corresponding memory kernel
obtained using sixteen input tokens.
The scale-free long-time memory remains robust against the input
sequence length.
}
\label{fig:phase_portrait}
\end{figure*}

\begin{figure*}[t]
\centering
\includegraphics[width=1.0\textwidth, trim=0cm 10cm 0cm 0cm]{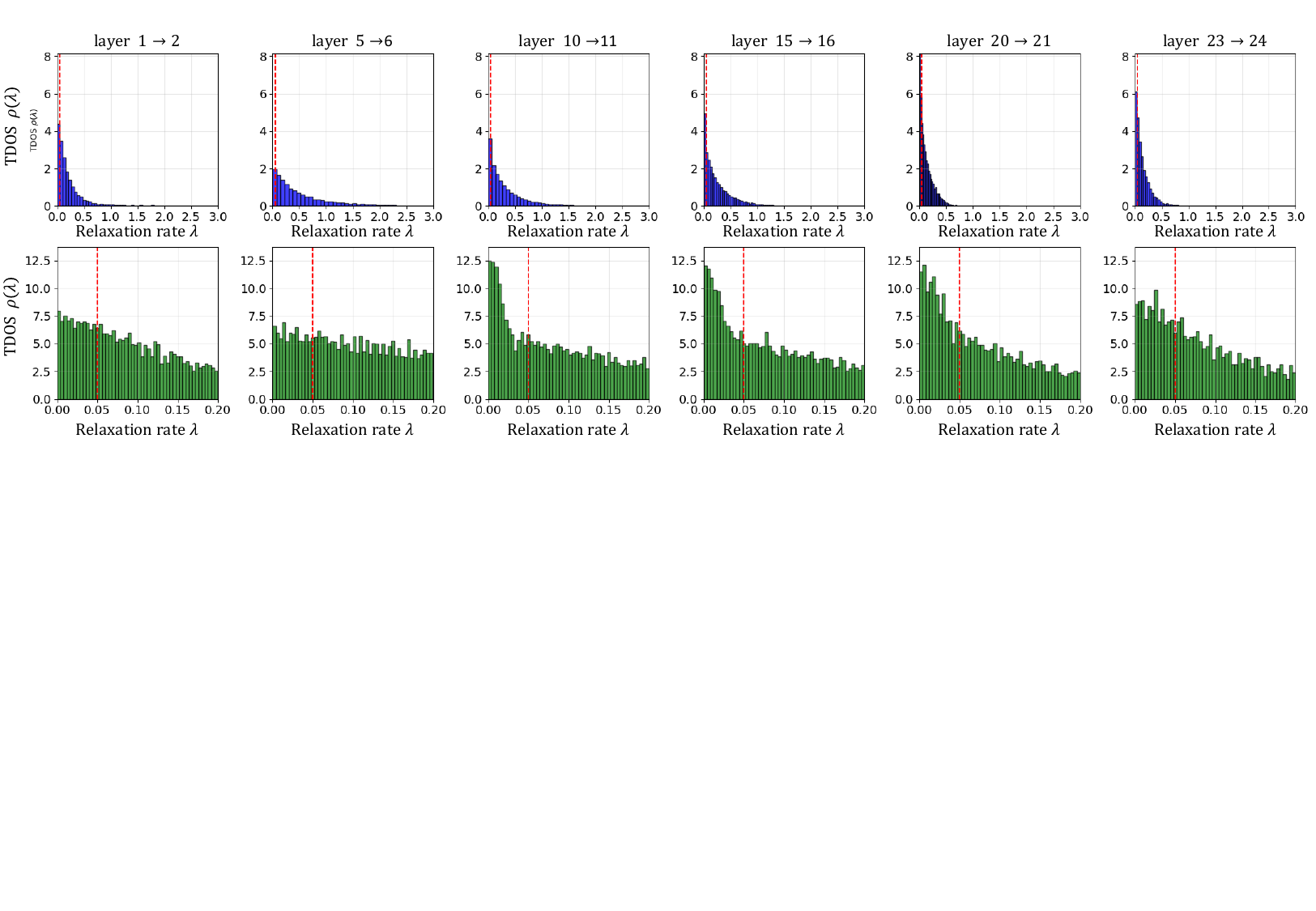}
\caption{
Layer-dependent TDOS obtained from the trained Pythia-410M model using
a sixteen-token input sequence.
Infrared accumulation of slow relaxation modes is observed throughout
the Transformer hierarchy.
}
\label{fig:phase_portrait}
\end{figure*}

\begin{figure*}[t]
\centering
\includegraphics[scale=0.55, trim= 0cm 0.5cm 9cm 0cm]{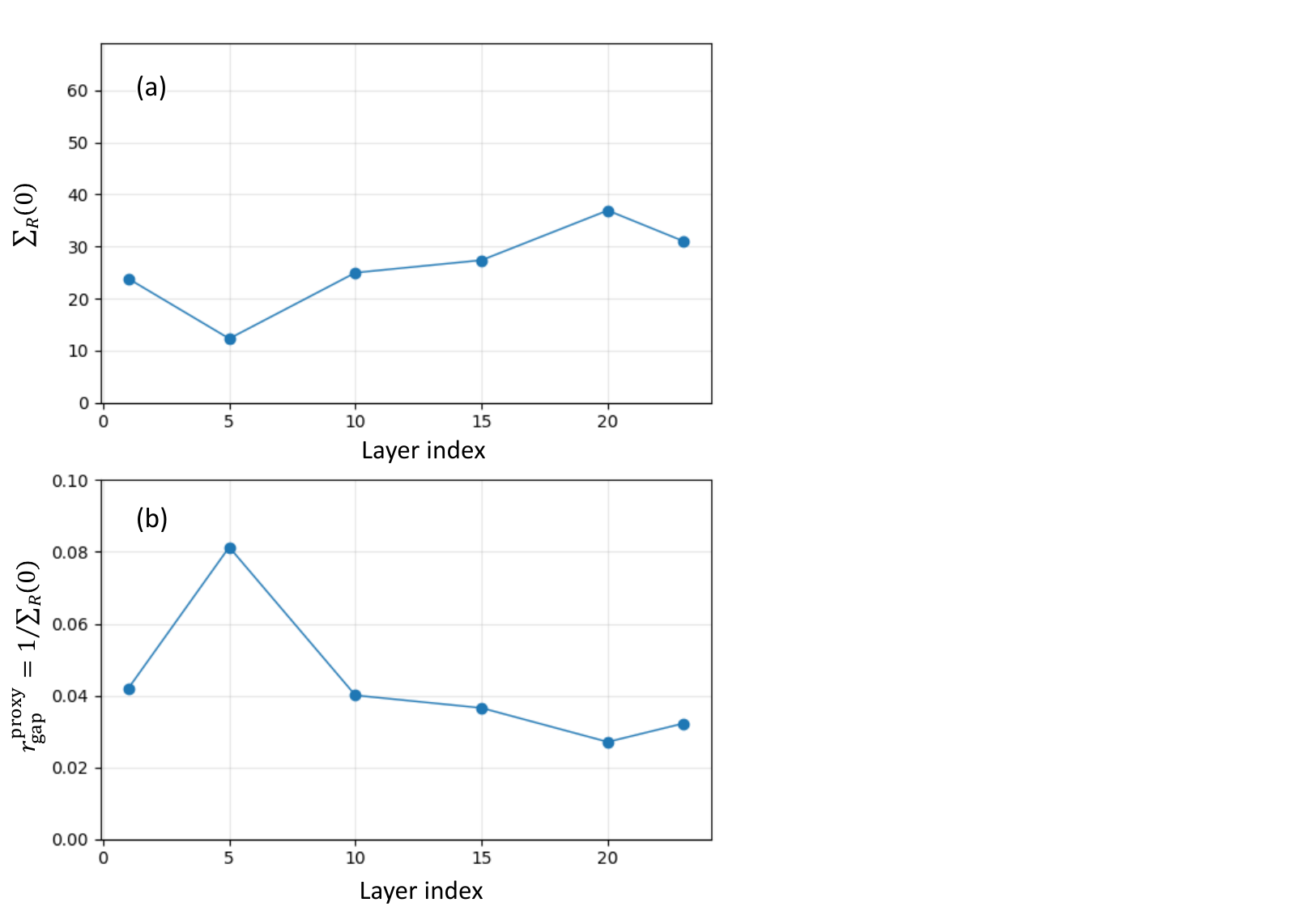}
\caption{
Layer dependence of the static memory self-energy and the
corresponding proxy forgetting gap obtained from the sixteen-token
analysis.
Intermediate and deeper layers exhibit stronger memory
renormalization, consistent with the results in the main text.
}
\label{fig:phase_portrait}
\end{figure*}

\begin{figure*}[t]
\centering
\includegraphics[scale=0.5, trim= 0cm 0.5cm 8cm 0cm]{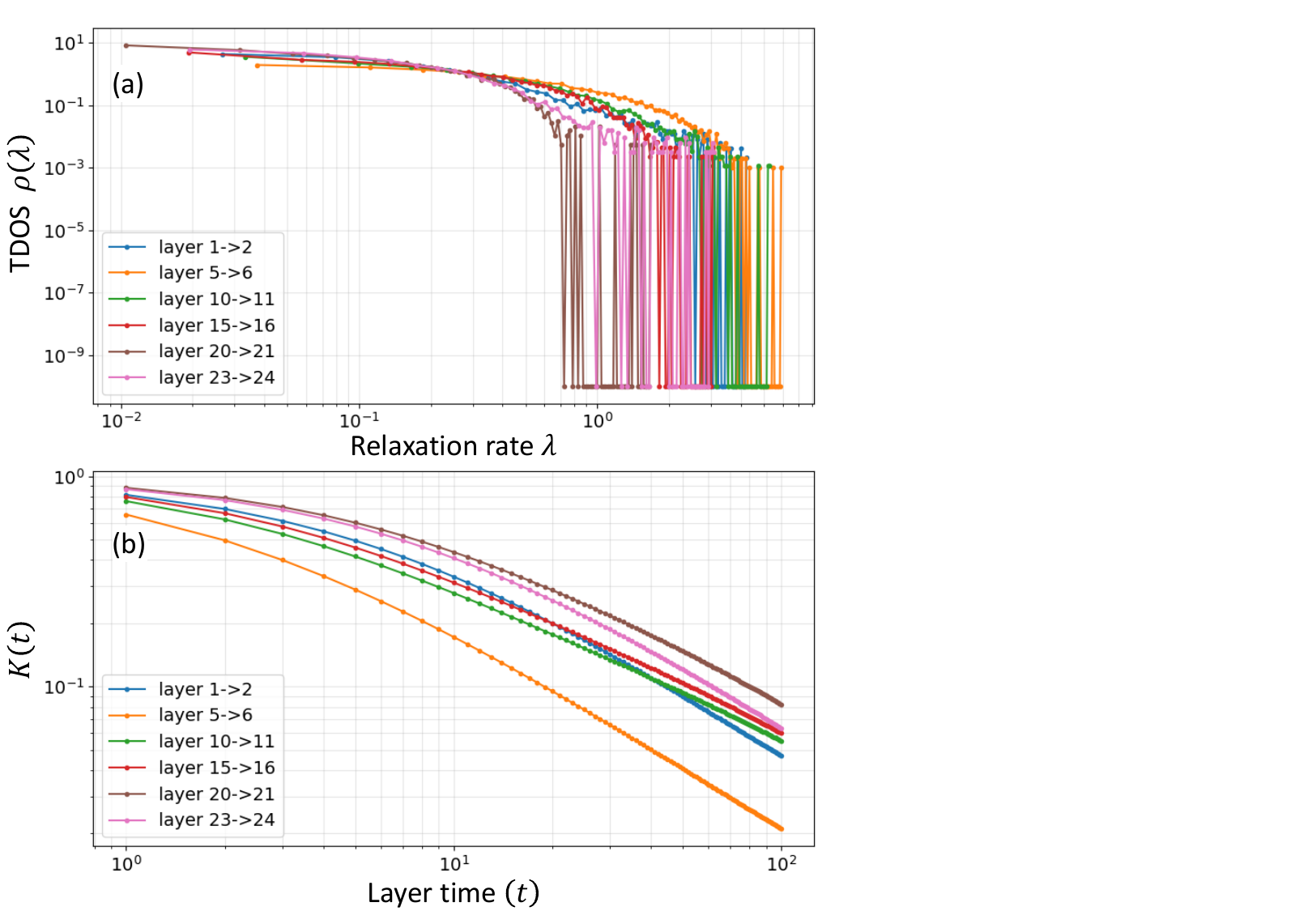}
\caption{
Layer-dependent infrared TDOS and memory kernels obtained using a
sixteen-token input sequence.
The common scale-free infrared organization is preserved throughout
the Transformer hierarchy.
}
\label{fig:phase_portrait}
\end{figure*}

%%%%%%%%%%%%%%%%%%%%%%%%%%%%%%%%%%%%%%%%%%%%%%%%%%%%%%%%%%%%%%%%%%%%%%%%%%%%%%
%% Appendix C
%%%%%%%%%%%%%%%%%%%%%%%%%%%%%%%%%%%%%%%%%%%%%%%%%%%%%%%%%%%%%%%%%%%%%%%%%%%%%%

\clearpage

\renewcommand{\thefigure}{C\arabic{figure}}
\setcounter{figure}{0}

\renewcommand{\thetable}{C\arabic{table}}
\setcounter{table}{0}

\section{Appendix C: Prompt-Ensemble Robustness}

The prompt-ensemble analysis presented in Sec.~IV was performed using
thirty representative prompts spanning diverse semantic domains,
including artificial intelligence, physics, mathematics,
neuroscience, natural language, technology, economics, and everyday
narratives.
The purpose of this prompt set is not to evaluate language
understanding itself, but to test whether the measured time-scale
density of states depends on prompt semantics.

For completeness and reproducibility, the complete prompt ensemble
used throughout the analysis is listed in
Tables~\ref{tab:prompt_list_1} and
\ref{tab:prompt_list_2}.
For each prompt, the Jacobian spectrum was evaluated using the same
Layer~10$\rightarrow$11 mapping of the fully trained Pythia-410M
model.

Figure~C1 presents the detailed prompt-resolved TDOS obtained using
24-token input subspaces, corresponding to the measurement condition
used for the prompt-ensemble analysis in Sec.~IV.
The individual spectra provide an expanded view of the
prompt-to-prompt robustness summarized by the ensemble statistics in
the main text.

To further examine the dependence on the retained state-space
dimension, we repeated the same prompt-ensemble analysis using
16-token and 28-token input subspaces.
For each token-subspace dimension, Jacobian spectra were computed from
the same prompt ensemble, from which the normalized TDOS,
prompt-ensemble averages, and spectral fluctuations were evaluated.

Figures~C2--C5 summarize these additional measurements.
Figures~C2 and C3 present the prompt-resolved TDOS and corresponding
prompt-ensemble statistics obtained using 16-token input subspaces,
while Figs.~C4 and C5 show the corresponding results for 28-token
input subspaces.

Across all three token-subspace dimensions, the measured TDOS exhibits
the same overall spectral organization and strong concentration across
different input prompts.  Increasing the retained token subspace also
extends the reliably resolved spectrum toward smaller relaxation
rates, consistent with the token-subspace convergence reported in
Sec.~IV.  These results confirm that the observed prompt robustness is
not specific to a particular retained token-subspace dimension.

%%%%%%%%%%%%%%%%%%%%%%%%%%%%%%%%%%%%%%%%%%%%%%%%%%%%%%%%%%%%%%%%%%%%%%%%%%%%%%
%% Table C1
%%%%%%%%%%%%%%%%%%%%%%%%%%%%%%%%%%%%%%%%%%%%%%%%%%%%%%%%%%%%%%%%%%%%%%%%%%%%%%

\begin{table*}[t]
\caption{
First fifteen representative prompts (Prompt 1--15) used in the
28-token prompt-ensemble TDOS analysis.
}
\label{tab:prompt_list_1}
\centering
\begin{tabular}{lp{16.5cm}}
\hline\hline
No. & Prompt \\
\hline

1 &
``The future of artificial intelligence depends on reliable reasoning adaptive memory robust planning transparent evaluation and the ability to maintain context across long sequences of information while integrating new evidence correcting earlier mistakes and preserving coherent goals during extended problem solving.'' \\

2 &
``Tiny cognitive field networks may support robust control in changing environments by combining stable internal geometry slow collective modes recursive feedback and adaptive responses to uncertainty while preserving memory across delayed observations and reorganizing behavior when external conditions suddenly change.'' \\

3 &
``Quantum field theory describes collective excitations through effective low energy degrees of freedom whose interactions determine symmetry breaking critical behavior long range correlations and observable macroscopic phases across different materials energy scales and experimentally accessible dynamical regimes.'' \\

4 &
``Machine learning models can reorganize internal representations during training and inference so that distributed features become coordinated across layers tokens tasks and changing contextual demands while preserving useful abstractions and suppressing unstable or irrelevant internal fluctuations.'' \\

5 &
``Once upon a time a curious child discovered a hidden path through the forest and followed it toward an ancient village surrounded by rivers mountains and forgotten stories where old travelers spoke of a library buried beneath the central stone tower.'' \\

6 &
``The global economy changes when technology productivity institutions human expectations energy systems and international trade interact across many spatial temporal political and financial scales creating feedback loops that influence employment investment inflation innovation and long term social stability.'' \\

7 &
``Neural networks learn distributed representations from many examples of structured data while optimization gradually reshapes internal geometry attention patterns feature hierarchies and collective dynamical responses that support generalization transfer contextual reasoning and adaptation to unfamiliar inputs.'' \\

8 &
``Mathematics provides a precise language for describing patterns symmetry transformation uncertainty geometry dynamics and the hidden relationships connecting apparently different physical and computational systems through shared structures conserved quantities and reusable abstract principles.'' \\

9 &
``Tomorrow the weather may change rapidly as a cold front crosses the region bringing stronger winds lower temperatures heavy rain unstable clouds and difficult travel conditions especially near coastal roads mountain passes and low lying areas vulnerable to flooding.'' \\

10 &
``Biological evolution produces complex adaptive systems through variation selection inheritance development ecological interaction and the accumulation of functional structure across extremely long periods of time while organisms continuously modify and respond to their changing environments.'' \\

11 &
``Large language models exhibit surprising abilities when context computation data diversity and optimization are scaled together while preserving coherent representations across many layers and token positions during reasoning generation translation planning and retrieval from long contextual sequences.'' \\

12 &
``Physics attempts to explain how simple microscopic laws generate complex macroscopic behavior through collective organization fluctuations symmetry breaking conservation principles and emergent effective degrees of freedom that govern phases transport memory response and universal critical phenomena.'' \\

13 &
``The little boy walked slowly toward the river while carrying an old wooden box that contained letters photographs maps and a mysterious key from his grandfather together with a handwritten note describing a house that no longer appeared on modern maps.'' \\

14 &
``Deep learning enables computers to recognize patterns in images speech and written language by constructing layered representations that capture local features global structure and contextual relationships while learning invariances useful for prediction classification reconstruction and controlled generation.'' \\

15 &
``Memory is an essential component of reasoning because past information guides present decisions supports future planning stabilizes context and enables coherent behavior over extended time intervals even when relevant evidence is separated by many intermediate computational steps.'' \\

\hline\hline
\end{tabular}
\end{table*}

%%%%%%%%%%%%%%%%%%%%%%%%%%%%%%%%%%%%%%%%%%%%%%%%%%%%%%%%%%%%%%%%%%%%%%%%%%%%%%
%% Table C2
%%%%%%%%%%%%%%%%%%%%%%%%%%%%%%%%%%%%%%%%%%%%%%%%%%%%%%%%%%%%%%%%%%%%%%%%%%%%%%

\begin{table*}[t]
\caption{
Second fifteen representative prompts (Prompt 16--30) used in the
28-token prompt-ensemble TDOS analysis.
}
\label{tab:prompt_list_2}
\centering
\begin{tabular}{lp{16.5cm}}
\hline\hline
No. & Prompt \\
\hline

16 &
``Transformer architectures process token sequences through attention nonlinear transformation residual feedback normalization and learned projection while information is repeatedly reorganized across depth and contextual position to support prediction retrieval composition abstraction and long range dependence.'' \\

17 &
``Natural language processing allows machines to analyze generate summarize translate and reason about human language while preserving semantic relations discourse structure contextual references and communicative intent across documents conversations technical domains and multilingual environments.'' \\

18 &
``Consciousness remains one of the most difficult scientific problems in neuroscience philosophy psychology and artificial intelligence because subjective experience must be connected to measurable physical dynamics distributed neural activity memory integration attention and self related representation.'' \\

19 &
``The universe began in a hot dense state and has expanded for billions of years while matter radiation galaxies stars planets and complex chemical structures gradually formed through gravitational collapse nuclear reactions cooling and large scale cosmic evolution.'' \\

20 &
``Statistical mechanics predicts collective behavior by connecting microscopic states with macroscopic observables through probability distributions entropy fluctuations correlations phase transitions and coarse grained effective descriptions that become increasingly universal near critical regimes.'' \\

21 &
``Scientists discovered that the material changes its properties near a critical temperature where fluctuations grow relaxation slows correlations spread and conventional microscopic descriptions become insufficient to explain the observed response transport and long range collective organization.'' \\

22 &
``Modern computers perform complex calculations by coordinating many layers of hardware software memory communication protocols operating systems compilers numerical libraries and application specific algorithms while managing errors latency energy consumption parallel execution and data movement.'' \\

23 &
``Energy is conserved in an isolated system although it may change from one form to another through motion heat radiation chemical reactions fields and collective excitations while the total remains fixed under time translation symmetry.'' \\

24 &
``Information theory studies communication uncertainty compression coding noise and the fundamental limits of reliable transmission across channels with finite bandwidth resources and imperfect observations while quantifying redundancy entropy mutual information and error correcting capacity.'' \\

25 &
``The human brain contains interconnected neural populations operating across many spatial and temporal scales while sensory memory motor emotional and cognitive processes continuously influence one another through recurrent activity neuromodulation plasticity and coordinated population dynamics.'' \\

26 &
``Critical phenomena emerge when fluctuations become correlated across increasingly large scales causing microscopic differences to become irrelevant while universal infrared structure controls observable collective behavior response functions relaxation laws and scaling relations.'' \\

27 &
``The cat sat quietly beside the window and watched the rain fall outside while distant thunder echoed across the street and evening lights appeared one by one as people hurried home beneath umbrellas and passing cars reflected on the wet road.'' \\

28 &
``Artificial neural systems may develop persistent internal states through recursive dynamical feedback stable attractors slow collective modes contextual gating and learned mechanisms that resist rapid forgetting while remaining flexible enough to incorporate new evidence.'' \\

29 &
``Cognitive dynamics exhibit slow relaxation when many interacting modes approach marginal stability allowing information to persist influence later computation and reorganize the trajectory of ongoing inference across multiple layers tokens and recursively coupled internal representations.'' \\

30 &
``Learning reorganizes internal geometry and changes the spectrum of accessible collective relaxation modes thereby modifying memory stability contextual persistence inference pathways and long range dynamical coordination while driving the system toward a structured near critical regime.'' \\

\hline\hline
\end{tabular}
\end{table*}

\begin{figure*}[t]
\centering
\includegraphics[width=1.0\textwidth, trim=0cm 3.5cm 0cm 0cm]{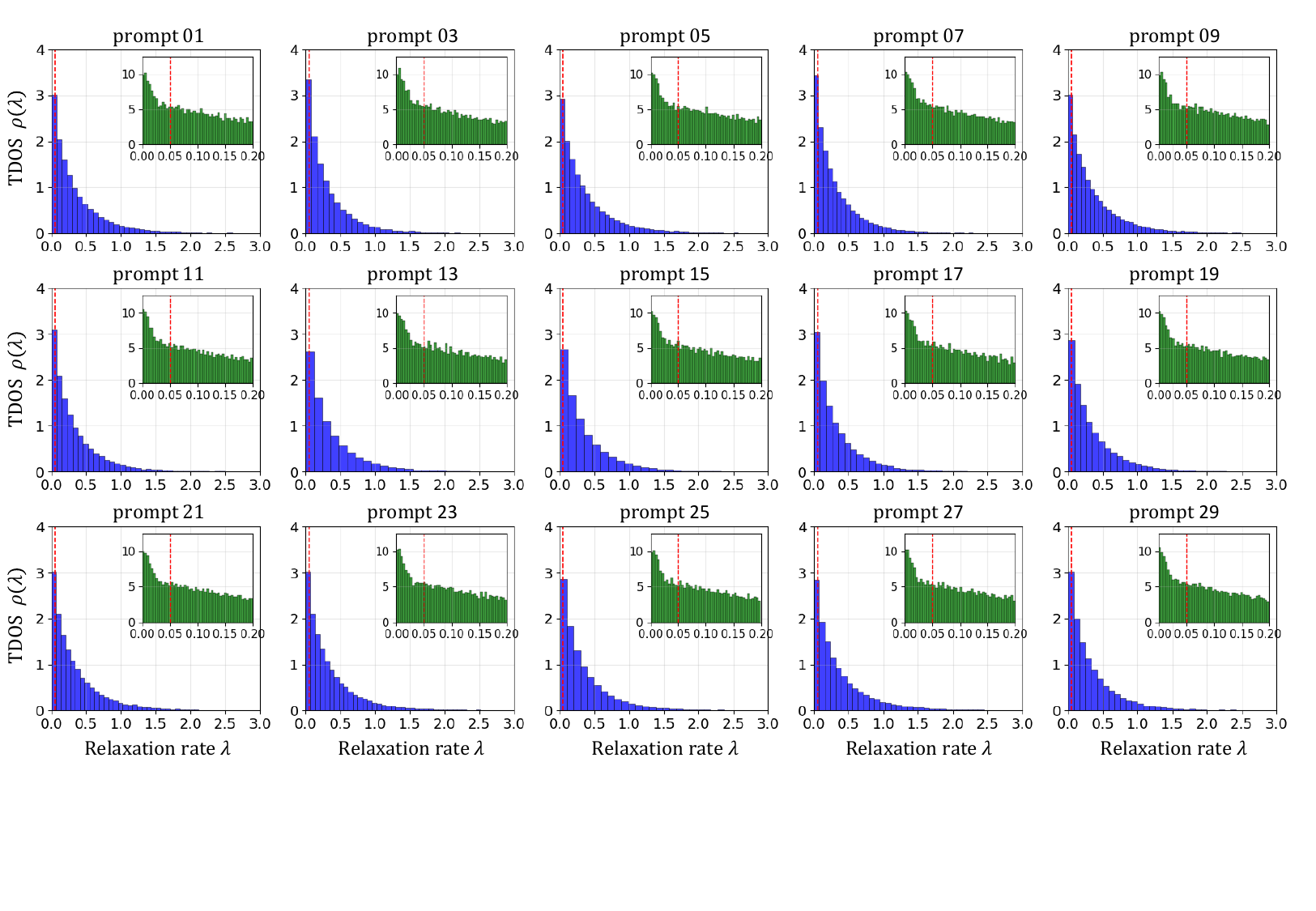}
\caption{
Prompt-resolved normalized time-scale density of states measured from
the Layer~10$\rightarrow$11 mapping of the fully trained Pythia-410M
model using twenty-four-token input subspaces.
Fifteen representative prompts from the thirty-prompt ensemble are shown.
The main panels display the full TDOS, while the insets enlarge the
infrared region \(0<\lambda<0.2\).
The red dashed line denotes the reference cutoff
\(\lambda_{\rm cut}=0.05\).
The spectra exhibit the same overall relaxation-rate distribution
across distinct prompt-dependent hidden states.
}
\label{fig:prompt_resolved_24token}
\end{figure*}

\begin{figure*}[t]
\centering
\includegraphics[width=1.0\textwidth, trim=0cm 3.5cm 0cm 0cm]{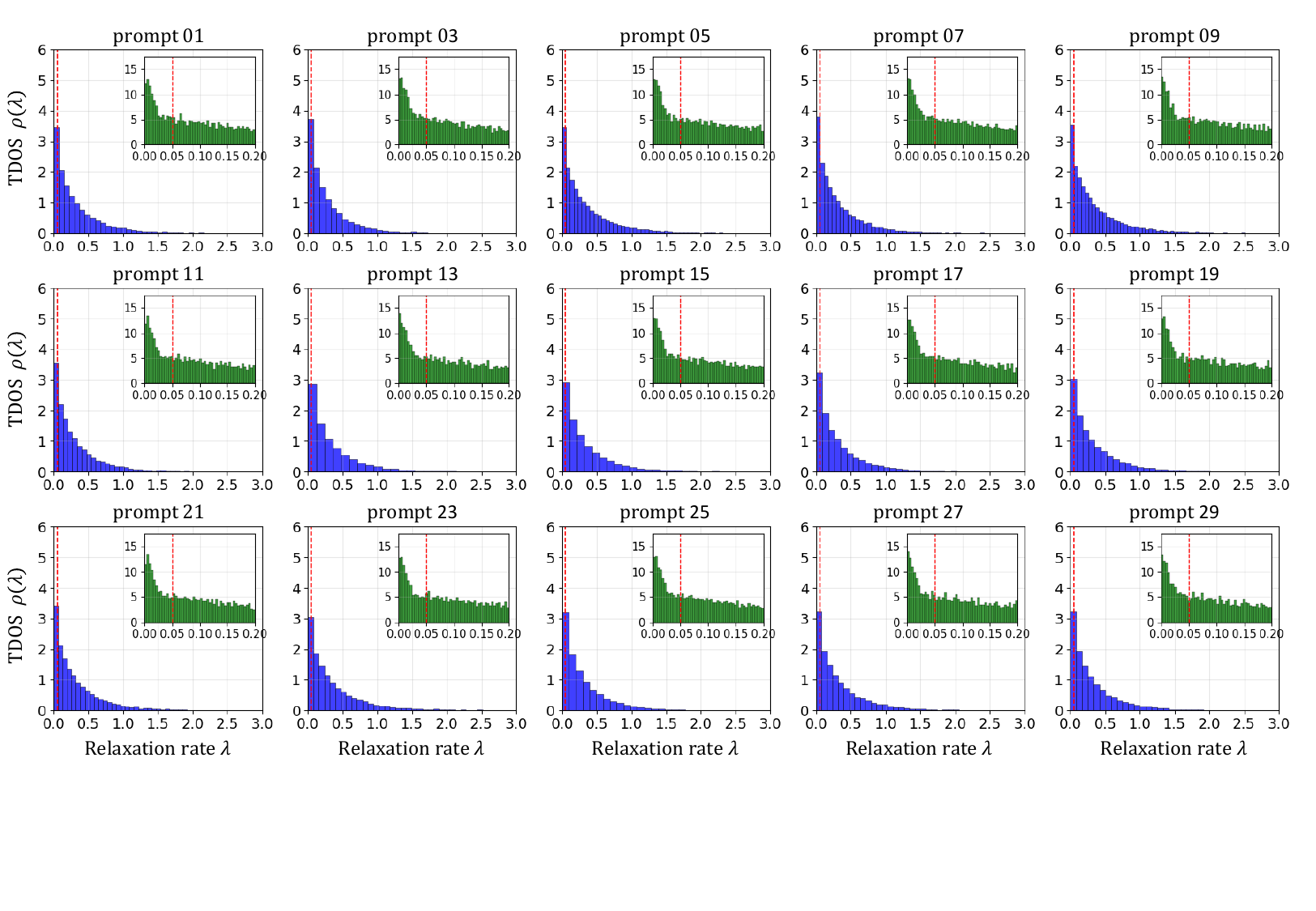}
\caption{
Prompt-resolved normalized time-scale density of states measured from
the Layer~10$\rightarrow$11 mapping of the fully trained Pythia-410M
model using sixteen-token input subspaces.
Representative prompts from the thirty-prompt ensemble are shown.
The main panels display the full TDOS, while the insets enlarge the
infrared region \(0<\lambda<0.2\).
The red dashed line denotes the reference cutoff
\(\lambda_{\rm cut}=0.05\).
}
\label{fig:phase_portrait}
\end{figure*}

\begin{figure*}[t]
\centering
\includegraphics[scale=0.4, trim= 0cm 3cm 0cm 0cm]{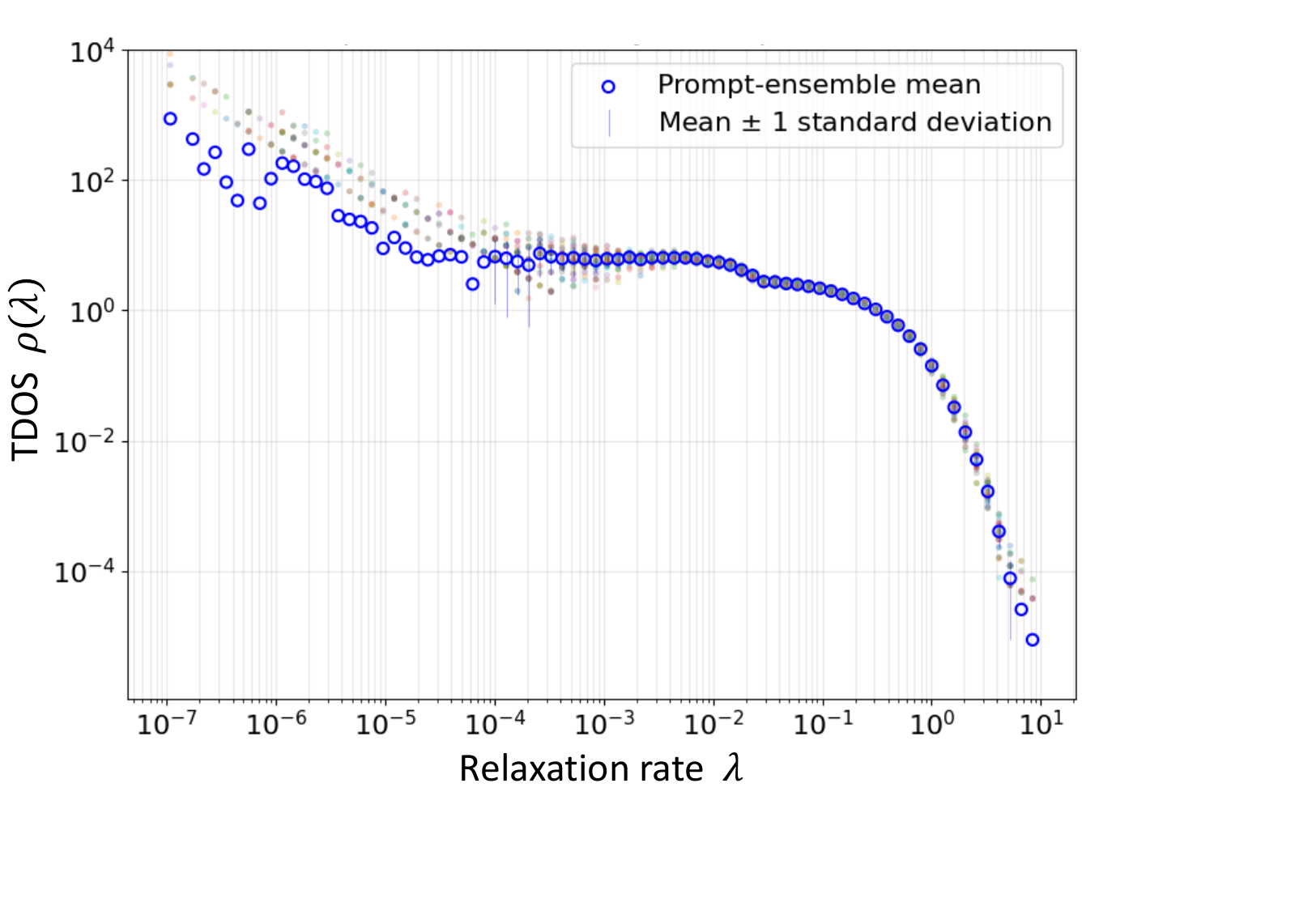}
\caption{
Prompt-ensemble concentration of the normalized TDOS obtained using
sixteen-token input subspaces.
Colored symbols denote the spectra measured from individual prompts,
while open blue circles indicate the ensemble mean.
Vertical bars represent one standard deviation across the prompt
ensemble.
The overall behavior agrees well with the corresponding 24-token
results presented in the main text.
}
\label{fig:phase_portrait}
\end{figure*}

\begin{figure*}[t]
\centering
\includegraphics[width=1.0\textwidth, trim=0cm 3.5cm 0cm 0cm]{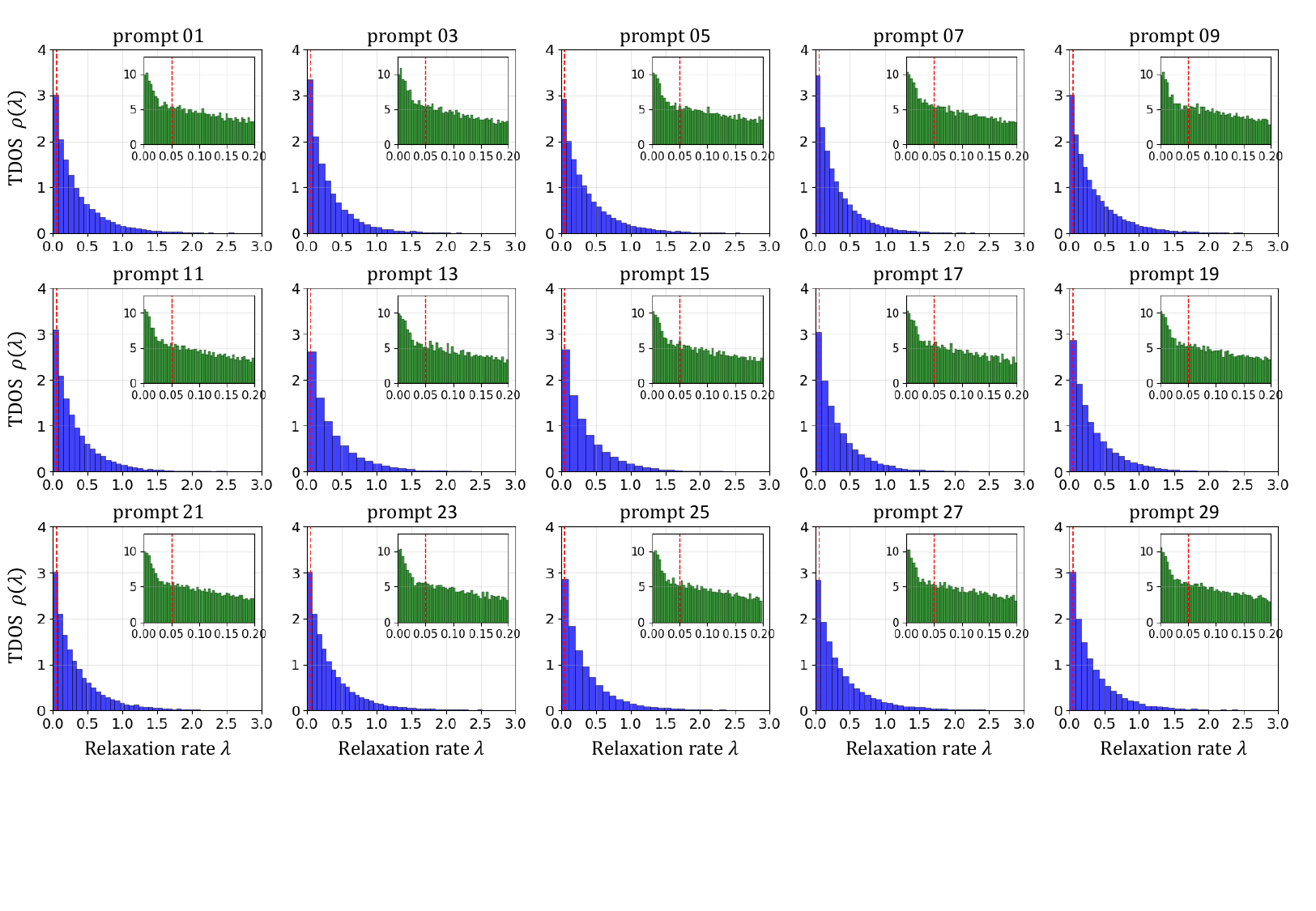}
\caption{
Prompt-resolved normalized time-scale density of states measured from
the Layer~10$\rightarrow$11 mapping of the fully trained Pythia-410M
model using twenty-eight-token input subspaces.
Representative prompts from the thirty-prompt ensemble are shown.
The main panels display the full TDOS, while the insets enlarge the
infrared region \(0<\lambda<0.2\).
The red dashed line denotes the reference cutoff
\(\lambda_{\rm cut}=0.05\).
}
\label{fig:phase_portrait}
\end{figure*}

\begin{figure*}[t]
\centering
\includegraphics[scale=0.4, trim= 0cm 3cm 0cm 0cm]{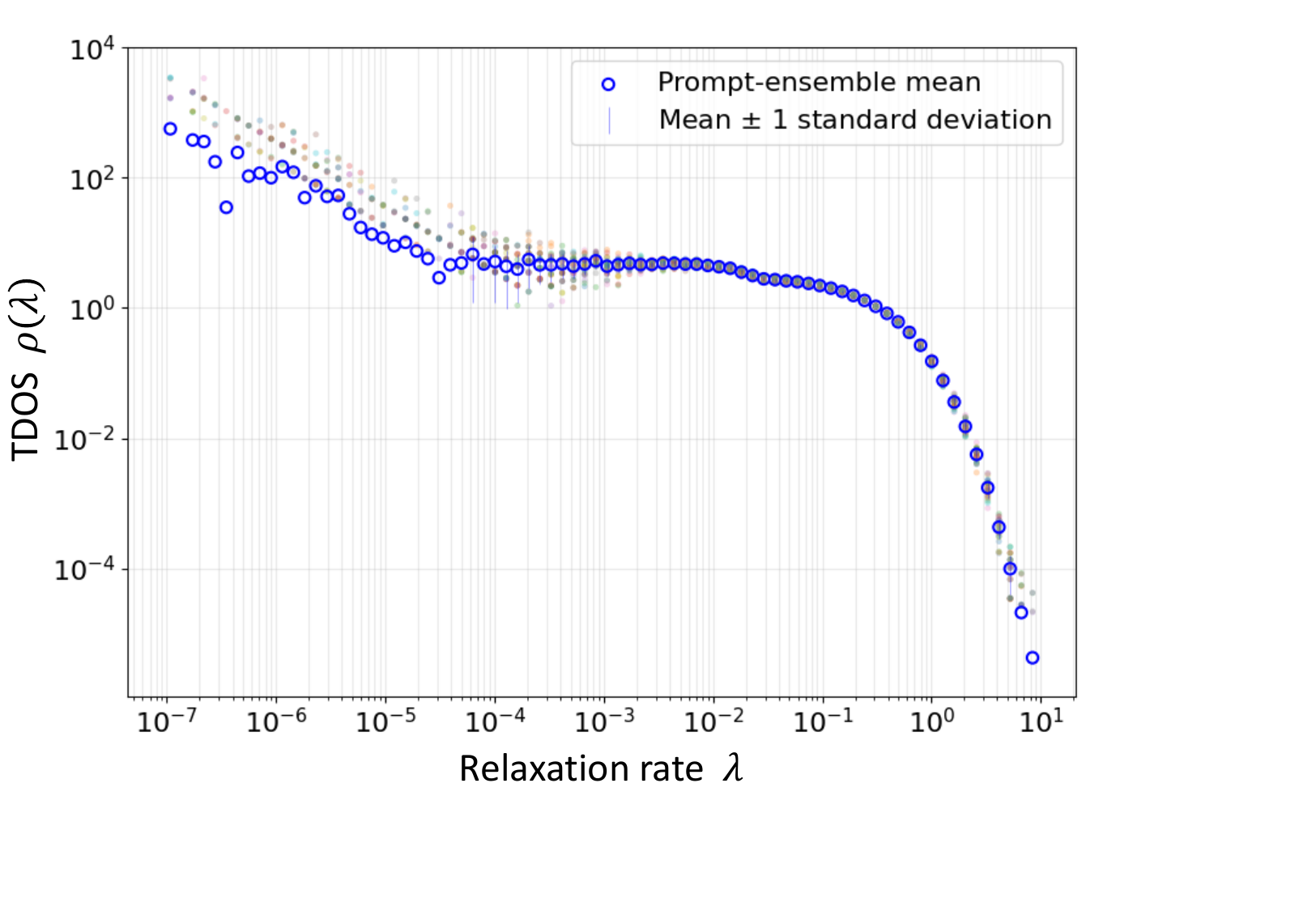}
\caption{
Prompt-ensemble concentration of the normalized TDOS obtained using
twenty-eight-token input subspaces.
Colored symbols denote the spectra measured from individual prompts,
while open blue circles indicate the ensemble mean.
Vertical bars represent one standard deviation across the prompt
ensemble.
The overall behavior is consistent with the corresponding 24-token
results presented in the main text.
}
\label{fig:phase_portrait}
\end{figure*}

%%%%%%%%%%%%%%%%%%%%%%%%%%%%%%%%%%%%%%%%%%%%%%%%%%%%%%%%%%%%%%%%%%%%%%%%%%%%%%
%% Appendix D
%%%%%%%%%%%%%%%%%%%%%%%%%%%%%%%%%%%%%%%%%%%%%%%%%%%%%%%%%%%%%%%%%%%%%%%%%%%%%%

\clearpage
\renewcommand{\thefigure}{D\arabic{figure}}
\setcounter{figure}{0}

%===============================================

\section{Appendix D: Pythia-70M model}

To examine the robustness of the observed infrared organization with
respect to model scale, we repeated the complete spectral analysis for
the Pythia-70M model.
The same procedure used in the main text was applied: Jacobian spectra
were extracted at representative training checkpoints, relaxation rates
were constructed from the eigenvalue magnitudes, and the corresponding
TDOS, memory self-energy, forgetting-gap proxy, and memory kernels were
evaluated.

Figures~D1--D7 summarize the results.
Figure~D1 shows that the TDOS of the Pythia-70M model undergoes
progressive infrared reorganization during training, with slow
relaxation modes accumulating near the infrared sector.
Although the smaller model size leads to stronger finite-size
fluctuations in the TDOS compared with Pythia-410M, the qualitative
evolution remains the same.

Figure~D2 shows the corresponding evolution of the relaxation-mode
populations.
The number of stable modes increases during optimization, whereas the
unstable-mode population decreases, while the total number of finite
modes remains approximately constant.
This confirms that training reorganizes the existing relaxation
spectrum rather than generating new collective degrees of freedom.

Figure~D3 presents the normalized memory self-energy and the
corresponding normalized cognitive forgetting-gap.
As in the larger Pythia-410M model, the memory self-energy exhibits a
transient maximum during training, while the forgetting-gap proxy
reaches its minimum at the same stage.
This reproduces the critical-formation scenario discussed in the main
text.

Figure~D4 compares the infrared TDOS with the memory kernel computed
from the measured relaxation spectrum.
The memory kernel exhibits an extended scale-free decay, demonstrating
that long-time collective memory is preserved even in the smaller
Pythia-70M model.

Figures~D5--D7 show the layer-dependent organization after training.
The TDOS is broadly distributed across all hidden-layer mappings,
while the memory self-energy increases toward later layers and the
forgetting-gap proxy decreases correspondingly.
The associated memory kernels retain a common scale-free temporal form
across layers, indicating that the cognitive field remains spatially
distributed and dynamically universal within the smaller Transformer
hierarchy.

Taken together, these results demonstrate that the principal infrared
features reported in the main text are already present in the
Pythia-70M model.
Despite its substantially smaller size, Pythia-70M exhibits slow-mode
accumulation, transient memory self-energy enhancement,
forgetting-gap reduction, scale-free memory kernels, and
layer-dependent collective organization.
This supports the conclusion that infrared slow-mode organization is
not specific to the representative Pythia-410M model but persists
across Transformer model scales.

\begin{figure*}[t]
\centering
\includegraphics[width=1.0\textwidth, trim=0cm 1.7cm 0cm 0cm]{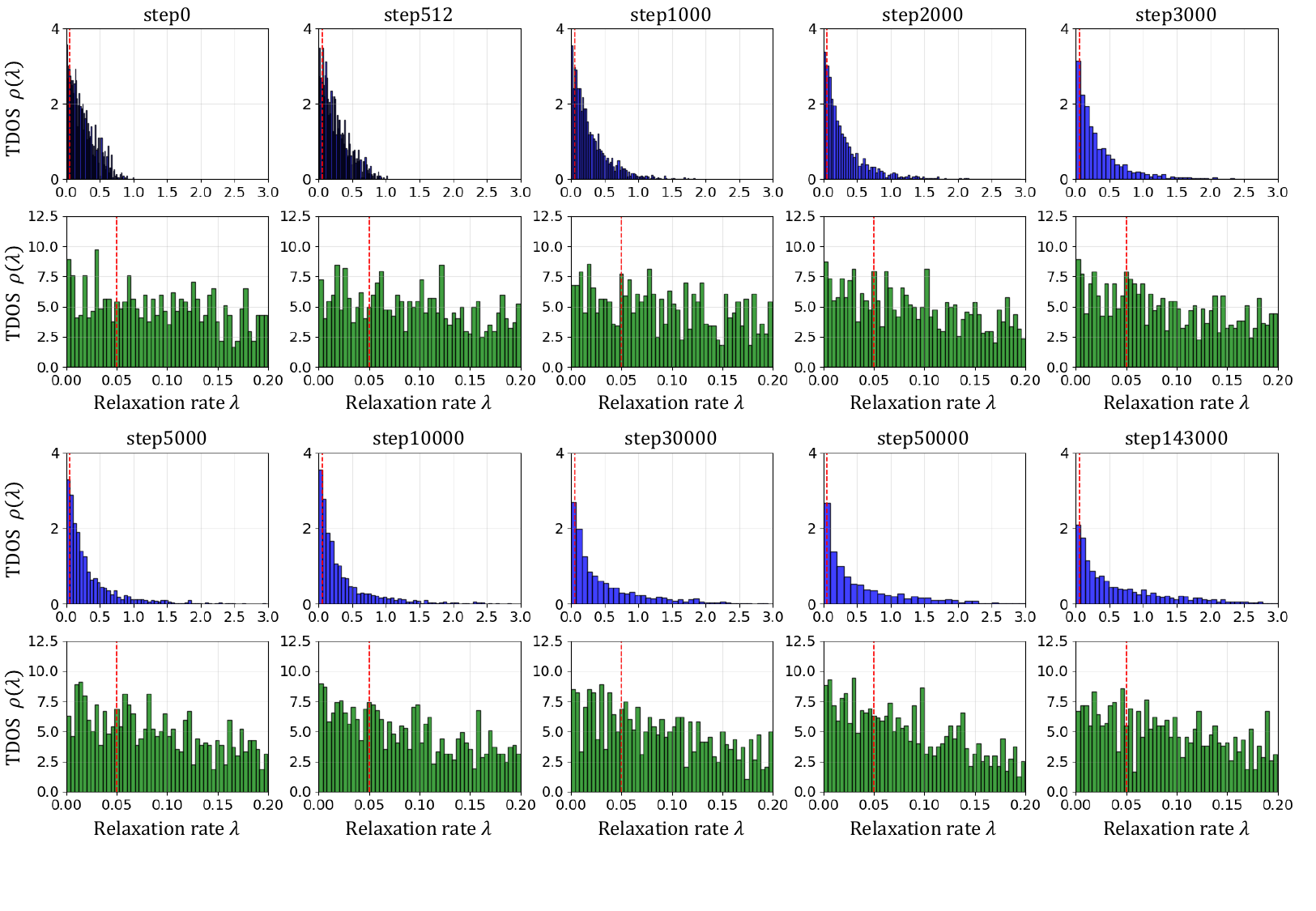}
\caption{
Evolution of the TDOS during training for the Pythia-70M model.
The TDOS exhibits progressive infrared reorganization, reproducing the
qualitative behavior observed in the larger Pythia models.
}
\label{fig:phase_portrait}
\end{figure*}

\begin{figure*}[t]
\centering
\includegraphics[scale=0.55, trim= 0cm 8cm 9cm 0cm]{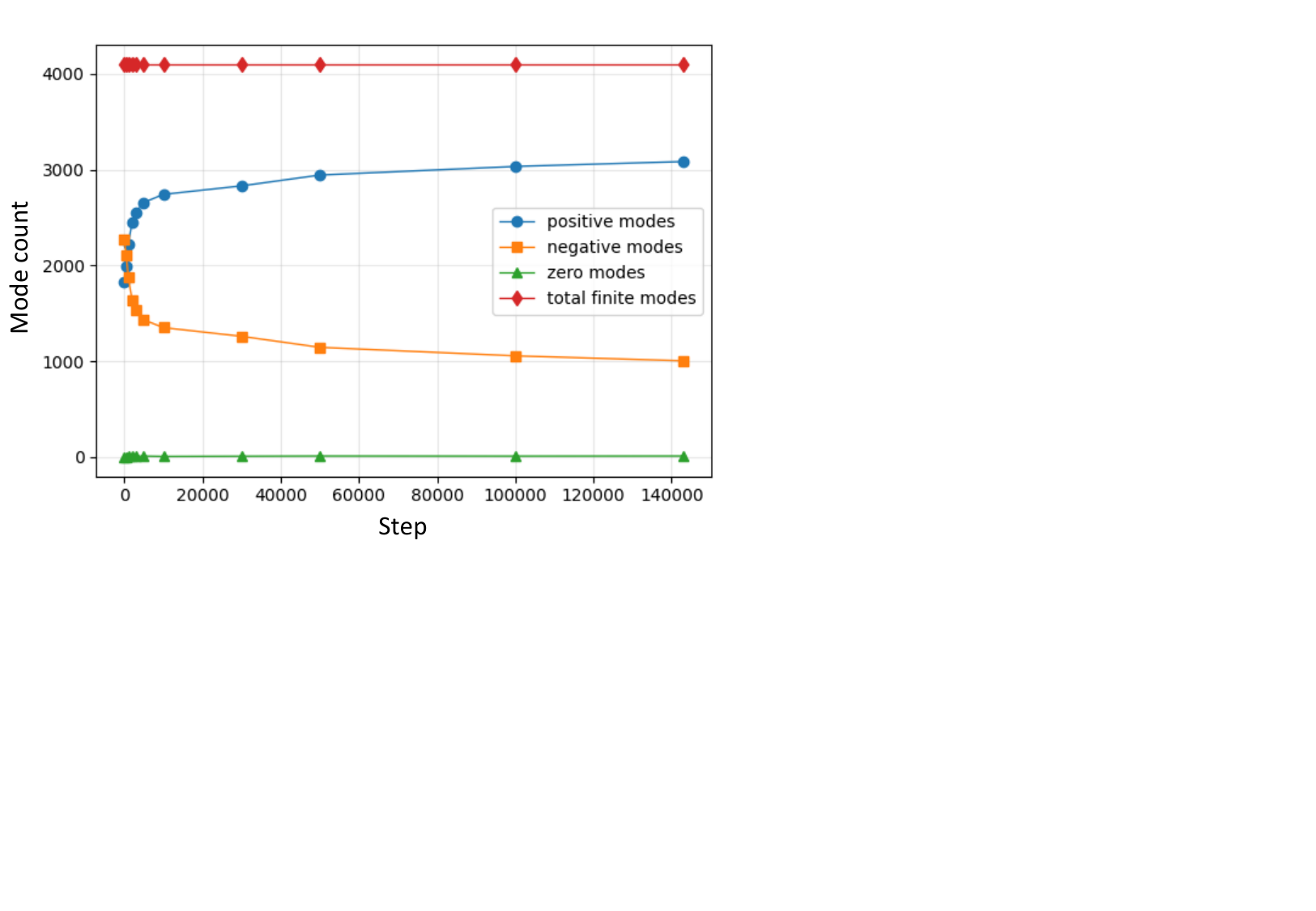}
\caption{
Evolution of stable, unstable, zero, and total finite relaxation modes
for the Pythia-70M model.
Training increases the stable-mode population while reducing the
unstable sector.
}
\label{fig:phase_portrait}
\end{figure*}

\begin{figure*}[t]
\centering
\includegraphics[scale=0.55, trim= 0cm 0.58cm 9cm 0cm]{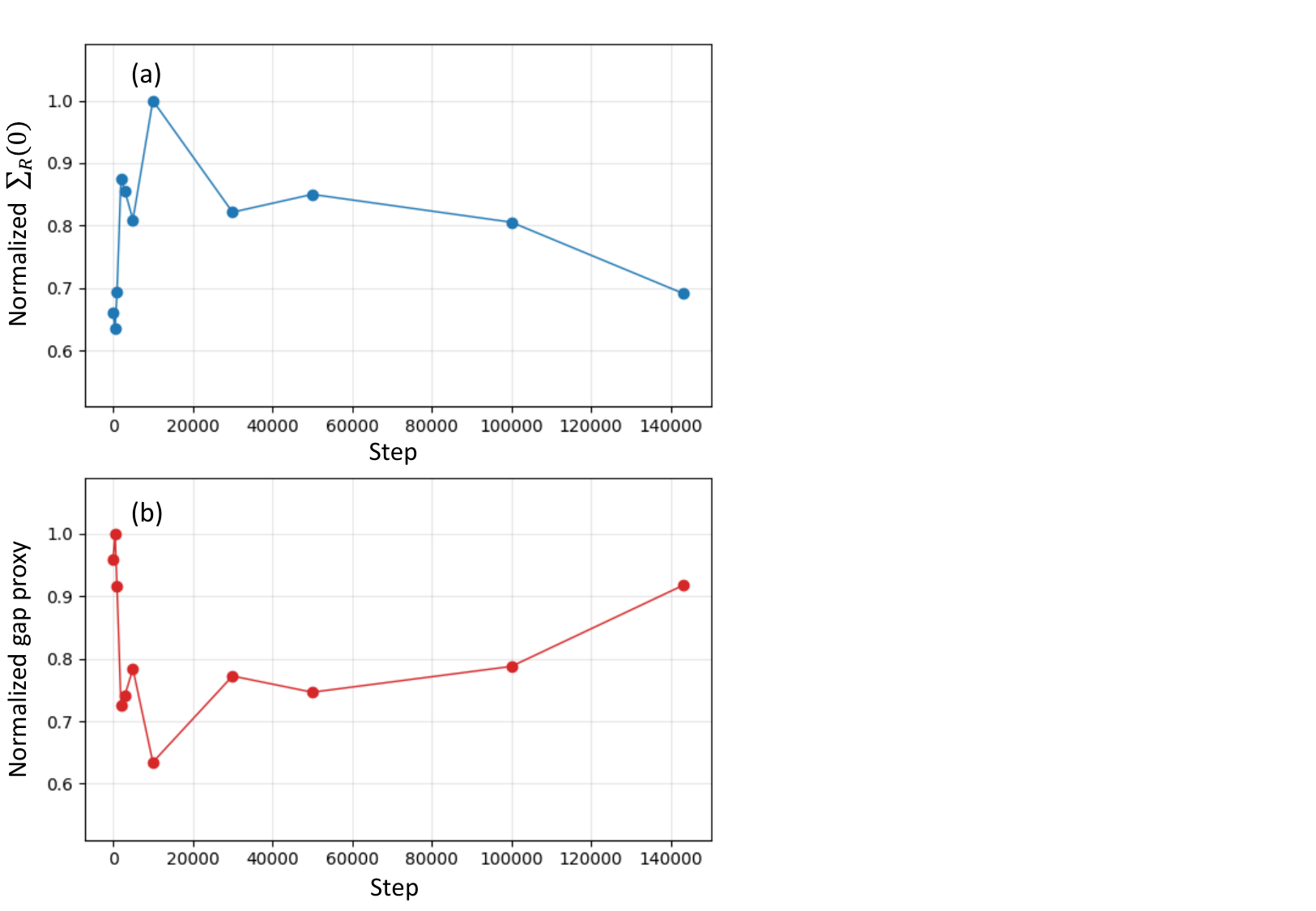}
\caption{
Normalized memory self-energy and corresponding normalized
forgetting-gap proxy for the Pythia-70M model.
The transient maximum of the memory self-energy and the associated
minimum of the forgetting gap reproduce the critical-formation
scenario.
}
\label{fig:phase_portrait}
\end{figure*}

\begin{figure*}[t]
\centering
\includegraphics[scale=0.5, trim= 0.2cm 0.5cm 8cm 0cm]{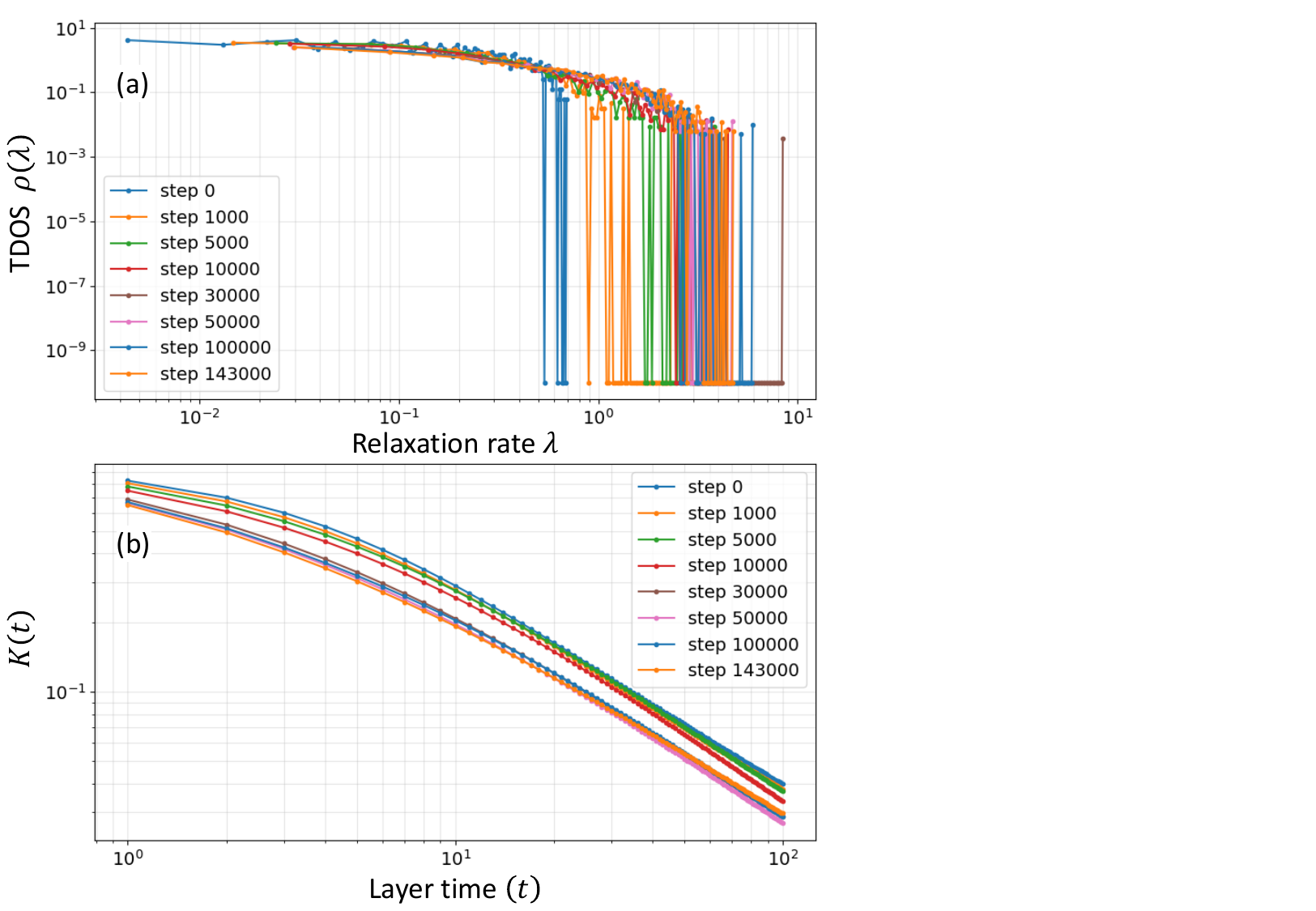}
\caption{
Infrared TDOS and memory kernel for the Pythia-70M model.
The measured relaxation spectrum generates a scale-free long-time
memory kernel.
}
\label{fig:phase_portrait}
\end{figure*}

\begin{figure*}[t]
\centering
\includegraphics[width=1.0\textwidth, trim=0cm 9cm 0cm 0cm]{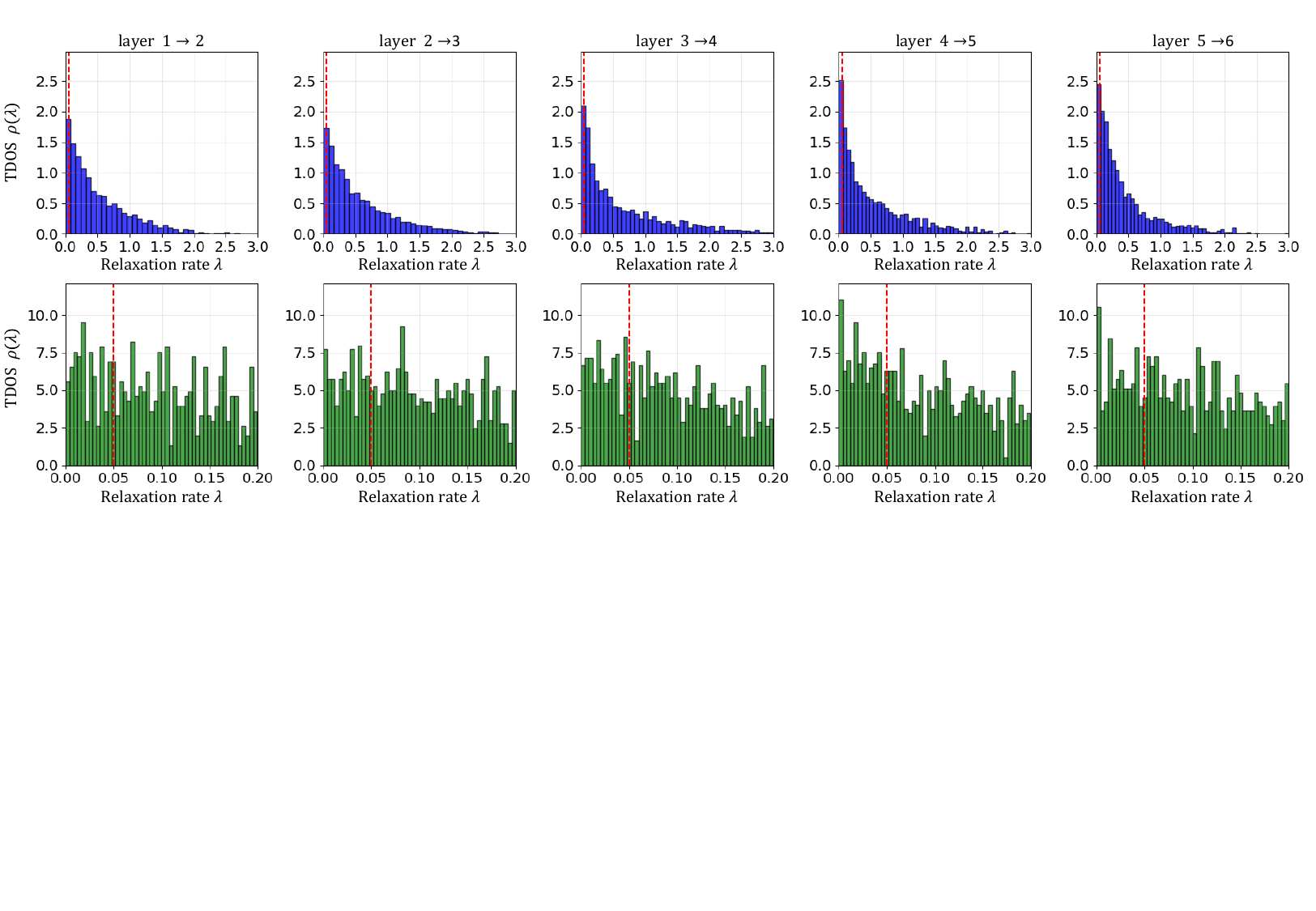}
\caption{
Layer-dependent TDOS of the fully trained Pythia-70M model.
Infrared slow-mode organization is observed across the Transformer
hierarchy.
}
\label{fig:phase_portrait}
\end{figure*}

\begin{figure*}[t]
\centering
\includegraphics[scale=0.55, trim= 0cm 0.5cm 9cm 0cm]{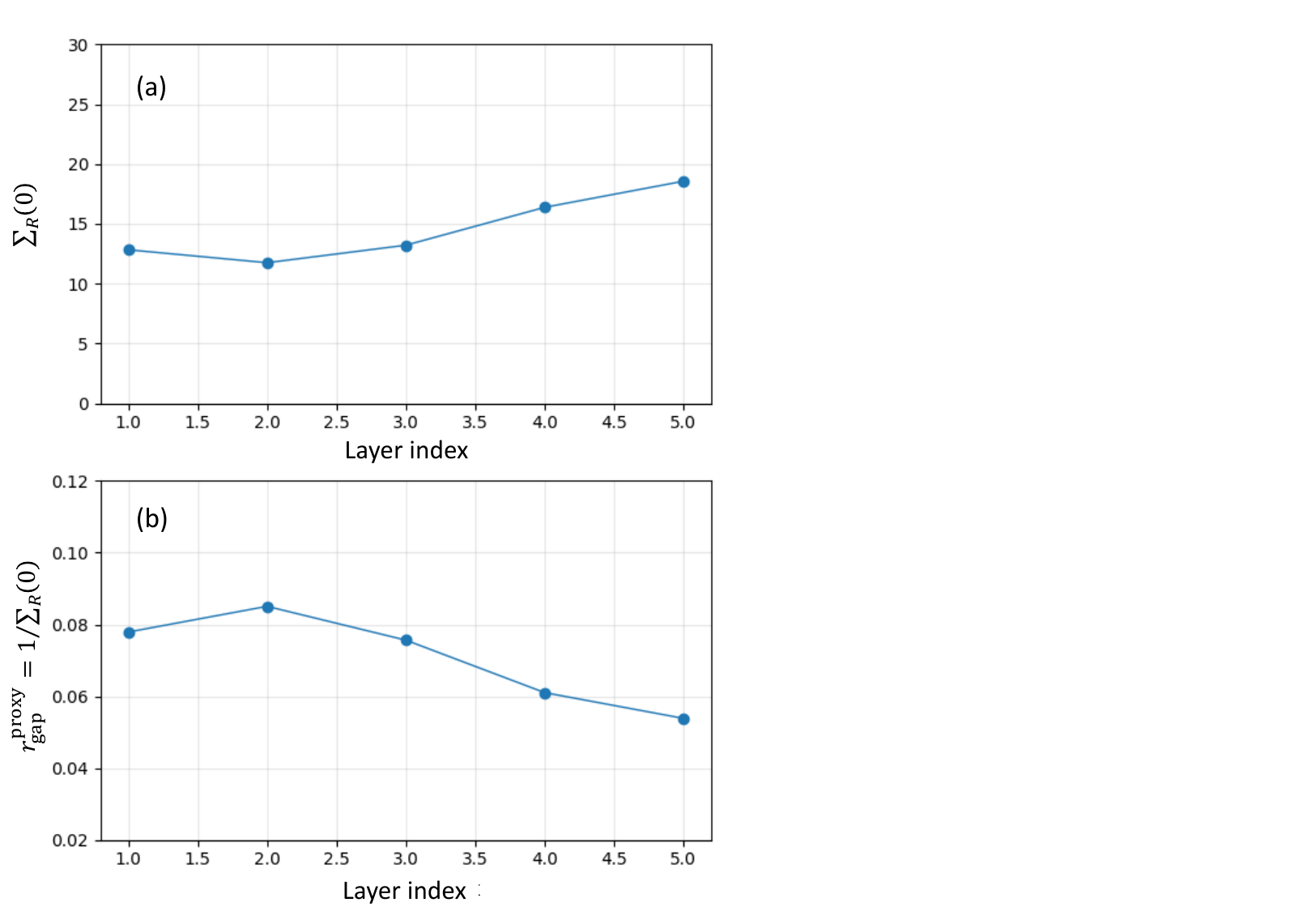}
\caption{
Layer-dependent memory self-energy and forgetting-gap proxy for the
fully trained Pythia-70M model.
Later hidden layers exhibit stronger memory renormalization and smaller
forgetting gaps.
}
\label{fig:phase_portrait}
\end{figure*}

\begin{figure*}[t]
\centering
\includegraphics[scale=0.5, trim= 0cm 0.5cm 8cm 0cm]{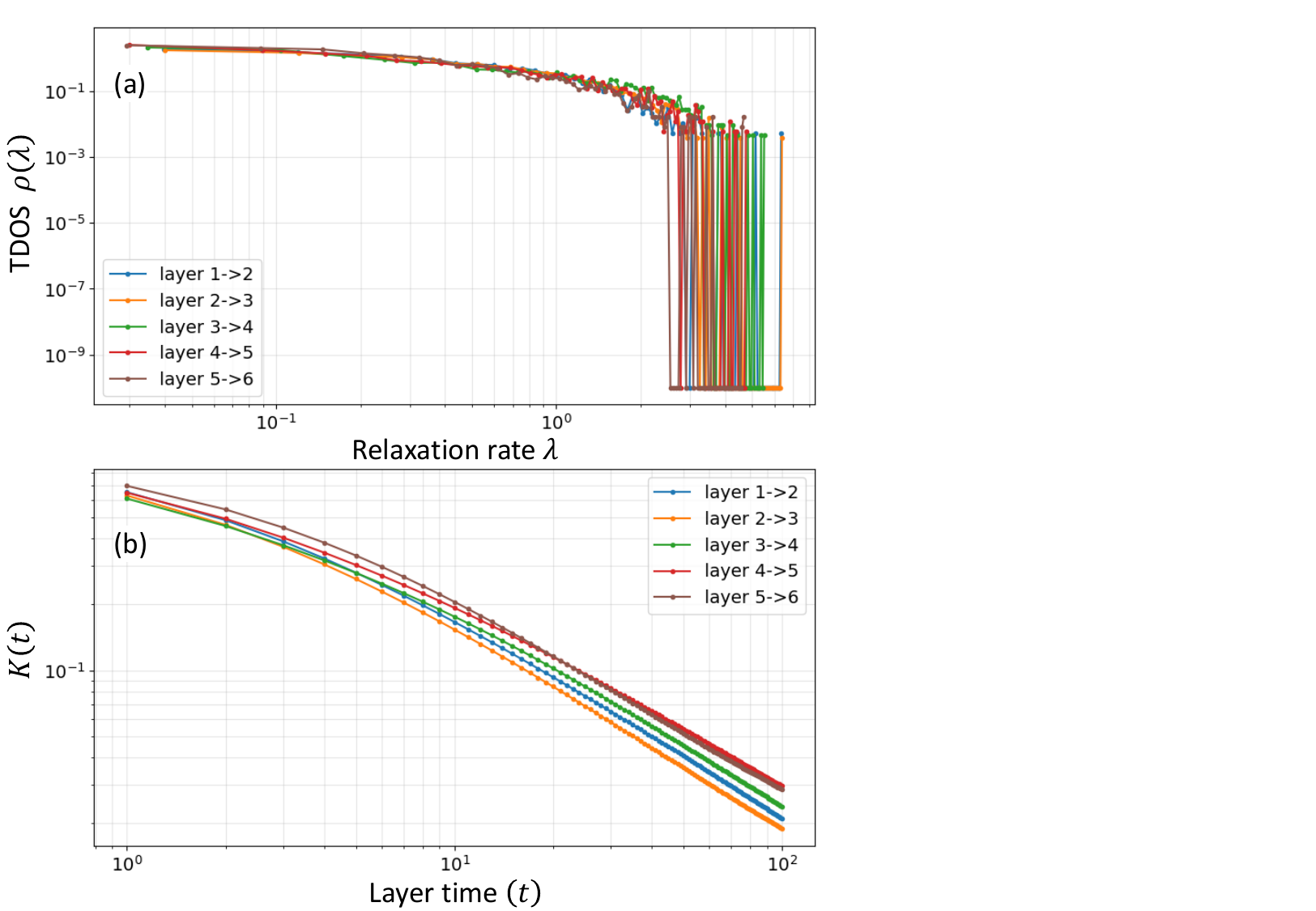}
\caption{
Layer-dependent TDOS and memory kernels for the fully trained
Pythia-70M model.
All investigated layers retain a common scale-free infrared
organization.
}
\label{fig:phase_portrait}
\end{figure*}

%%%%%%%%%%%%%%%%%%%%%%%%%%%%%%%%%%%%%%%%%%%%%%%%%%%%%%%%%%%%%%%%%%%%%%%%%%%%%%
%% Appendix D
%%%%%%%%%%%%%%%%%%%%%%%%%%%%%%%%%%%%%%%%%%%%%%%%%%%%%%%%%%%%%%%%%%%%%%%%%%%%%%

\clearpage
\renewcommand{\thefigure}{E\arabic{figure}}
\setcounter{figure}{0}

\section{Appendix E: Pythia-1.4B model}

To further examine the universality of the infrared organization at
larger model scales, we repeated the complete analysis for the
Pythia-1.4B model using exactly the same measurement protocol as in the
main text.
Jacobian spectra were extracted at representative training
checkpoints, from which the relaxation spectrum, TDOS, memory
self-energy, forgetting-gap proxy, and memory kernels were evaluated.

Figures~E1--E7 summarize the corresponding results.
Figure~E1 shows the evolution of the TDOS throughout training.
As in the Pythia-410M and Pythia-70M models, optimization progressively
transfers spectral weight toward the infrared sector, producing a
pronounced accumulation of slow relaxation modes.
The infrared organization therefore persists even in the largest model
investigated in the present work.

Figure~E2 presents the corresponding evolution of the relaxation-mode
populations.
The number of stable relaxation modes increases continuously during
training, whereas the unstable-mode population decreases, while the
total number of finite modes remains approximately constant.
The overall spectral evolution closely follows the behavior reported
for the smaller Pythia models.

Figure~E3 shows the normalized memory self-energy together with the
corresponding normalized cognitive forgetting-gap.
The memory self-energy again exhibits a pronounced transient maximum
during the early stage of optimization, while the forgetting-gap proxy
reaches its minimum at the same stage.
This reproduces the transient critical formation of the cognitive
field discussed in the main text.

Figure~E4 compares the infrared TDOS with the corresponding memory
kernels.
Despite the larger model size, the memory kernel retains an extended
power-law decay, demonstrating that the scale-free long-time memory
predicted by Cognitive Field Theory remains robust.

Figures~E5--E7 summarize the layer-dependent analysis after completion
of training.
The TDOS exhibits pronounced infrared accumulation throughout the
Transformer hierarchy.
The memory self-energy increases systematically toward the deeper
hidden layers, accompanied by a corresponding reduction of the
forgetting-gap proxy.
Meanwhile, the memory kernels preserve the same scale-free temporal
dependence across all investigated layers, indicating that the
cognitive field remains spatially distributed while sharing a common
infrared dynamical organization.

Taken together, these measurements demonstrate that the principal
results reported in the main text remain valid for the largest
Transformer investigated in the present work.
The progressive infrared accumulation of slow relaxation modes, the
transient enhancement of the memory self-energy, the reduction of the
forgetting gap, the emergence of scale-free memory kernels, and the
distributed layer hierarchy are all reproduced in the Pythia-1.4B
model.
These results provide additional support for the conclusion that
infrared slow-mode organization represents a universal property of
Transformer dynamics over a broad range of model scales.

\begin{figure*}[t]
\centering
\includegraphics[width=1.0\textwidth, trim=0cm 1.7cm 0cm 0cm]{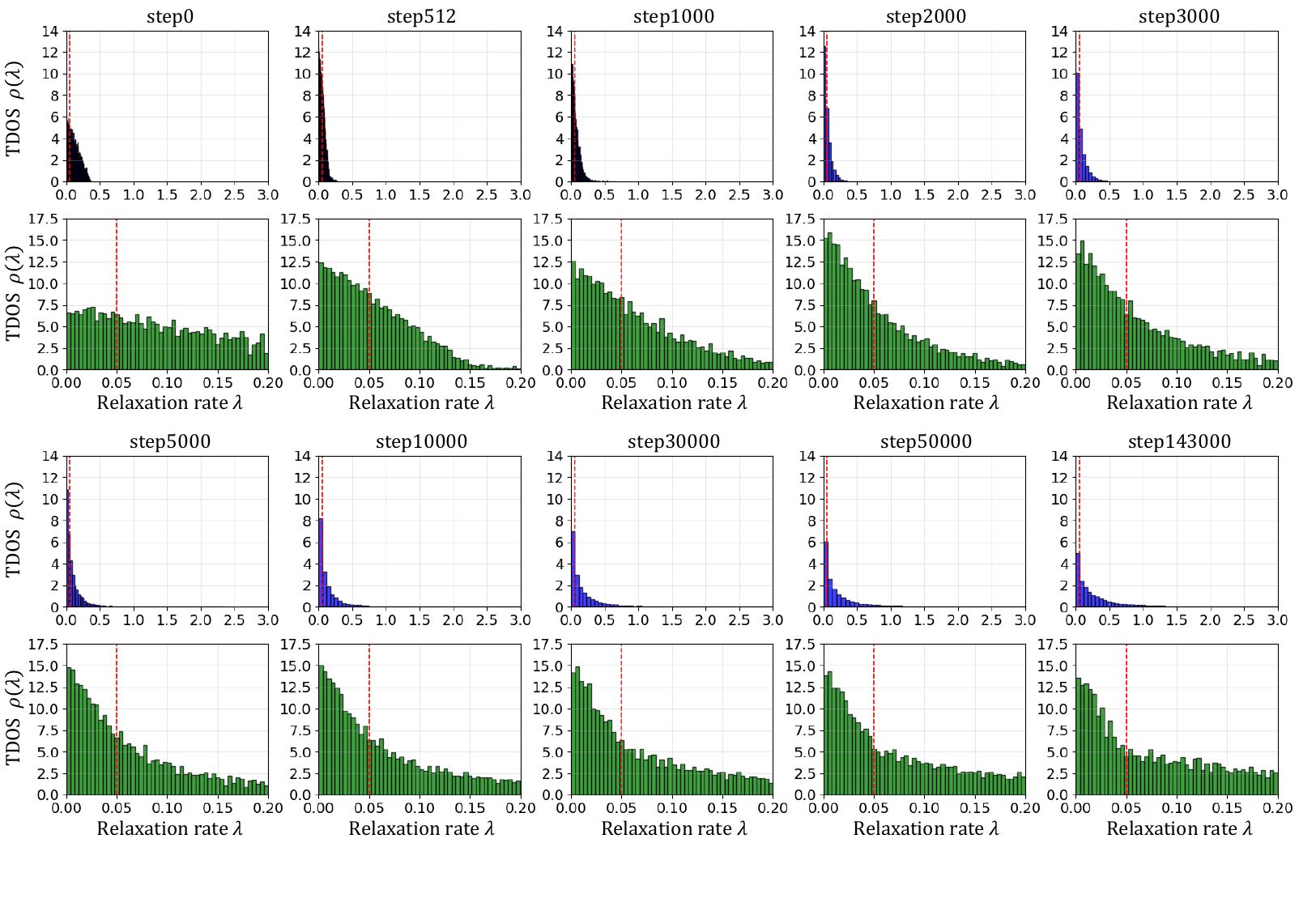}
\caption{
Evolution of the TDOS during training for the Pythia-1.4B model.
The TDOS exhibits progressive infrared reorganization, reproducing the
qualitative behavior observed for the smaller Pythia models.
}
\label{fig:phase_portrait}
\end{figure*}

\begin{figure*}[t]
\centering
\includegraphics[scale=0.55, trim= 0cm 8cm 9cm 0cm]{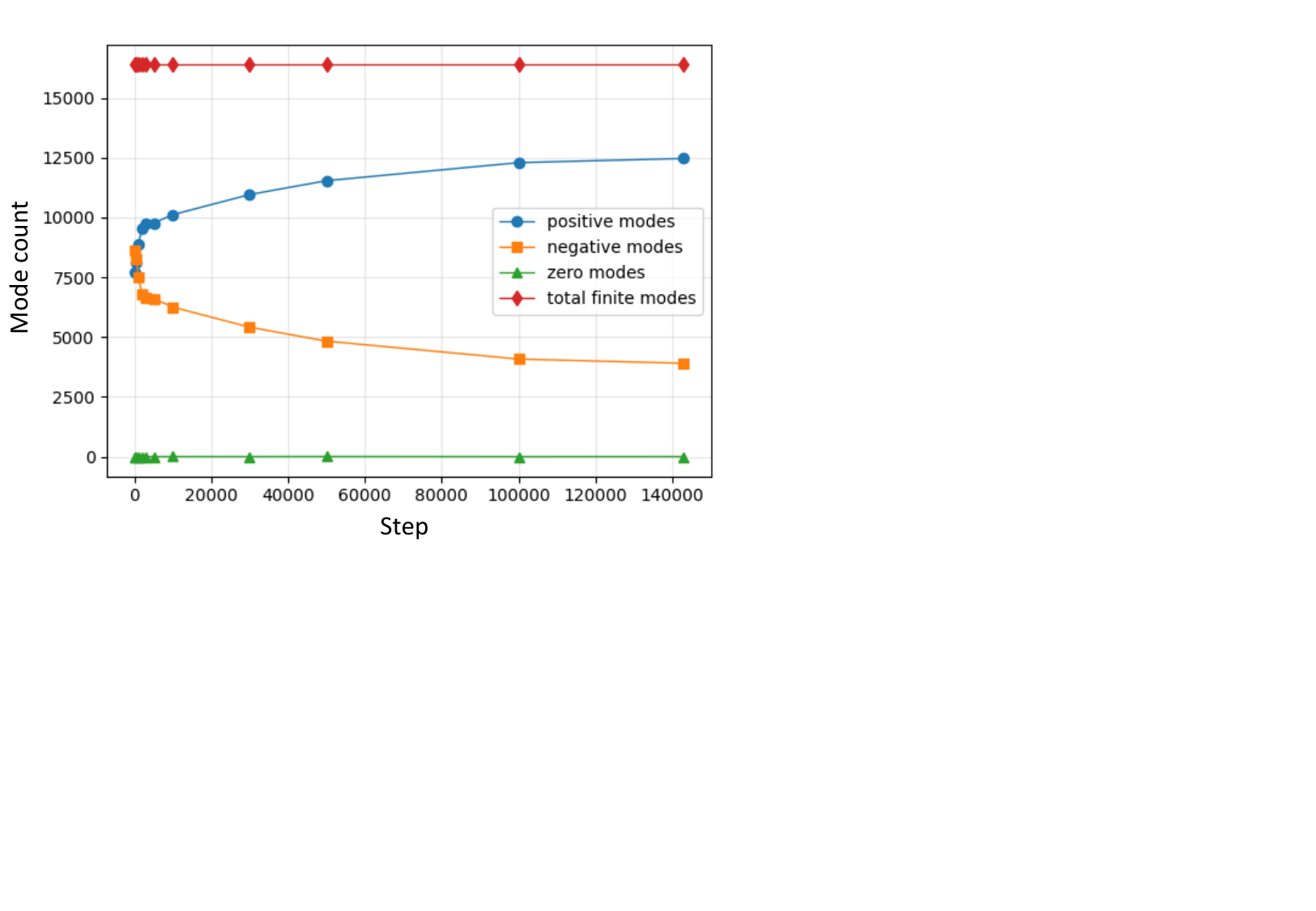}
\caption{
Evolution of stable, unstable, zero, and total finite relaxation modes
for the Pythia-1.4B model.
Training progressively increases the stable-mode population while
reducing the unstable sector.
}
\label{fig:phase_portrait}
\end{figure*}

\begin{figure*}[t]
\centering
\includegraphics[scale=0.55, trim= 0cm 0.58cm 9cm 0cm]{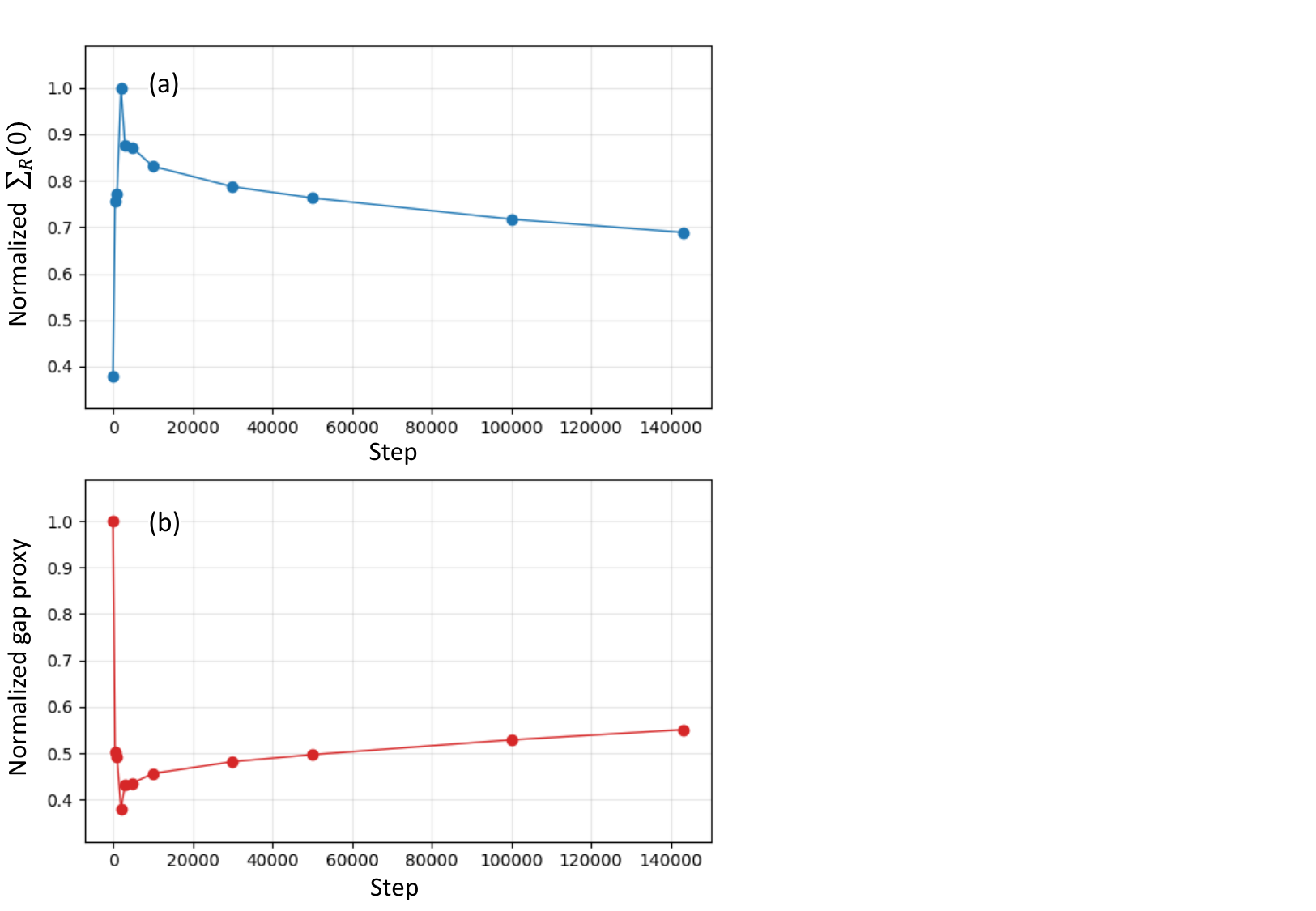}
\caption{
Normalized memory self-energy and corresponding normalized
forgetting-gap proxy for the Pythia-1.4B model.
The transient maximum of the memory self-energy and the associated
minimum of the forgetting-gap proxy reproduce the critical-formation
scenario described in the main text.
}
\label{fig:phase_portrait}
\end{figure*}

\begin{figure*}[t]
\centering
\includegraphics[scale=0.5, trim= 0.2cm 0.5cm 8cm 0cm]{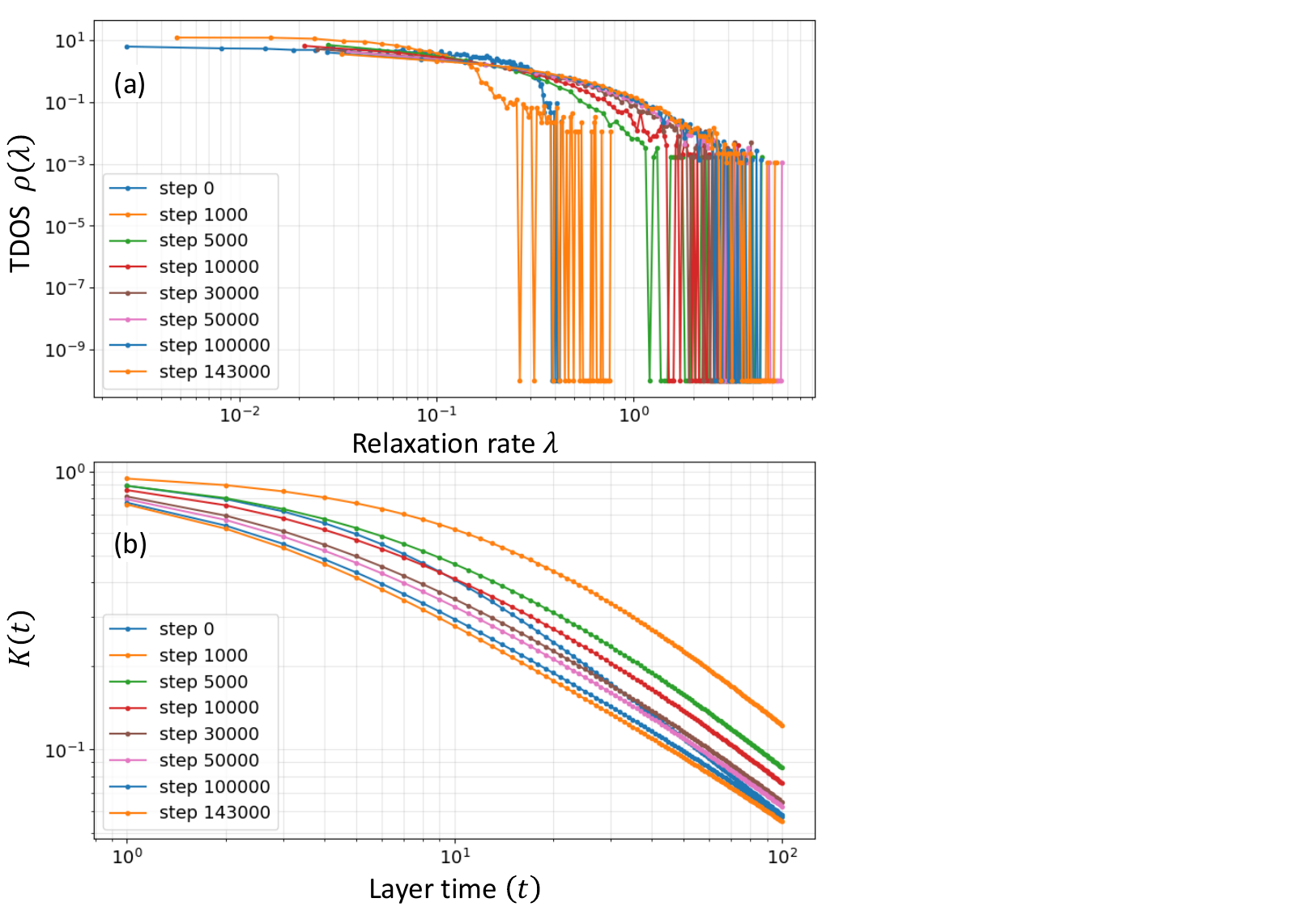}
\caption{
Infrared TDOS and memory kernel for the Pythia-1.4B model.
The measured relaxation spectrum generates robust scale-free
long-time memory throughout training.
}
\label{fig:phase_portrait}
\end{figure*}

\begin{figure*}[t]
\centering
\includegraphics[width=1.0\textwidth, trim=0cm 10cm 0cm 0cm]{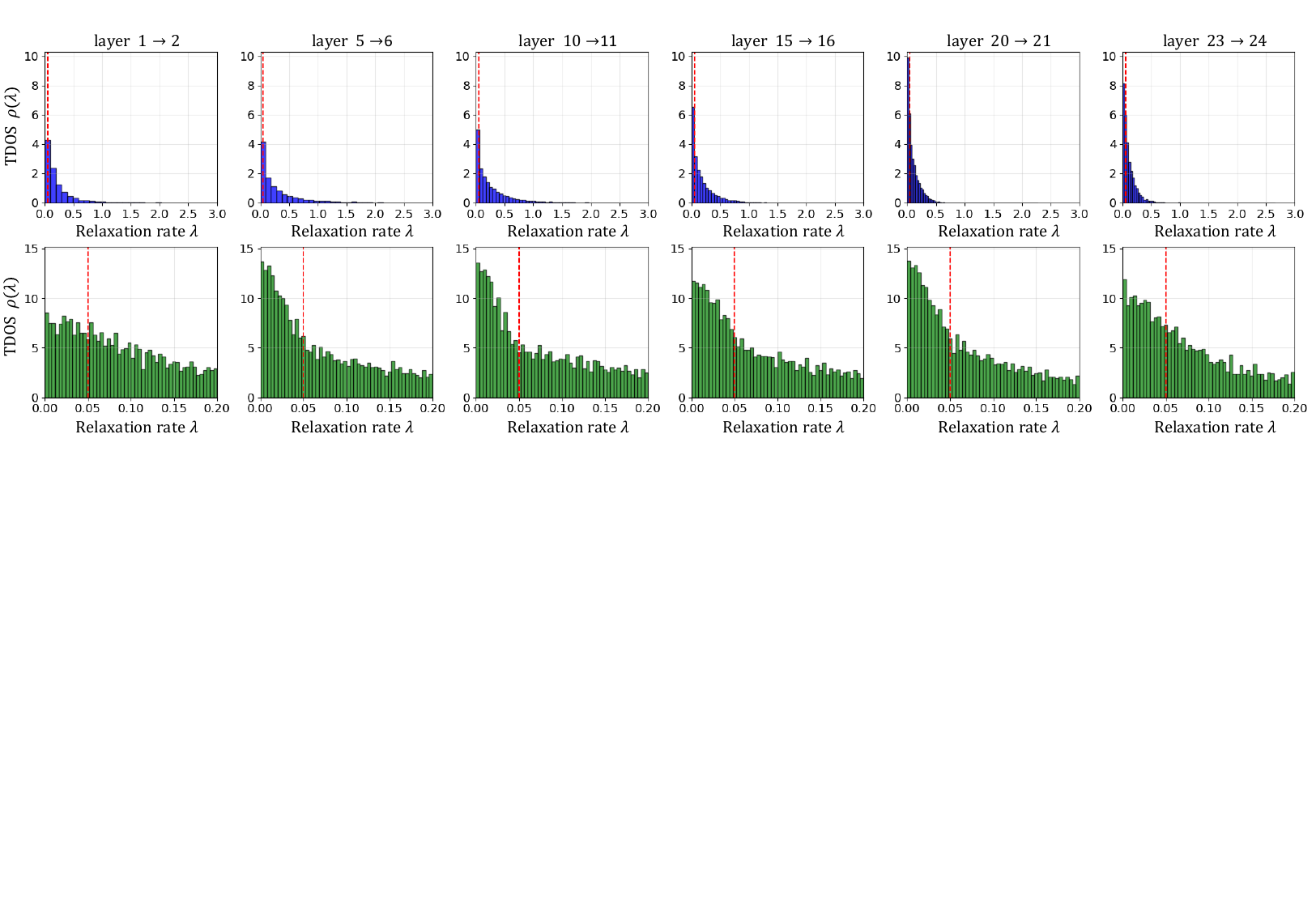}
\caption{
Layer-dependent TDOS of the fully trained Pythia-1.4B model.
Pronounced infrared accumulation of slow relaxation modes is observed
throughout the Transformer hierarchy.
}
\label{fig:phase_portrait}
\end{figure*}

\begin{figure*}[t]
\centering
\includegraphics[scale=0.55, trim= 0cm 0.5cm 9cm 0cm]{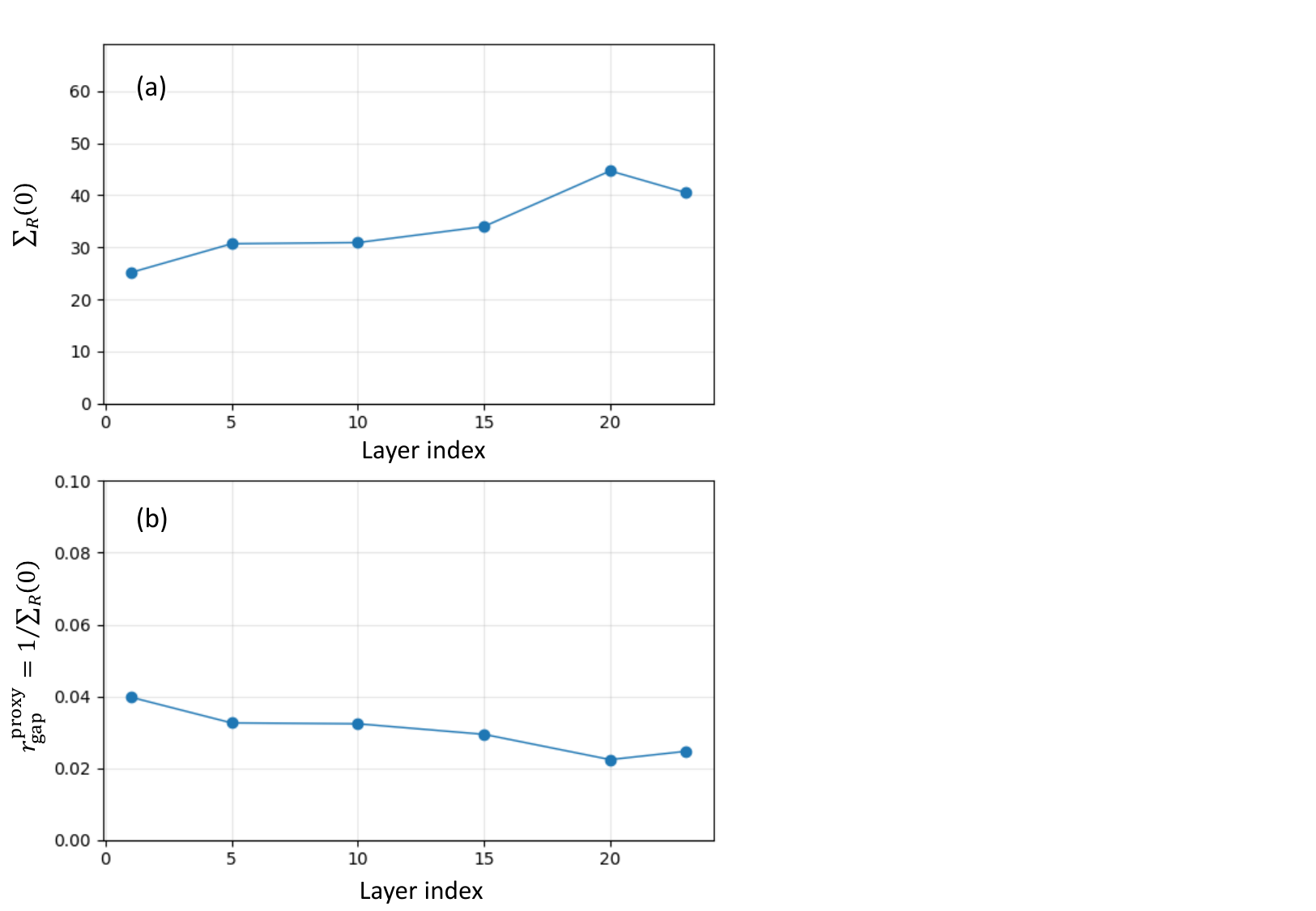}
\caption{
Layer-dependent memory self-energy and forgetting-gap proxy for the
fully trained Pythia-1.4B model.
Deeper hidden layers exhibit stronger memory renormalization and
smaller forgetting-gap proxies than the earlier layers.
}
\label{fig:phase_portrait}
\end{figure*}

\begin{figure*}[t]
\centering
\includegraphics[scale=0.5, trim= 0cm 0.5cm 8cm 0cm]{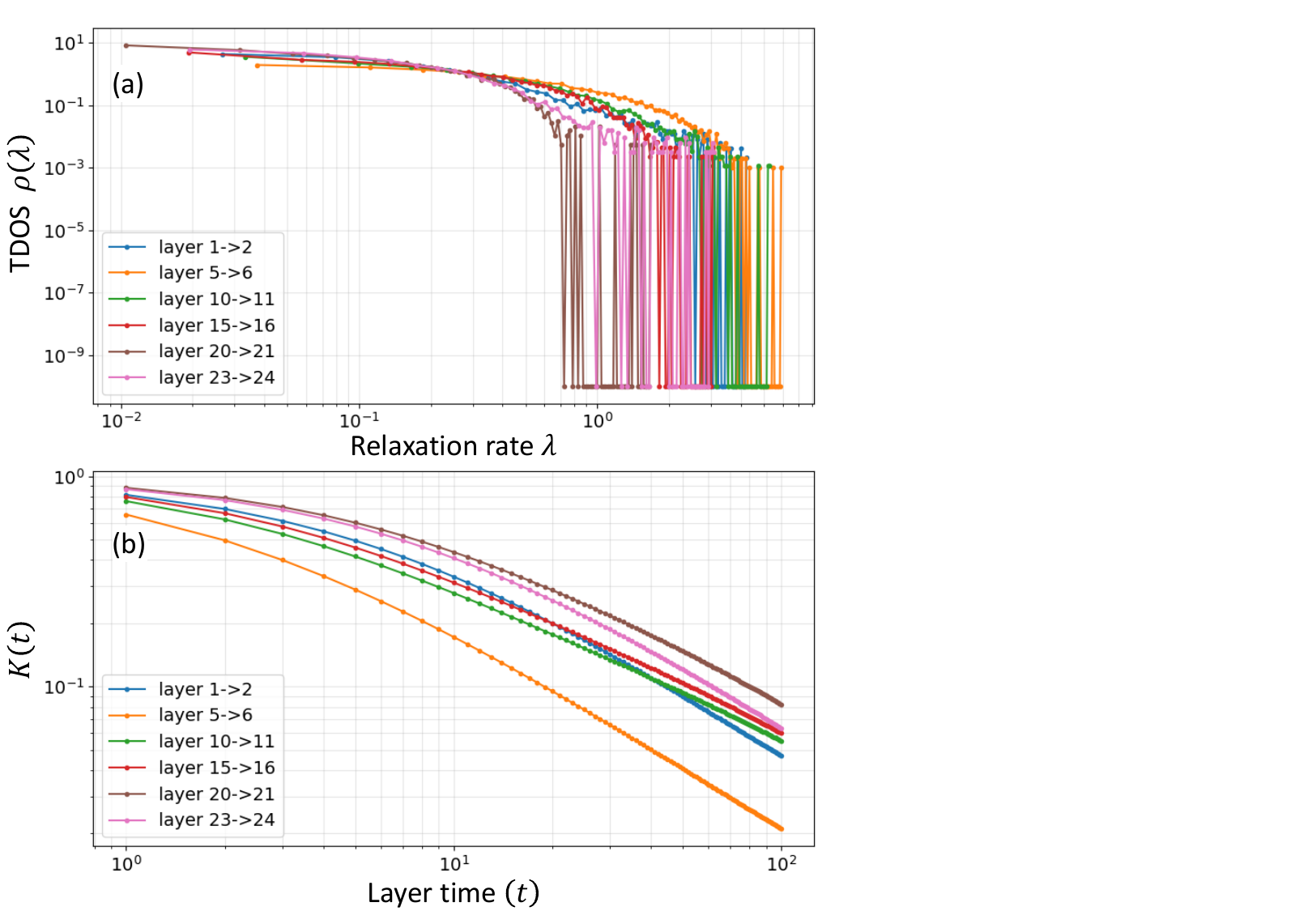}
\caption{
Layer-dependent TDOS and memory kernels for the fully trained
Pythia-1.4B model.
All investigated layers retain the same scale-free infrared
organization despite quantitative variations in memory strength.
}
\label{fig:phase_portrait}
\end{figure*}

\end{document}